\documentclass[10pt,twocolumn,letterpaper]{article}

 \usepackage[pagenumbers]{cvpr} 

\usepackage{graphicx}
\usepackage{amsmath}
\usepackage{amssymb}
\usepackage{booktabs}

\usepackage{multirow}
\usepackage{algorithmic}
\usepackage{algorithm}
\usepackage{caption}
\usepackage[pagebackref,breaklinks,colorlinks]{hyperref}

\usepackage[capitalize]{cleveref}
\crefname{section}{Sec.}{Secs.}
\Crefname{section}{Section}{Sections}
\Crefname{table}{Table}{Tables}
\crefname{table}{Tab.}{Tabs.}

\def\cvprPaperID{863} 
\def\confName{CVPR}
\def\confYear{2023}

\begin{document}

\title{Binary Latent Diffusion}


\author{Ze Wang \\
Purdue University \\
\texttt{{\small zewang@purdue.edu}} \\
\and
Jiang Wang $\quad$ Zicheng Liu\\
Microsoft Corporation \\
\texttt{{\small\{jiangwang, zliu\}@microsoft.com}} \\
\and
Qiang Qiu \\
Purdue University \\
\texttt{{\small qqiu@purdue.edu}} \\
}
\maketitle

\begin{abstract}

In this paper, we show that a binary latent space can be explored for compact yet expressive image representations. We model the bi-directional mappings between an image and the corresponding latent binary representation by training an auto-encoder with a Bernoulli encoding distribution. On the one hand, the binary latent space provides a compact discrete image representation of which the distribution can be modeled more efficiently than pixels or continuous latent representations. On the other hand, we now represent each image patch as a binary vector instead of an index of a learned cookbook as in discrete image representations with vector quantization. In this way, we obtain binary latent representations that allow for better image quality and high-resolution image representations without any multi-stage hierarchy in the latent space. In this binary latent space, images can now be generated effectively using a binary latent diffusion model tailored specifically for modeling the prior over the binary image representations. We present both conditional and unconditional image generation experiments with multiple datasets, and show that the proposed method performs comparably to state-of-the-art methods while dramatically improving the sampling efficiency to as few as 16 steps without using any test-time acceleration. The proposed framework can also be seamlessly scaled to $1024 \times 1024$ high-resolution image generation without resorting to latent hierarchy or multi-stage refinements. 

\end{abstract}


\begin{figure}[t]
    \centering
    \includegraphics[width=\linewidth]{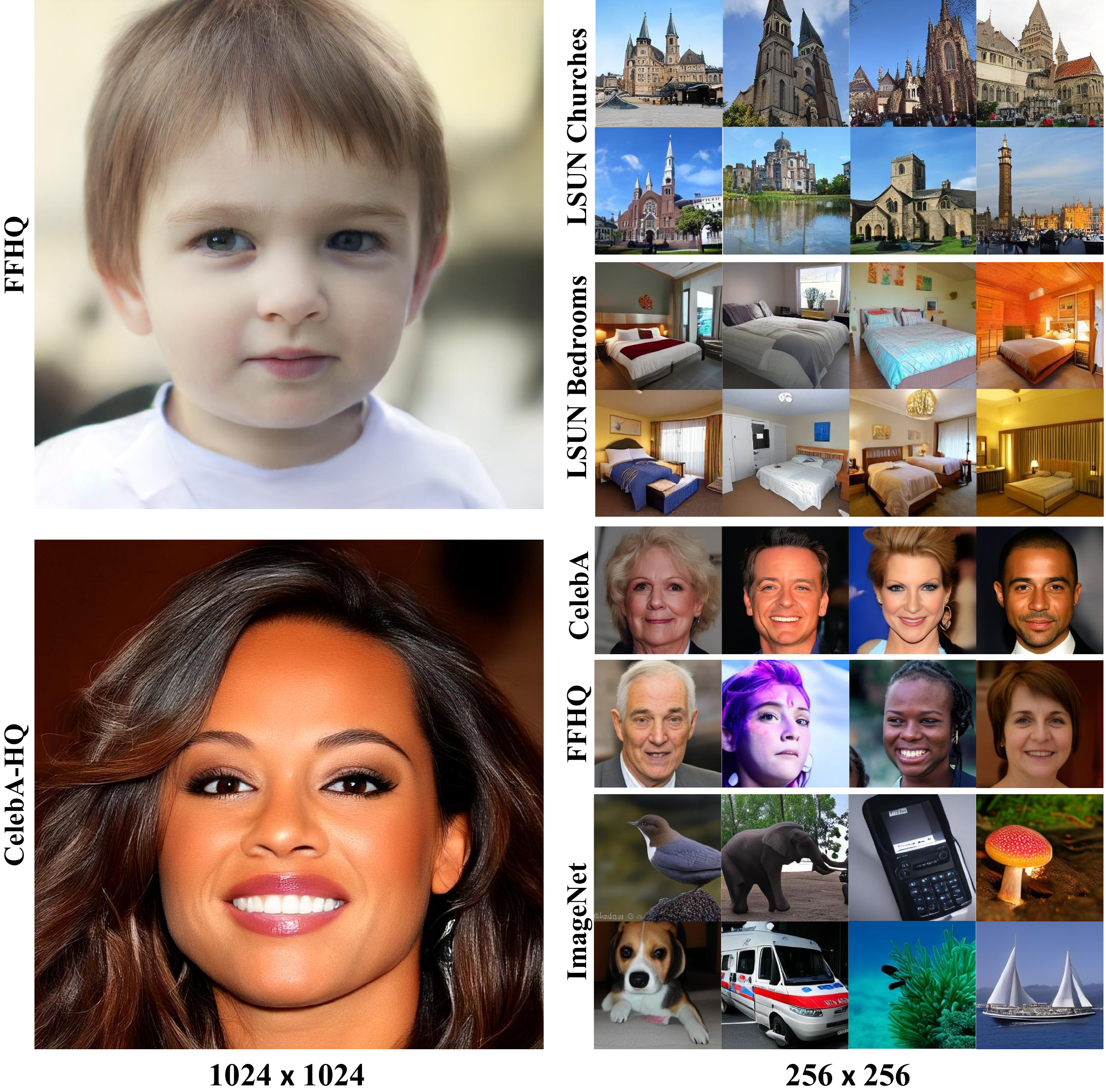}
    \caption{Examples of generated images with different resolutions using the proposed method.}
    \label{fig:show}
\end{figure}

\section{Introduction}
The goal of modeling the image distribution that allows the efficient generation of high-quality novel samples drives the research of representation learning and generative models. 
Directly representing and generating images in the pixel space stimulates various research such as generative adversarial networks \cite{gan,wgan,biggan,pggan}, flow models \cite{flows,kobyzev2020normalizing,nice,papamakarios2021normalizing}, energy-based models \cite{xieebm,implicit,improved,lecunebm}, and diffusion models \cite{diffusion,ddpm,ddim,improved_ddpm}.
As the resolution grows, it becomes increasingly difficult to accurately regress the pixel values. And this challenge usually has to be addressed through hierarchical model architectures \cite{pggan,c2f} or at a notably high cost \cite{ddpm}. Moreover, while demonstrating outstanding generated image quality, GAN models suffer from issues including insufficient mode coverage \cite{metz2016unrolled} and training instability \cite{gulrajani2017improved}.

Representing and generating images in a learned latent space \cite{latent,vae,latentebm} provides a promising alternative.
Latent diffusion \cite{latent} performs denoising in latent feature space with a lower dimension than the pixel space, therefore reducing the cost of each denoising step. However, regressing the real-value latent representations remains complex and demands hundreds of diffusion steps. 
Variational auto-encoders (VAEs) \cite{vae,betavae,rezende2014stochastic} generate images without any iterative steps. However, the static prior of the latent space restricts the expressiveness, and can lead to posterior collapse.
To achieve higher flexibility of the latent distribution without significantly increasing the modeling complexity, VQ-VAE \cite{vqvae} introduces a vector-quantized latent space, where each image is represented as a sequence of indexes, each of which points to a vector in a learned codebook. The prior over the vector-quantized representations is then modeled by a trained sampler, which is usually parametrized as an autoregressive model. The success of VQ-VAE stimulates a series of works that model the discrete latent space of codebook indexes with different models such as accelerated parallel autoregressive models \cite{maskgit} and multinomial diffusion models \cite{unleashing,vqdiff}.
VQ-based generative models demonstrate surprising image synthesis performance and model coverage that is better than the more sophisticated methods like GANs without suffering from issues like training instability. 
However, the hard restriction of using one codebook index to represent each image patch introduces a trade-off on the codebook size, as a large enough codebook to cover more image patterns will introduce an over-complex multinomial latent distribution for the sampler to model.

In this research, we explore a compact yet expressive representation of images in a binary latent space, where each image patch is now represented as a binary vector, and the prior over the discrete binary latent codes is effectively modeled by our improved binary diffusion model tailored for Bernoulli distribution. 
Specifically, the bi-directional mappings between images and the binary representations are modeled by a feed-forward autoencoder with a binary latent space. 
Given an image, the encoder now outputs the normalized parameters of a sequence of independently distributed Bernoulli variables, from which a binary representation of this image is sampled, and fed into the decoder to reconstruct the image. 
The discrete sampling in Bernoulli distribution does not naturally permit gradient propagation. We find that a simple straight-through gradient copy \cite{estimating,latentb} is sufficient for high-quality image reconstruction while maintaining high training efficiency. 

With images compactly represented in the binary latent space, we then introduce how to generate novel samples by modeling the prior over binary latent codes of images. 
To overcome the shortcomings of many existing generative models such as being uni-directional \cite{vqvae,vqvae2} and the non-regrettable greedy sampling \cite{maskgit,unleashing}, we introduce binary latent diffusion that generates the binary representations of novel samples by a sequence of denoising starting from a random Bernoulli distribution. 
Performing diffusion in a binary latent space, modeled as Bernoulli distribution, reduces the need for precisely regressing the target values as in Gaussian-based diffusion processes \cite{ddpm,ddim,latent}, and permits sampling at a higher efficiency.
We then introduce how to progressively reparametrize the prediction targets at each denoising step as the residual between the inputs and the desired samples, and train the proposed binary latent diffusion models to predict such `flipping probability' for improved training and sampling stability. 


We support our findings with both conditional and unconditional image generation experiments on multiple datasets. We show that our method can deliver remarkable image generation quality and diversity with more compact latent codes, larger image-to-latent resolution ratios, as well as fewer sampling steps, and faster sampling speed.
We present some examples with different resolutions generated by the proposed method in Figure~\ref{fig:show}. 

We organize this paper as follows:
Related works are discussed in Section~\ref{related}.
In Section~\ref{bae}, we introduce binary image representations by training an auto-encoder with a binary latent space. 
We then introduce in Section~\ref{bdiffusion} binary latent diffusion, a diffusion model for multi-variate Bernoulli distribution, and techniques tailored specifically for improving the training and sampling of binary latent diffusion. We present both quantitative and qualitative experimental results in Section~\ref{exp} and conclude the paper in Section~\ref{conclusion}.

\section{Related Work}
\label{related}

\paragraph{Diffusion models and image generation.}

Diffusion models \cite{diffusion} are proposed as a flexible way of modeling complex data distribution using tractable families of probability distributions. 
Denoising diffusion probabilistic models (DDPM) \cite{ddpm} build a reparameterization of the learning objective that connects diffusion probabilistic models and denoising score matching \cite{score}. Gaussian diffusion models scale up the resolution and achieve image generation with quality comparable with strong generative families such as GANs \cite{stylegan2,sngan,biggan}, score-based models \cite{score,scorev2,song2020score}, and energy-based models \cite{clsy,ebm}, without suffering from issues like mode collapse and training instability. 
Further improvements have been proposed to encourage higher log-likelihoods \cite{improved_ddpm}, reduce computational demands \cite{latent}, and extend diffusion models to broader applications including text-to-image generation \cite{dalle2,imagen}, planning \cite{janner2022planning}, super-resolution \cite{superres}, temporal data modeling \cite{alcaraz2022diffusion,kong2020diffwave,tashiro2021csdi}, and adversarial robustness \cite{blau2022threat,wang2022guided}.
Particularly, diffusion models with discrete states are advocated in \cite{structured} for text generation. A series of structured transition matrices are introduced in \cite{structured}, among which diffusion with an absorbing state is further extended to image generation with VQ codes in \cite{unleashing}. \cite{meng2022concrete} introduces a novel score function that generalizes score-based models to discrete representations.

\noindent \textbf{Discrete representations.}
Modern deep neural networks trained with back-propagation and gradient descent prevalently advocate continuous features across all layers.
However, discrete representations still demonstrate their unique advantages and applications.
\cite{estimating} discusses several approaches to estimating the gradient through stochastic neurons. Among the four approaches introduced in \cite{estimating}, the practically simplest one with heuristic gradient copies is further adopted in learning fully discrete vector-quantized auto-encoders \cite{vqvae}, which builds the foundations for many follow-up methods \cite{unleashing,maskgit,vqgan,vqvae2}. 
\cite{deja2021binplay} proposes rehearing training samples for continual learning in a binary latent space. 
\cite{latentb} introduces a binary latent space by a hard threshold instead of sampling in our methods. And the sampling with random hyperplane rounding in \cite{latent} can hardly be scaled to a higher resolution.

\newcommand{\img}{\textbf{x}}
\newcommand{\latent}{\textbf{y}}
\newcommand{\blatent}{\textbf{z}}
\newcommand{\encoder}{\boldsymbol{\Psi}}
\newcommand{\decoder}{\boldsymbol{\Phi}}
\newcommand{\bnl}{\mathcal{B}}
\newcommand{\R}{\mathbb{R}}


\begin{figure*}[t]
    \centering
	\includegraphics[width=\linewidth]{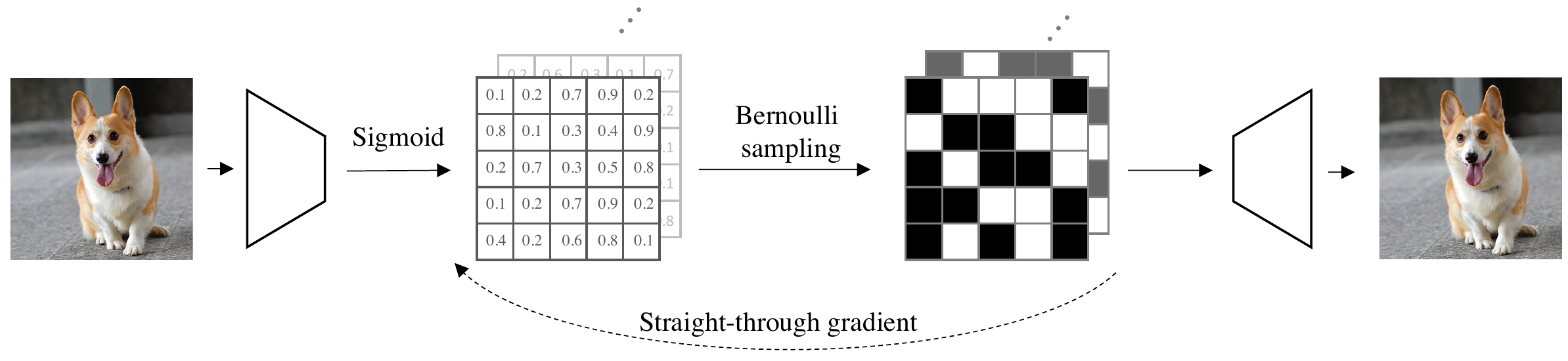}
\caption{Illustration of the proposed binary auto-encoder. The gradient flow estimated with a straight-through surrogate function is denoted as the dashed line.}
\label{fig:bae}
\end{figure*}

\section{Binary Image Representations}
\label{bae}

Given an image dataset, we begin with learning the bi-directional mappings between images and their binary representations.
This is achieved by training an auto-encoder with a binary latent space, where the binary code of each image is obtained as a sample of a sequence of independently distributed Bernoulli variables inferred from the image.
Specifically, denoting an image as $\img \in \R^{h \times w \times 3}$, 
we train an image encoder $\encoder$ that outputs the unnormalized parameters for the corresponding Bernoulli distribution $\encoder(\img)$. A Sigmoid non-linearity $\sigma$ is then adopted to normalize the parameters $\latent \in \R^{\frac{h}{k} \times \frac{w}{k} \times c} = \sigma(\encoder(\img))$, where $h$ and $w$ denote the spatial resolution of the image, $k$ is the downsampling factor of the encoder $\encoder$, and $c$ is the number of encoded feature channels. 

To obtain the binary representations of images, we perform Bernoulli sampling given the normalized parameters $\blatent = \text{Bernoulli}(\latent)$. Note that the Bernoulli sampling operation here does not naturally permit gradient propagation through it, and prevents the end-to-end training of the encoder-decoder architecture. 
In practice, we consistently observe that a straight-through gradient estimation by direct copying the gradients and skipping the non-differentiable sampling in backpropagation as in \cite{gumbel} can maintain both stable training and superior performance. 
The straight-through gradient estimation can be easily implemented by a surrogate function as:
\begin{equation}
\label{eq:binarize}
\small
    \tilde{\blatent} = \oslash(\blatent) + \latent - \oslash(\latent),
\end{equation}
where $\oslash(\cdot)$ denotes the stop gradient operation. $\tilde{\blatent}$, which is binary and identical to $\blatent$, will be sent to the following decoder network $\decoder$ for image reconstruction. The gradient propagated back from the decoder $\decoder$ to $\tilde{\blatent}$ is directly sent to $\latent$, which is differentiable w.r.t. the encoder $\encoder$ and permits the end-to-end training of the entire auto-encoder with a discrete binary latent space. 

The reconstruction of the image is obtained using a decoder network $\decoder$ as $\hat{\img} = \decoder(\tilde{\blatent})$. The overall framework of the binary autoencoder is visualized in Figure~\ref{fig:bae}.
With the gradient surrogate function ensuring the end-to-end gradient propagation, the network is trained by minimizing the final objective
\begin{equation}
\label{eq:train_bae}
\small
    \mathcal{L} = \sum_{i}^{||\mathcal{C}||} \omega_i \mathcal{C}[i] (\hat{\img}, \img),
\end{equation}
where $\mathcal{C}$ denotes a collection of loss functions such as mean squared error, perception loss, and adversarial loss. And $\omega_i$ denotes the weights that balance each loss term.

\begin{figure}[b]
    \centering
	\includegraphics[width=\linewidth]{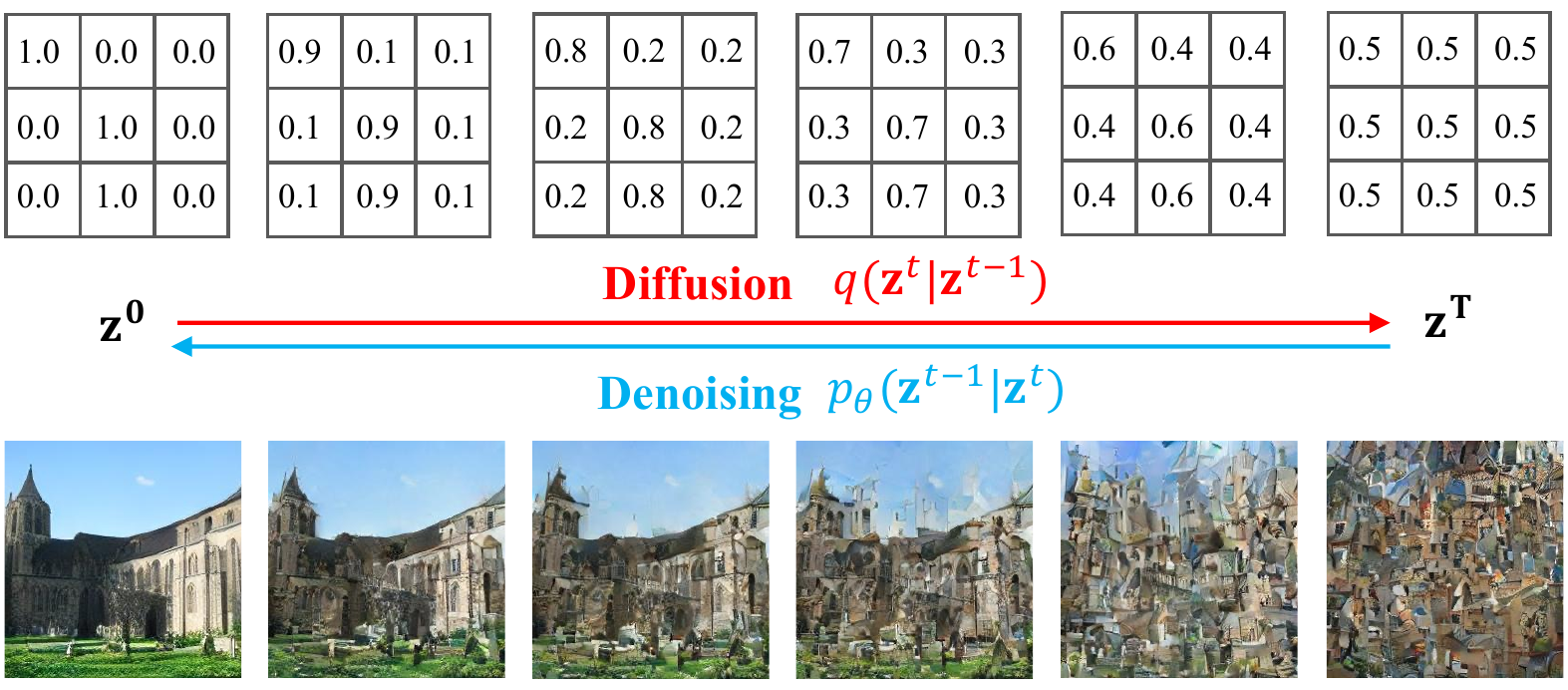}
\caption{Illustration of the binary latent diffusion and denoising processes. A binary representation is progressively diffused towards a random Bernoulli distribution with 0.5 everywhere. }
\label{fig:diffusion}
\end{figure}

\section{Bernoulli Diffusion Process}
\label{bdiffusion}
Given an image dataset and the corresponding binary latent representations of each image, we then discuss how to effectively model the prior over the binary latent codes with a parametrized model $p_\theta(\blatent)$, from which novel samples of the binary latent codes can be efficiently sampled. 
To do so, we introduce binary latent diffusion, a diffusion model tailored specifically for Bernoulli distribution accompanied by improvement techniques that promote stable and effective training and sampling.

A diffusion model is usually established by first defining a $T$-step diffusion process consisting of a sequence of variation distributions $q(\blatent^t | \blatent^{t-1})$, with $t \in \{1, \dots T\}$.
Each variation distribution in $q(\blatent^t | \blatent^{t-1})$ is defined to progressively add noise to $\blatent^{t-1}$, so that with sufficient steps $T$ and a valid noise scheduler defining $q(\blatent^t | \blatent^{t-1})$, the final state $q(\blatent^T | \blatent^{0})$ converges to a known random distribution that is easy to evaluate and sample from, and conveys almost no valid information of $\blatent^0$. 
Specifically, in our case, we are interested in modeling the prior distribution of the binary latent code of images. 
Therefore, we define the starting point of the diffusion process with our latent code distribution $\blatent \sim q(\blatent^0)$, where $q(\blatent^0)$ is a Bernoulli distribution as the prior of latent codes. 
And the full diffusion process $q$ can be defined as 
\begin{equation}
\small
    q(\blatent^{1:T}) := \prod_{t=1}^{T} q(\blatent^t | \blatent^{t-1}), \quad \text{with}
\end{equation}
\begin{equation}
\small
    q(\blatent^t| \blatent^{t-1}) = \bnl(\blatent^t; \blatent^{t-1}(1 - \beta^t) + 0.5 \beta^t),
\end{equation}
where $\bnl$ denotes a Bernoulli distribution and $\beta^t$ defines the noise scale at each step $t$. 
With sufficient steps $T$, and $\beta^t$ at adequate scales, the forward diffusion process will converge to $\bnl(\blatent^T; 0.5)$, which is a random Bernoulli distribution that is easy to sample from. Figure~\ref{fig:diffusion} illustrates an example of the binary latent diffusion process.

Given an arbitrary time step $t$ and a sample $\blatent^0$, the posterior distribution can be easily obtained as 
\begin{equation}
\label{eq:q_at_anytime}
\small
\begin{aligned}
     & q(\blatent^t | \blatent^0, \blatent^T) = \bnl(\blatent^t; k^t \blatent^0 + b^t), \quad \text{with} \\
      & k^t =  \prod_{i=1}^{t} (1 - \beta^i), \\
      & b^t = (1 - \beta^t)b^{t-1} + 0.5 \beta^t, \quad \text{and} \quad b^1 = 0.5 \beta^1,
\end{aligned}
\end{equation}
where $k^t$ and $b^t$ here jointly define the accumulated noise scale till time step $t$. The Markov chain and the analytic form of the posterior in (\ref{eq:q_at_anytime}) permit stochastic sampling in each training batch.

The noise schedulers of the proposed binary latent diffusion process can be constructed by either simply defining the noise scale $\beta^t$ at each step, or directly defining the accumulated noise factors $k^t$ and $b^t$. 
Note that even with $k^t$ and $b^t$ directly defined, the corresponding $\beta^t$ can still be obtained as $\beta^t = 1 - \frac{k^t}{k^{t-1}}$. We present in Appendix Section~\ref{noise} discussions on different choices of noise schedulers.

With the forward diffusion process properly defined. the goal now is to train a function $f_\theta$ with $\hat{\blatent}^{t-1} = f_\theta(\blatent^t, t)$ to model the reverse diffusion (denoising) process 
\begin{equation}
\small
    p_\theta(\blatent^{t-1} | \blatent^t) = \bnl({\blatent^{t-1}; f_\theta(\blatent^t, t)}), 
\end{equation}
which allows sampling to be performed by reversing a sample from $q(\blatent^T)$ to a sample in $q_\theta(\blatent^0)$.
The diffusion process $q$ and the denoising process $p$ jointly define a variational auto-encoder \cite{vae}, with the variational lower bound:
\begin{equation}
\small
\begin{aligned}
\label{eq:vlb}
    \mathcal{L}_{\text{vlb}} &:= \mathcal{L}_0 + \sum_{t=1}^{T-1} \mathcal{L}_t + \mathcal{L}_T \\
                             &:= -\log p_\theta(\blatent^0 | \blatent^1) \\
                             &+ \sum_{t=1}^{T-1} \text{KL}(q(\blatent^{t-1}|\blatent^t, \blatent^0) || p_\theta(\blatent^{t-1}|\blatent^t)) \\
                             &+\text{KL}(q(\blatent^T|\blatent^0)|| p(\blatent^T)). 
\end{aligned}
\end{equation}
Note that in this paper we focus on the diffusion process with fixed predefined noise scheduler $\beta^t$, therefore the third term on the RHS of (\ref{eq:vlb}) does not depend on $\theta$ and is always nearly $0$.
All the KL divergence and likelihood terms in (\ref{eq:vlb}) can be analytically calculated in a closed form as all distributions involved are Bernoulli.

\subsection{Binary Latent Diffusion Reparameterization}
To learn the reverse diffusion process, a straightforward way is to train a neural network $f_\theta$ using (\ref{eq:vlb}) to model $p_\theta(\blatent^{t-1}|\blatent^t)$.
According to (\ref{eq:q_at_anytime}), each $\blatent^t$ can be considered as a \textit{linear interpolation} between $\blatent^0$ and $\blatent^T$, whose parameters can take arbitrary numbers between 0 and 1 depending on the noise scheduler. Therefore,
directly training $f_\theta$ to model $p_\theta(\blatent^{t-1}|\blatent^t)$ can be challenging as the model needs to accurately regress such sophisticated interpolations. 

\noindent \textbf{Predicting $\blatent^0$.}
Inspired by \cite{vqdiff}, 
we first introduce a reparameterization to the prediction target as $p_\theta(\blatent^0|\blatent^t)$ and train the model $f_\theta$ to directly predict $\blatent^0$ as $\hat{\blatent}^0 = f_\theta(\blatent^t, t)$ at each step.
During sampling, at each step, with $p_\theta(\blatent^0|\blatent^t)$ predicted, we can recover the corresponding 
$p_\theta(\blatent^{t-1} | \blatent^t)$ by:
\begin{equation}
\label{eq:one_step}
\small
\begin{aligned}
    p_\theta(\blatent^{t-1} | \blatent^{t}) = & q(\blatent^{t-1} | \blatent^t, \blatent^0 = \textbf{0}) p_\theta(\blatent^0 = \textbf{0} | \blatent^t) \\& + q(\blatent^{t-1} | \blatent^t, \blatent^0 = \textbf{1}) p_\theta(\blatent^0 = \textbf{1} | \blatent^t), 
\end{aligned}
\end{equation}
with 
\begin{equation}
\small
    q(\blatent^{t-1} | \blatent^{t}, \blatent^0) = \frac{q(\blatent^t | \blatent^{t-1}, \blatent^0) q(\blatent^{t-1}| \blatent^0)}{q(\blatent^t | \blatent^0)}.
\end{equation}
Specifically, with a binary code $\blatent^t$ and the noise scheduler defined in (\ref{eq:q_at_anytime}), we have 
\begin{equation}
\small
\begin{aligned}
\label{eq:rep0}
    &p_\theta(\blatent^{t-1}|\blatent^t) \\
    &= \bnl \big(\blatent^{t-1} | \frac{[(1-\beta^t)\blatent^t + 0.5\beta^t] \odot [k^t f_\theta(\blatent^t, t) + 0.5b^t]}{\mathbf{Z}} \big),
\end{aligned}
\end{equation}
where 
\begin{equation}
\small
\begin{aligned}
    \mathbf{Z} &= [(1-\beta^t)\blatent^t + 0.5\beta^t] \odot [k^t f_\theta(\blatent^t, t) + 0.5b^t] \\
    &+ [(1-\beta^t)(1-\blatent^t) + 0.5\beta^t] \odot [k^t (1-f_\theta(\blatent^t, t)) + 0.5b^t]
\end{aligned}
\end{equation}
is a normalization term that guarantees valid probabilities in (\ref{eq:rep0}), and $\odot$ denotes element-wise product.
With the introduced reparameterization, the prediction targets at all time steps become $\blatent^0$, which is strictly binary, and eases the training according to our observations. 

\noindent \textbf{Predicting the residual.}
A reparameterization to the prediction target as the residual is introduced in \cite{ddpm}, which bridges the connections between diffusion and denoising score matching \cite{score} and improves the sampling results. 
In the proposed binary latent diffusion, the decreasing noise scales result in $\blatent^t$ getting closer to $\blatent^0$ as $t$ decreases. 
Thus, the sampling in the binary latent space naturally favors fewer flipping to the binary codes as $t$ gets closer to $t = 0$. To better capture this \textit{sparsity} in predictions and stabilize the sampling to prevent divergence, we show that, while predicting the residual is considered non-trivial for discrete-state diffusion models \cite{multinomial}, the prediction targets can be further reparametrized as the residual between $\blatent^0$ and $\blatent^t$ in our binary latent diffusion. Specifically, we train $f_\theta(\blatent^t, t)$ to fit $\blatent^t \oplus \blatent^0$, where $\oplus$ denotes the element-wise logic XOR operation. The model $f_\theta$ is now trained to predict the `flipping probability’ of the binary code. And the prediction targets remain strictly binary.


\noindent \textbf{Final training objective.}
The final simplified training objective can be formulated as:
\begin{equation}
\small
    \mathcal{L}_{\text{residual}} = \mathbb{E}_{t, \blatent^0}\text{BCE}(f_\theta(\blatent^t, t), \blatent^t \oplus \blatent^0),
\end{equation}
where $\text{BCE}(\cdot, \cdot)$ denotes the binary cross-entropy loss.
In practice, following \cite{improved_ddpm}, we find it beneficial to set the final learning objective as a combination of $\mathcal{L}_{\text{residual}}$ and $\mathcal{L}_{\text{vlb}}$ as 
\begin{equation}
\small
\label{eq:final}
    \mathcal{L} = \mathcal{L}_{\text{residual}} + \lambda\mathcal{L}_{\text{vlb}},
\end{equation}
with $\lambda$ is a small number.

\begin{figure}[t]
    \centering
	\includegraphics[width=\linewidth]{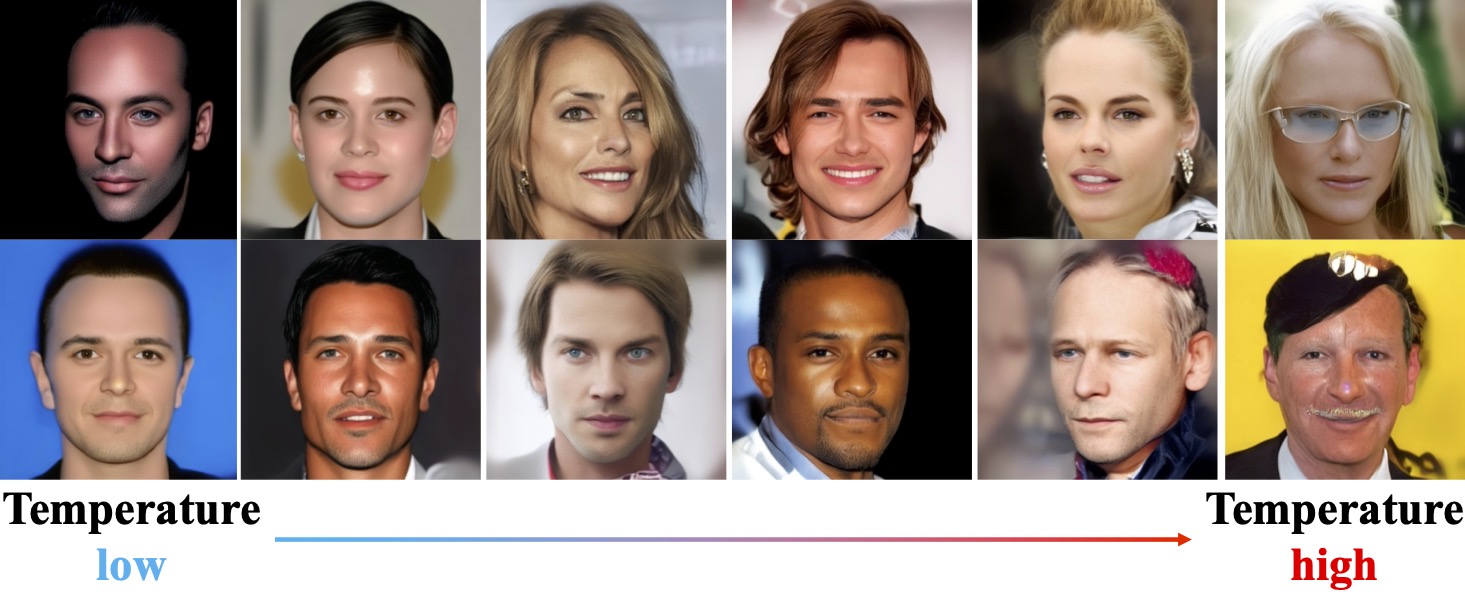}
    \caption{Temperature values directly decide the diversity of the sampled images. High temperatures improve coverage and reveal rare patterns such as reflections. But overly high temperatures can result in sampling procedures failing to converge and produce artifacts. Demonstrative results are obtained on the $1024 \times 1024$ CelebA-HQ experiment with temperature values from 0.2 to 1.2.}
\label{fig:temp}
\end{figure}

\noindent \textbf{Sampling temperatures.}
In practice, the prediction of the residual $f_\theta(\blatent^t, t)$ is implemented as $f_\theta(\blatent^t, t) = \sigma(\mathcal{T}_\theta(\blatent^t, t))$, where $\mathcal{T}_\theta$ is a plain transformer that outputs the unnormalized flipping probability, which is then normalized by a Sigmoid function $\sigma$. During sampling, we can manually adjust the sampling diversity by inserting a temperature $\tau$ which turns the prediction to $f_\theta(\blatent^t, t) = \sigma(\mathcal{T}_\theta(\blatent^t, t) / \tau)$. Note that $\tau$ is only used to adjust the diversity during post-training sampling, and is not included as a hyperparameter in training. 
Examples of how $\tau$ affects the sample diversity are shown in Figure~\ref{fig:temp}.

\subsection{Comparisons with Existing Methods}
\label{comp}

\begin{figure}[t]
    \centering
	\includegraphics[width=\linewidth]{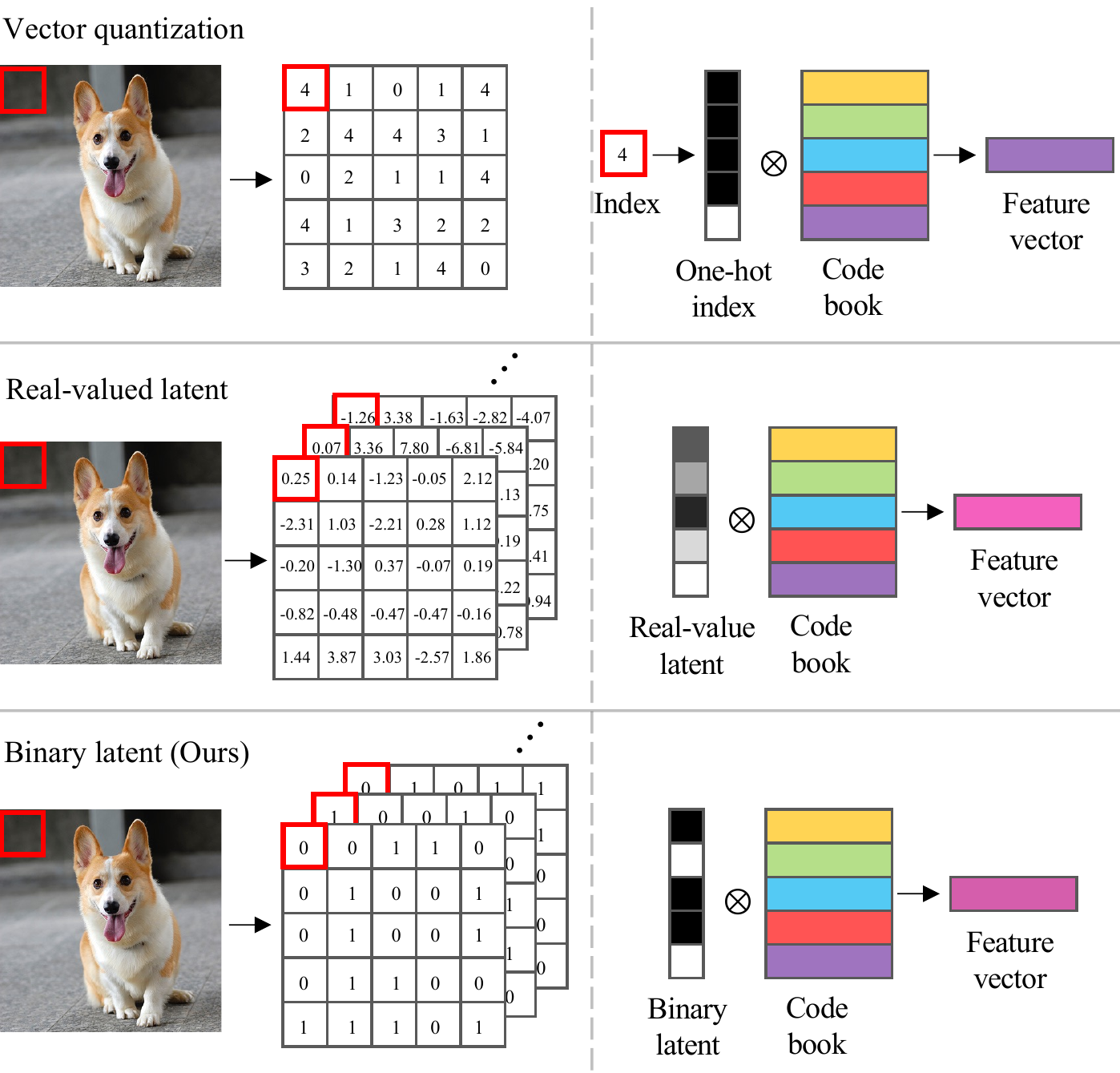}
\caption{Interpreting different latent representations from a code-book view. Vector-quantized (top) methods perform one-hot selections of codes. Latent diffusion composes codes with arbitrary linear combinations. Our method also permits the composition of codes but restricts the composition to be binary for improved efficiency. }
\label{fig:compare}
\end{figure}

In this section, we briefly discuss the advantages of adopting the proposed binary representations over other alternatives of latent space image representations. 
As visualized in Figure~\ref{fig:compare}, vector-quantized latent space \cite{vqvae,vqgan} represents each image patch as a discrete index, or equivalently, a one-hot vector. The one-hot vector then multiplies with the learned codebook to obtain the feature representation of an image patch. 
Image representations in continuous latent space \cite{latent} can be interpreted in a similar way by simply treating the first weight matrix in the decoder network as the learned codebook.
The real-value latent code of an image patch performs arbitrary linear combinations of vectors in the codebook and is able to cover diverse feature space even with a low-dimensional code for higher efficiency. 
Our method with binary representation strives for a balance between these two methods by restricting the vectors composing the codebook to be binary. 
On the one hand, the binary composition of a codebook offers a much more diverse and flexible composition of features compared to the vector-quantized representations. For example, a compact 32-bit binary vector can represent over 4,000 million patterns, which is much larger than the 1,024 patterns commonly used in vector-quantized representations \cite{maskgit,unleashing,vae}. As we will further show empirically in Section~\ref{exp}, this improved coverage of patterns allows for higher expressiveness and enables high-resolution image generation that can hardly be accomplished by VQ representations at high quality.
On the other hand, the binary restriction guarantees that the representations remain compact. As we will show in Section~\ref{discuss}, our 8k-bit binary representations perform comparably with a latent diffusion model \cite{latent} with 131k-bit representations. 

\section{Experiments}
\label{exp}
In this section, we present both unconditional and conditional image generation experiments with multiple datasets.
Following common practice \cite{maskgit,unleashing}, we train a plain transformer network \cite{atten} to parametrize the sampler $\mathcal{T}_\theta$. We use a temperature of $\tau = 0.9$ as the default setting for sampling, and $\lambda = 0.1$ in all the experiments, if not otherwise specified. 
For the number of channels in the binary latent space, we use $c = 32$ for $256 \times 256$ unconditional image generation, $c = 64$ for class-conditional image generation, and $1024 \times 1024$ high-resolution unconditional image generation.
Additional details regarding the implementation and evaluation metrics can be found in Appendix Section~\ref{details}.







\subsection{Unconditional Image Generation}
\label{exp_uncond}
In this section, we present results and comparisons on unconditional image generation experiments with multiple datasets including LSUN Bedrooms and Churches \cite{lsun}, FFHQ \cite{stylegan}, and CelebA-HQ \cite{pggan}. 
To allow for fair comparisons, we first present $256\times256$ image generation. We apply our method on a $16\times16\times32$ latent space, which corresponds to the same $16\times$ downsampling of the spatial resolution as in methods like \cite{unleashing,maskgit}. Note that as we further show in Section~\ref{discuss}, the improved expressiveness of our binary representation compared to the vector-quantized representation allows our method to perform well with a larger downsampling ratio such as $32\times$.
Each image patch is now represented with a 32-bit binary vector, and an image is represented as an 8k-bit tensor, which is more compact compared to the 131k-bit representations in Latent Diffusion models \cite{latent}.

For comprehensive evaluations and comparisons, we adopt Fr\'echet Inception Distance (FID) and precision-recall \cite{pr} as the evaluation metrics.
We present comparisons against various state-of-the-art methods in Table~\ref{tab:uncond}.
We conduct comparisons by generating 50,000 samples and comparing them with the corresponding training dataset in every experiment. Quantitative comparisons against state-of-the-art methods are presented in Table~\ref{tab:uncond}. Despite using much fewer denoising steps compared to latent diffusion models (LDM) \cite{latent} and absorbing diffusion models \cite{unleashing}, our binary latent diffusion models achieve comparable image generation quality with desirable diversity.
Note that our results are obtained without any test-time accelerations or skipping denoising steps. While LDM \cite{latent} results are reported with accelerated 200 DDIM \cite{ddim} steps at test time, our method still demonstrates significantly higher sampling efficiency as we will show in Section~\ref{discuss}.
We present further qualitative results in Appendix Section~\ref{uncond_imgs}.

\begin{table}[]
    \centering
    \caption{Quantitative comparisons on unconditional image generation with LSUN Bedrooms, LSUN Churches, FFHQ, and CelebA-HQ. We also provide comparisons on the number of denoising steps in \textit{training}. All results are obtained with a resolution of $256\times256$.}
    \label{tab:uncond}
    
    \resizebox{\linewidth}{!}{
    \begin{tabular}{c|c| ccc | ccc }
    \toprule
         \multirow{2}{*}{Methods} & \multirow{2}{*}{Steps} &     \multicolumn{3}{c|}{Bedrooms} & \multicolumn{3}{c}{Churches} \\
         & & FID $\downarrow$ & P $\uparrow$ & R $\uparrow$ & FID $\downarrow$ & P $\uparrow$ & R $\uparrow$ \\
    \midrule
    DCT \cite{dct} & - & 6.40 & 0.44. & 0.56 & 7.56 & 0.60 & 0.48 \\
    ImageBART \cite{imagebart} & & 5.51 & - & - & 7.32 & - & -\\
    VQGAN \cite{vqgan} & 256 & 6.35 & 0.61 & 0.33 & 7.81 & 0.67 & 0.29\\
    DDPM \cite{ddpm} & 1,000 & 6.36 & - & - & 7.89 & - & -  \\
    PGGAN \cite{pggan} & - & 8.34 & 0.43 & 0.40 & 6.42 & 0.61 & 0.38 \\
    StyleGAN \cite{stylegan,stylegan2} & - & 2.35 & 0.55 & 0.48 & 3.86 & 0.60 & 0.43 \\
    Absorbing \cite{unleashing} & 256 & 3.64 & 0.67 & 0.38 & 4.07 & 0.71 & 0.45\\
    LDM \cite{latent} & 1,000 & 2.95 & 0.66 & 0.48 & 4.02 & 0.64 & 0.52\\
    \textbf{Ours} & 16 & 4.32 & 0.62 & 0.44 & 4.96 & 0.66 & 0.48\\
    \textbf{Ours} & 64 & 3.85 & 0.65 & 0.44 & 4.36 & 0.68 & 0.50 \\

    \midrule
    \midrule
    
    \multirow{2}{*}{Methods} & \multirow{2}{*}{Steps} &     \multicolumn{3}{c|}{FFHQ} & \multicolumn{3}{c}{CelebA-HQ} \\
         & & FID $\downarrow$ & P $\uparrow$ & R $\uparrow$ & FID $\downarrow$ & P $\uparrow$ & R $\uparrow$ \\
    \midrule
    VQGAN & 400 & - & - & - & 10.2 & - & - \\
    StyleGAN \cite{stylegan} & - & 4.16 & 0.71 & 0.46 & - & - & -\\
    UDM \cite{udm} & 1,000 & 5.52 & - & - & 7.16 & - & - \\
    Absorbing \cite{unleashing} & 256 & 6.11 & 0.73 & 0.48 & - & - & -\\
    LDM \cite{latent} & 1,000 & 4.98 & 0.73 & 0.50 & 5.11 & 0.72 & 0.49 \\
    \textbf{Ours} & 16 & 6.48 & 0.71 & 0.45 & 7.80 & 0.67 & 0.44\\
    \textbf{Ours} & 64 & 5.85 & 0.73 & 0.50 & 6.24 & 0.71 & 0.48 \\
         
    \midrule
    \end{tabular}}
\end{table}

    
      

\begin{table*}[t]
    \resizebox{\linewidth}{!}{%
	\begin{minipage}{0.70\linewidth}
    \centering
    \small
    \caption{Quantitative comparisons on $256 \times 256$ class-conditional image generation. We report both the top-1 and top-5 accuracy of CAS. Using the official training set obtains Top-1 and Top-5 accuracy of $76.6\%$ and $93.1\%$, respectively.}
    \label{tab:imagenet}
    
    \resizebox{\linewidth}{!}{
    \begin{tabular}{ c | cccccccc}
    \toprule
        Methods & Params & Step & FID $\downarrow$ & IS $\uparrow$ & P $\uparrow$ & R $\uparrow$  & Top-1 $\uparrow$  & Top-5 $\uparrow$\\
    \midrule
        DCT \cite{dct} &  738M & $>$1024 & 36.51 & - & 0.36 & 0.67 & - & -\\ 
        BigGAN-deep \cite{biggan} & 160M & 1 & 6.95 & 198.2 & 0.87 &  0.28 & 43.99 & 67.89\\
        Improved DDPM \cite{improved_ddpm} & 280M & 250 & 12.26 & - & 0.70 & 0.62 & - & - \\
        ADM \cite{guidance} & 554M & 250 & 10.94 & 101.0 & 0.69 & 0.63 & - & -\\
        VQVAE2 \cite{vqvae2} & 13.5B & 5120 & 31.11 & $\sim$ 45 & 0.36 & 0.57 & 54.83 & 77.59\\
        VQGAN \cite{vqgan} & 1.4B & 256 & 15.78 & 78.3 & - & - & - & - \\
        MaskGIT \cite{maskgit} & 227M & 8 & 6.18& 182.1 & 0.80 & 0.51 & 63.14 & 84.45 \\
        LDM \cite{latent} & 400M & 250 &  10.56 & 103.49 & 0.71 & 0.62 & - & -\\
      
        \textbf{Ours} & 172M & 64 & 8.21 & 162.32 & 0.72 & 0.64 & 65.07 & 85.60 \\
    \bottomrule
    \end{tabular}}
    \end{minipage}
    \hspace{2mm}
    \begin{minipage}{0.26\linewidth}
    \small
    \includegraphics[width=\linewidth]{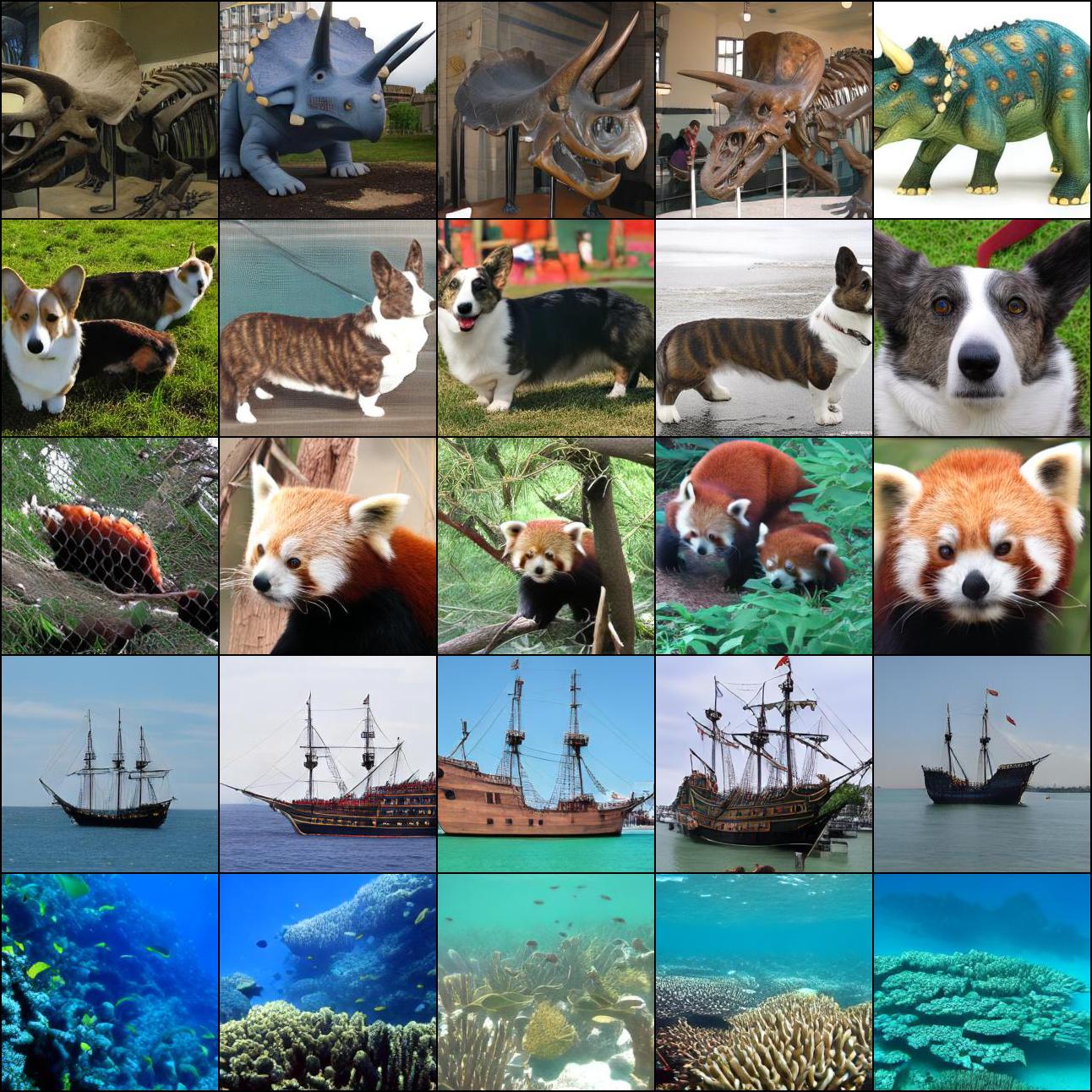}
    \captionof{figure}{Class-conditional image generation with classes: 51, 264, 387, 724, 973.}
    \label{fig:imagenet}
    \end{minipage}
    }
\end{table*}

\noindent \textbf{High-resolution image generation in one shot.} 
Compared to vector-quantization based methods, the proposed binary latent space allows each image patch to be represented as a composition of features, and therefore potentially permits latent space image generation with a large downsampling (image-to-latent resolution) ratio. 
And the larger resolution ratio permits image generation with larger image resolution without increasing the difficulty of modeling the latent prior. 
To show this, we directly present high-resolution image generation experiments with FFHQ and CelebA-HQ by increasing the target image resolution to $1024\times1024$. Generating such high-resolution images in a discrete latent space used to be handled by a multi-stage hierarchy of the latent representations \cite{vqvae2,yan2021videogpt,yu2021vector}. We show that in our method, high-resolution image generation can be directly achieved by a single latent space without significantly scaling up the size of the latent space. 
Modeling the discrete latent space of such high-resolution images was previously handled by sophisticated designs, such as a three-layer hierarchy with $32 \times 32$, $64 \times 64$, and $128 \times 128$ latent space as in \cite{vqvae2}, which leads to over 5,000 sampling steps. 
In our method, we adopt a single latent space with a resolution of $32\times32$. Note that this setting gives a higher $32\times$ downsampling ratio since the commonly used $16\times$ downsampling ratio will lead to a $64\times64$ latent resolution that is unaffordable for the plain transformer architecture we adopt. 
As we will further discuss and compare in Section~\ref{discuss}, we consistently notice little compromise in the image quality with such a higher downsampling ratio. 
We present quantitative comparisons in Table~\ref{tab:highres} and qualitative results in Appendix Section~\ref{highres_imgs}.
To the best of our knowledge, our binary latent diffusion is the first diffusion model that generates such high-resolution image \textit{in one-shot}, i.e., without any latent hierarchy \cite{vqvae2} or multi-stage upsampling \cite{dalle2}.


\begin{table}[]
    \centering
    \small
    \caption{FID comparisons on $1024\times1024$ high-resolution image generation with FFHQ and CelebA-HQ. $\dagger$ indicates the results we obtain based on their official implementations. }
    \label{tab:highres}
    \resizebox{\linewidth}{!}{
    \begin{tabular}{c|c |c c}
    \toprule
         Methods & Steps & FFHQ $\downarrow$ & CelebA-HQ $\downarrow$\\
    \midrule
    
         StyleSwin \cite{styleswin} & - & 5.07 & 4.43  \\
         StyleGAN-XL \cite{sauer2022stylegan} & - & 2.02 & -	\\
         Diffusion StyleGAN2 \cite{diffusiongan} & $<$1,000 & 2.83 & - \\
         LDM$^\dagger$ \cite{latent} & 1,000 & 10.09 & 8.68 \\
         Absorbing$^\dagger$ \cite{unleashing} & 1,024 & 14.12 & 13.82\\ 
         
         \textbf{Ours} & 64 & 6.75 & 6.12 \\
         \textbf{Ours} & 256 & 6.35 & 6.03\\
    \bottomrule
    \end{tabular}}
\vspace{-2mm}
\end{table}

\subsection{Conditional Image Generation}

We adopt the ImageNet-1K dataset for image generation with class labels as conditions. 
In addition to FID and precision-recall, we follow the common practice and include the inception score (IS) \cite{salimans2016improved}  and the classification accuracy score (CAS) \cite{cas} with ResNet-50 \cite{resnet} as the evaluation metrics. 
Following the same settings as in the unconditional image generation experiments, we generate images at a resolution of $256\times256$, with a latent space of $16\times16\times64$. 
We present quantitative comparisons against state-of-the-art methods in Table~\ref{tab:imagenet}.
We adopt classifier-free guidance \cite{ho2022classifier} as explained in Appendix Section~\ref{cfg} with a small guidance scale of $\omega = 2.0$.
Our results demonstrate comparable image quality to state-of-the-art image generation methods and appealing sample diversity. The effectiveness is validated by the improved CAS scores. 
We present qualitative results in Figure~\ref{fig:imagenet} and Appendix Section~\ref{cond_imgs}.

\subsection{Discussions}
\label{discuss}

\noindent \textbf{Image Reconstruction.}
In this section, we present comparisons of the image reconstruction quality with different ways of formulating the latent space.
To assess the reconstruction quality, we use the peak signal-to-noise ratio (PSNR) and the structural similarity index measure (SSIM) \cite{ssim} as the metrics to measure the reconstruction quality. We measure the performance on the validation set of the LSUN bedrooms dataset at a resolution of $256 \times 256$, and present the results in Table~\ref{tab:recon}.
Compared to VQ-based representation, our method permits better image reconstruction quality with a compact $32$-bit binary representation for each image patch. The advantage is further amplified when a more compact spatial size ($8 \times 8$), which corresponds to a $32\times$ downsampling ratio is adopted. The higher downsampling ratio allows our method to generate high-resolution images in a one-shot fashion as reported in Section~\ref{exp_uncond}.

\newcommand*{\x}{\mathsf{x}\mskip1mu}
\newcommand{\jj}{\scriptsize $\pm\text{  }$}

\begin{table}[]
    \caption{Image reconstruction quality with different formats of latent code. The sizes are formulated as height $\times$ width $\times$ code (or codebook) size.}
    \label{tab:recon}
    \centering
    \small
    \begin{tabular}{c|c c c }
    \toprule
        Methods & Size & PSNR $\uparrow$ &  SSIM $\uparrow$\\
    \midrule
    
     VQ \cite{vqvae} & $16 \x 16 \x 1024$ & 20.08 \jj 1.84 & 0.62 \jj 0.09 \\
     Real value \cite{latent} & $16 \x 16 \x 16$ & 24.08 \jj 4.22 & 0.70 \jj 0.12 \\
     Binary (Ours) & $16 \x 16 \x 32$ & 24.02 \jj 2.11 & 0.75 \jj 0.06 \\
     
     \midrule
     \midrule
     
     VQ \cite{vqvae} & $8 \x 8 \x 2048$ & 17.20 \jj 1.45 & 0.53 \jj 0.09 \\
     Real value \cite{latent} & $8 \x 8 \x 64$ & 22.74 \jj 1.87 & 0.62 \jj 0.15 \\
     Binary (Ours) & $8 \x 8 \x 128$ & 22.82 \jj 1.84 & 0.64 \jj 0.07 \\
     
    \bottomrule
    \end{tabular}
\vspace{-2mm}
\end{table}

\noindent \textbf{Alternative samplers.}
To show the advantages of using the binary latent diffusion for the modeling of the prior over the Bernoulli latent distribution, we present in Table~\ref{tab:samplers} quantitative results with the samplers parametrized by alternative architectures including autoregressive models as in \cite{vqvae} and absorbing diffusion \cite{unleashing}. We report FID results on $256\times 256$ unconditional image generation with LSUN Churches and FFHQ. The proposed binary latent diffusion model tailored for the Bernoulli distribution 
achieves the best results on modeling the prior over the binary latent image representations. Compared to autoregressive models and absorbing diffusion models which perform greedy sampling, our method allows the prediction at previous denoising steps to be modified and improved in later steps, and prevents error accumulation across steps. 

\begin{table}[]
    \caption{FID results with alternative samplers for the modeling of the prior over the Bernoulli latent distribution. }
    \label{tab:samplers}
    \centering
    \small
    \begin{tabular}{c| c c c }
    \toprule
        Samplers & Steps & Churches $\downarrow$ & FFHQ $\downarrow$\\
    \midrule
        Autoregressive & 256 &7.25 & 8.23 \\
        Absorbing diffusion & 256 & 5.44 & 7.64 \\
        \textbf{Binary Diffusion (Ours)} & 64 & 4.36 & 5.85  \\
     
    \bottomrule
    \end{tabular}
\end{table}



\begin{figure}[]
    \centering
	\includegraphics[width=\linewidth]{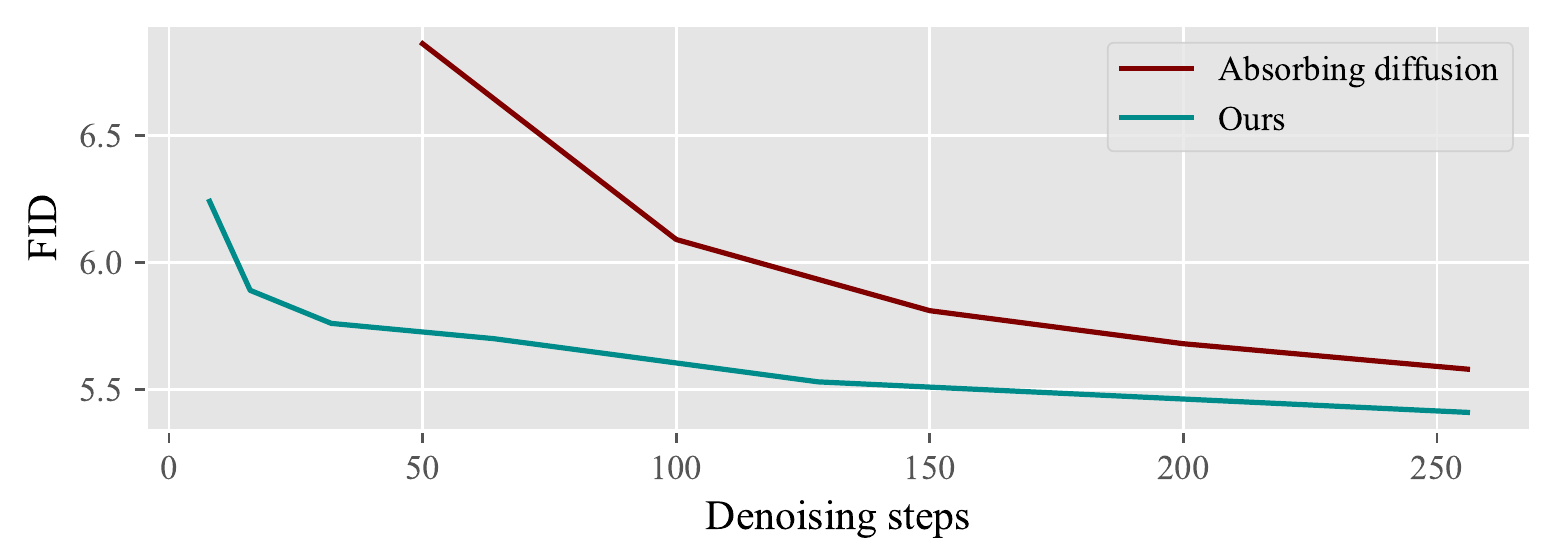}
\caption{FID with different denoising steps on the $256 \times 256$ LSUN churches experiment. For fair comparisons, we use $\tau = 1.0$ in all experiments. }
\label{fig:fid}
\vspace{-2mm}
\end{figure}

\noindent \textbf{Efficiency.}
One of the major advantages of the proposed efficient image generation with binary latent diffusion is that good results can be achieved with fewer steps of denoising, which results in a faster sampling speed in practice. We compare sampling speed against SyleGAN-2 \cite{stylegan2}, absorbing diffusion \cite{unleashing}, latent diffusion \cite{latent}, and DDPM \cite{ddpm} in Table~\ref{tab:speed}. Our method demonstrates clear advantages in image sampling speed compared to other diffusion-based methods, and further closes the gap between the speed of GANs and diffusion models. We further present comparisons on the generation results with different denoising steps against absorbing diffusion models \cite{unleashing} in Figure~\ref{fig:fid}. Our method demonstrates noticeably higher robustness to the number of steps, and fair results can be obtained with as few as 8 denoising steps. Meanwhile, we consistently observe that it is extremely hard to train Gaussian-based diffusion models such as latent diffusion models \cite{latent} and DDPM \cite{ddpm} to perform reasonably with fewer than 100 denoising steps in \textit{training}.

\begin{table}[]
    \caption{Comparisons on image generation speed with seconds per sample (s/sample). All results are obtained by averaging 1,000 times sampling with a batch size of 1. 64s and 16s denote 64 and 16 denoising steps, respectively. }
    \label{tab:speed}
    \centering
    \small
    \resizebox{\linewidth}{!}{
    \begin{tabular}{c|c c c c c c}
    \toprule
        Methods & StyleGAN-2 & Absorbing & LDM & DDPM & Ours 64s & Ours 16s\\
        
    \midrule
        s/sample & 0.04 & 3.40 & 15.68 & 63.85& 0.82 & 0.20 \\

    \bottomrule
    \end{tabular}}
\end{table}

\noindent \textbf{Ablation studies.}
We present in Table~\ref{tab:ablation} quantitative results that validate the values of $\lambda$ in (\ref{eq:final}) and the prediction targets of our binary latent diffusion. We consistently observe that the value of $\lambda$ correlates noticeably with the sample diversity, while a very large value of $\lambda$, such as $\lambda = 1.0$, frequently causes sampling divergence, which results in samples that appear as random compositions of multiple images, and thus degrades the quality. 
Setting the prediction targets as $\blatent^{t-1}$ corresponds to using only $\mathcal{L}_\text{vlb}$ as the supervision. While different parameterizations to the prediction targets achieve similar sample diversity, the proposed reparameterization of the targets with the flipping probability $\blatent^t \oplus \blatent^0$ achieves the best overall performance as it noticeably reduces the samples with strong artifacts caused by potentially diverged sampling according to our observations. 

\begin{table}[t]
    \caption{Performance (FID / Recall) with different values of $\lambda$ and prediction targets. All results are obtained on the LSUN Churches experiments. }
    \label{tab:ablation}
    \centering
    \small
    \resizebox{\linewidth}{!}{
    \begin{tabular}{c|c c c c }
    \toprule
        $\lambda$ values & $\lambda = 0$ & $\lambda = 0.01$ & $\lambda = 0.1$  & $\lambda = 1.0$ \\ 
        FID / R & 5.07 / 0.42 & 4.68 / 0.46 & \textbf{4.36 / 0.50} &  5.12 / 0.48\\
    \midrule
    \midrule
        Targets  & $\blatent^{t-1}$ & $\blatent^{0}$ & $\blatent^t \oplus \blatent^0$ & \multirow{2}{*}{-}\\
        FID / R & 5.09 / 0.48 & 4.72 / 0.50 & \textbf{4.36 / 0.50} & \\
    \bottomrule
    \end{tabular}}
\end{table}

\begin{figure}[]
    \centering
	\includegraphics[width=\linewidth]{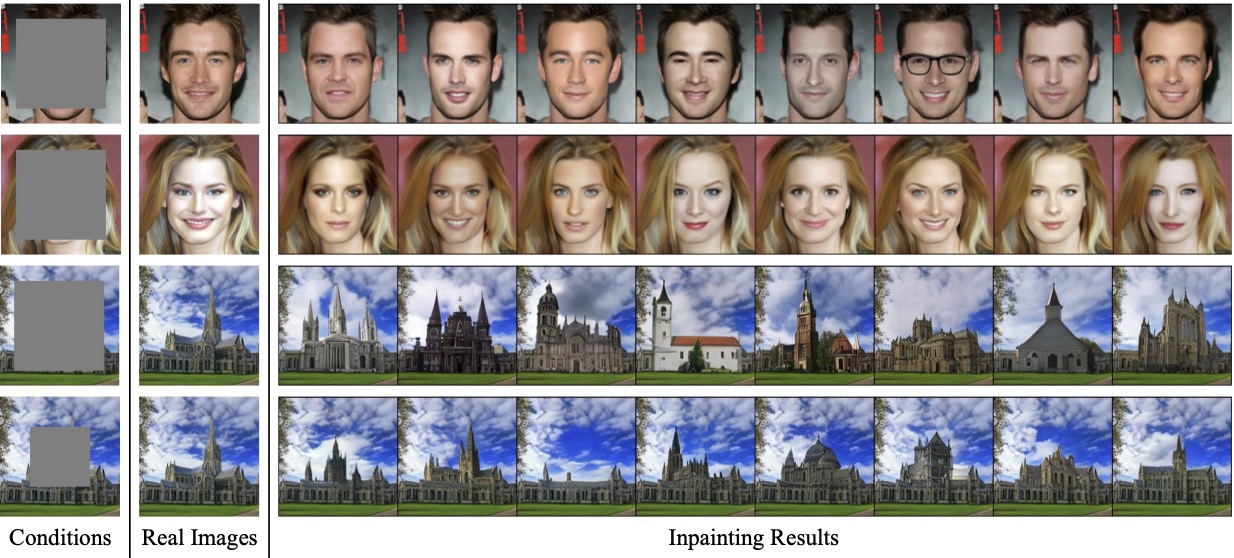}
\caption{Inpainting results. Given partial observations as conditions, our learned model can generate diverse samples with strong correspondence to the conditions. Diverse results can be obtained even with strong conditions given (last row). Zoom in for details.}
\label{fig:inpaint}
\end{figure}

\noindent \textbf{Image inpainting.}
Compared to the autoregressive-based samplers, one of the clear advantages of diffusion models is that the iterative sampling process in diffusion models does not assume any specific generation order of the spatial positions. This means that after training the generative models, we can generate images with partial observations as conditions. We present in Figure~\ref{fig:inpaint} examples of inpainting results generated by our models. Additional inpainting results can be found in Appendix Section~\ref{inpaint_imgs}.

\section{Conclusion}
\label{conclusion}

In this paper, we presented representing and generating images in a binary latent space. 
We learned binary image representations by training an auto-encoder with a multi-variate Bernoulli latent distribution and simple gradient copying to bypass the non-differential Bernoulli sampling operation. 
Compared to the real-value image representations in either the pixel or latent space, we showed that binary representation allows for a compact representation with the corresponding distribution that can be effectively modeled by a binary latent diffusion model. 
Compared to vector-quantized discrete representation, the binary representation in our work enables a higher expressiveness that permits high-resolution images to be generated without any multi-stage latent hierarchy. 
We examined our idea on multiple datasets with both conditional and unconditional image generation experiments. The comparable results to the recent state-of-the-art methods validated the effectiveness of the proposed method of representing and generating images in a binary latent space. 

\noindent \textbf{Social impacts.} As a framework for representing and generating images, the proposed binary latent diffusion model shares with other generative models the risk of malicious uses such as deepfake. And the generated sample may exhibit bias residing in the training datasets. 

{\small
\bibliographystyle{ieee_fullname}
\bibliography{egbib}
}

\clearpage
\appendix
\onecolumn 

\newcount\cvprrulercount
\setcounter{page}{1}

\setcounter{table}{0}
\setcounter{equation}{0}
\setcounter{figure}{0}

\renewcommand{\thetable}{\Alph{table}}
\renewcommand{\thefigure}{\Alph{figure}}

\section{Noise Scheduler}
\label{noise}

\begin{figure}[h]
    \centering
    \includegraphics[width=\textwidth]{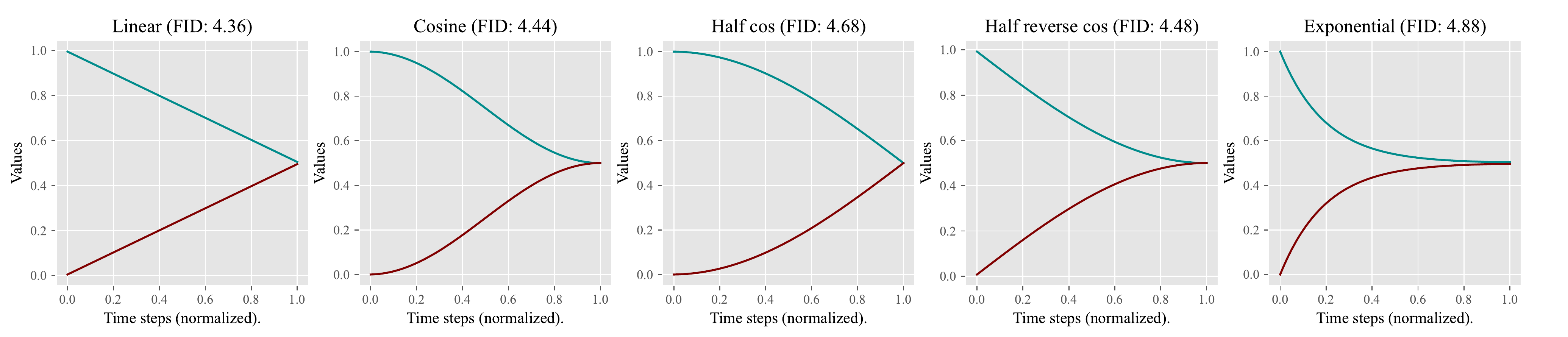}
    \caption{Noise schedulers. FID numbers are obtained with the $256\times 256$ LSUN Churches experiments. }
\label{fig:noise}
\end{figure}

We show in Figure~\ref{fig:noise} that with different noise schedulers, how $1$ (blue curves) and $0$ (red curves) in the binary latent representations progressively reach $0.5$ in the diffusion processes. We use the \textit{Linear} noise scheduler in all the experiments. 

\section{Implementation Details}
\label{details}

We use Adam as the default optimizer for all experiments. 
For the training of the binary auto-encoder in Section~\ref{bae}, we use a consistent learning rate of $5\times 10^{-4}$ with a linear learning rate warm-up for 10k iterations, and a batch size of 8. 
For $\mathcal{C}$ in (\ref{eq:train_bae}), we use mean squared error, perception loss, and adversarial loss, with an equal $\omega_i = 1.0$ for each term.

For the training of the binary latent diffusion models in Section~\ref{bdiffusion}, we use a consistent learning rate of $1\times 10^{-4}$ with a linear learning rate warm-up for 10k iterations. 
All models are trained for 20K iterations. We do not use any additional regularizations such as drop-out or weight decay. We summarize the general training and sampling of the proposed binary diffusion models in Algorithm~\ref{alg:train} and Algorithm~\ref{alg:sampling}, respectively.

All $256\times256$ unconditional binary latent diffusion models are trained with a batch size of 32. 
All $1024\times1024$ unconditional binary latent diffusion models are trained with a batch size of 16. 
All $256\times256$ conditional binary latent diffusion models are trained with a batch size of 256.  

All training is conducted on cloud servers with V100 GPUs. All the speeding testing is conducted on a single RTX3090 GPU. 

\subsection{Network Architectures}

For the binary auto-encoder in Section~\ref{bae}, we adopt a standard architecture using convolutional layers enhanced with self-attention layers, which is nearly identical to the one used in \cite{vqgan}. The only major difference is that the vector quantization layer is replaced by our binary latent layer implementing (\ref{eq:binarize}). 
This architecture gives a downsampling ratio of $16\times$. For the $1024 \times 1024$ high-resolution image generation experiments, of which the downsampling ratio raises to $32\times$, we simply insert one more downsampling block and one more upsampling block into the architecture.

For the conditional image generation experiments with ImageNet-1K, we train transformer networks with 24 layers and 768 feature channels with 12 heads in the self-attention layers. 
For the unconditional image generation experiments, we train transformer networks with 24 layers and 512 feature channels with 8 heads
in the self-attention layers. 
In both conditional and unconditional image generation experiments, we use an extra time embedding token to specify the time step $t$. And the class labels in conditional image generation with ImageNet are also specified by an additional token. Both class tokens and time-step tokens are trainable. 
We initialize the trainable position embedding in the transformer networks with 2D sine embedding.

\subsection{Evaluation Metrics}

\noindent \textbf{FID.} Fr\'echet Inception Distance (FID) is a commonly used metric for evaluating the quality of the generated images. It estimates the distance between two image distributions by comparing the mean and standard deviation of the deep image features extracted by a trained Inception network \cite{inception}.
We conduct comparisons by generating 50,000 samples and comparing them with the corresponding training dataset in every experiment.

\noindent \textbf{PR.} To comprehensively evaluate both the quality and mode-coverage of the generated samples compared to the training datasets, we further include precision-recall \cite{pr} as additional performance measurements.

\noindent \textbf{IS.} Inception score \cite{salimans2016improved} is a commonly used metric for evaluating the performance of class-conditional image generation. It favors generated images with low entropy of label predictions and diverse labels given a pretrained Inception-V3 network. 

\noindent \textbf{CAS.} Classification accuracy score \cite{cas} works by first training a ResNet-50 \cite{resnet} using the generated image across classes, and measuring the results of applying the trained classifier to the ImageNet validation set. CAS offers a comprehensive measurement of the generative quality as a robust classifier demands the generated images used for network training to be both diverse and of high quality.

\begin{algorithm}[]
	\small
	\caption{Training procedure. We assume unconditional image generation with a batch size of \textit{one} for the sake of discussion. The described training process can be easily extended to practical cases with arbitrary batch sizes by batching multiple samples. }
	\label{alg:train}
	\begin{algorithmic}[1]
		\STATE {\textbf{Given}: Trained encoder $\encoder$; Binary diffusion model $f_\theta$ parametrized by $\mathcal{T}_\theta$; An image dataset $\mathbf{X}$.} 
		\STATE{\textbf{Given}: Diffusion steps $T$; Noise scheduler defined by $\{k^t\}_{t=1}^{T}$ and $\{b_t\}_{t=1}^{T}$; Training steps $I$; and $\lambda$ in (\ref{eq:final}).}
		\STATE{Initializing $\mathcal{T}_\theta$.}
        \FOR{Step $i = 1:I$}
		\STATE{Sampling image $\img \sim \mathbf{X}$, and time step $t \sim \{1, \dots, T\}$.}
		\STATE{Obtaining binary code $\blatent^0 = \text{Bernoulli}(\sigma(\encoder(\img)))$.}
		\STATE{Obtaining $\blatent^t$ using $\blatent^0$, $t$, and noise scheduler with (\ref{eq:q_at_anytime}).}
		\STATE{Predicting flipping probability $f_\theta(\blatent^t, t)$.}
		\STATE{Obtaining predicted $\blatent^0$ as $p_\theta(\blatent^0) = (1 - \blatent^t) \odot f_\theta(\blatent^t, t) + \blatent^t \odot (1 - f_\theta(\blatent^t, t))$.}
		\STATE{Obtaining predicted $p_\theta(\blatent^{t-1})$ using $p_\theta(\blatent^{0})$ and $\blatent^t$ with (\ref{eq:one_step}). }
		\STATE{Calculating loss $\mathcal{L}$ using (\ref{eq:final}).}
		\STATE{Backpropagating $\mathcal{L}$ and updating $\theta$.}
		\ENDFOR
		\STATE{\textbf{Return} Binary diffusion model $f_\theta$.}
	\end{algorithmic}
\end{algorithm}

\begin{algorithm}[]
	\small
	\caption{Sampling procedure. We assume unconditional image generation with a batch size of \textit{one} for the sake of discussion. }
	\label{alg:sampling}
	\begin{algorithmic}[1]
		\STATE {\textbf{Given}: Trained decoder $\decoder$; Trained binary diffusion model $f_\theta$.} 
		\STATE{\textbf{Given}: Diffusion steps $T$; Noise scheduler defined by $\{k^t\}_{t=1}^{T}$ and $\{b_t\}_{t=1}^{T}$; Temperature $\tau$; Latent dimension specified by $h^\prime, w^\prime, c$, e.g., $h^\prime = w^\prime = 16, c = 32$ for the $256\times 256$ image generation experiments. }
        \STATE{Sampling $\blatent^T = \text{Bernoulli}(\blatent^{\text{init}})$, where $\blatent^{\text{init}} \in \mathbb{R}^{h^\prime \times w^\prime \times c}$ and contains $0.5$ only. }
        \FOR{Step $t = T:2$}
        \STATE{Predicting $p_\theta(\blatent^{t-1})$ with $f_\theta(\blatent^t, t) = \sigma(\mathcal{T}_\theta(\blatent^t, t) / \tau)$ and (\ref{eq:one_step}).}
        \STATE{Sampling $\blatent^{t-1} = \text{Bernoulli}(p_\theta(\blatent^{t-1}))$}
		\ENDFOR
		\STATE{\textbf{Return} the sampled image as $\decoder(\blatent^{t-1})$.}
	\end{algorithmic}
\end{algorithm}

\section{Classifier-free Guidance}
\label{cfg}
Classifier-free guidance \cite{ho2022classifier} is introduced to improve the generation fidelity and promote high correspondence to the conditions for conditional diffusion models. 
While it is previously believed that classifier-free guidance can hardly be extended to diffusion models, we show that the reparameterization to the prediction target as the binary residuals allows classifier-free guidance to apply seamlessly to the proposed binary latent diffusion for improved image fidelity in conditional image generation. 
In conditional image generation, we introduce an extra condition token $c$ that carries the conditions such as class embedding in ImageNet class-conditional image generation or text embedding in the text-to-image generation. The model performs class-conditional prediction as $f_\theta^\text{c}(\blatent^t, t, c) = \sigma(\mathcal{T}_\theta(\blatent^t, t, c))$. 
An unconditional prediction can be drawn by simply dropping the condition token $c$ as $f_\theta^\text{u}(\blatent^t, t) = \sigma(\mathcal{T}_\theta(\blatent^t, t))$. The final prediction at each step with classifier-free guidance can then be implemented as  $f_\theta(\blatent^t, t, c) =  \sigma((1 + \omega) \mathcal{T}_\theta(\blatent^t, t, c) - \omega \mathcal{T}_\theta(\blatent^t, t) )$, where $\omega$ is a non-negative scalar controlling the strength of the guidance. Note that same the temperature scale $\tau$, the guidance strength $\omega$ is only effective in the sampling stage and is not involved in training, which allows us to arbitrarily adjust the sampling quality without retraining the models. While it is inevitable that the classifier-free guidance doubles that computation cost at each sampling step due to the extra unconditional predictions, we consistently observe performing classifier-free guidance allows us to skip sampling steps. For example, performing only a quarter of the sampling steps enhanced with classifier-free guidance reduce the overall computation cost by half, and does not noticeably reduce the sample quality. We present in Appendix Figure~\ref{fig:app_omega} results generated with different scales of classifier-free guidance.

\section{Additional Qualitative Results}

\subsection{Unconditional Image Generation}
\label{uncond_imgs}

\begin{figure}[t]
    \centering
	\includegraphics[width=0.92\linewidth]{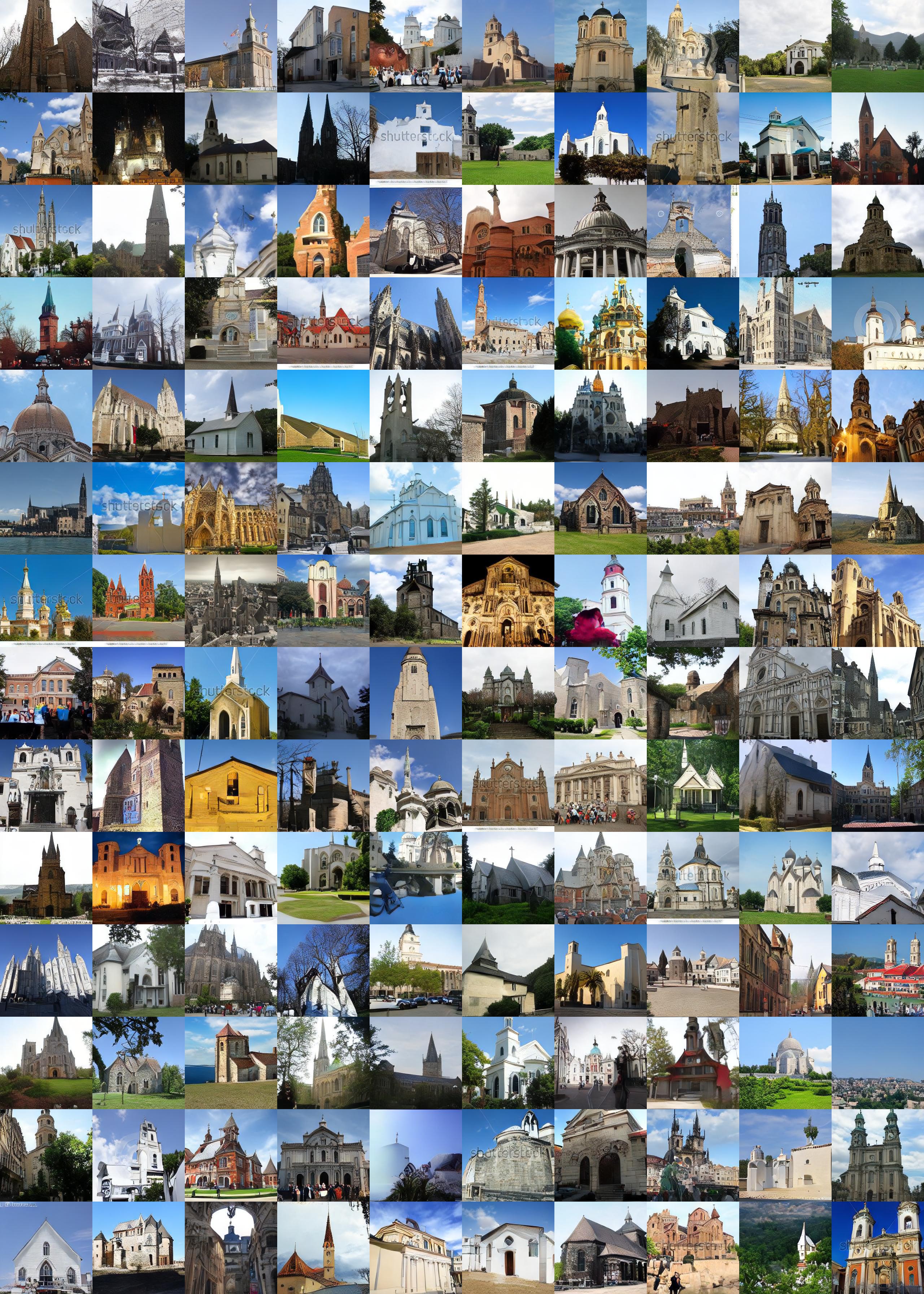}
\caption{Additional unconditional image generation results and comparisons at $256\times256$ with the LSUN Churches dataset. }
\label{fig:app_churches256}
\end{figure}

\begin{figure}[t]
    \centering
	\includegraphics[width=0.92\linewidth]{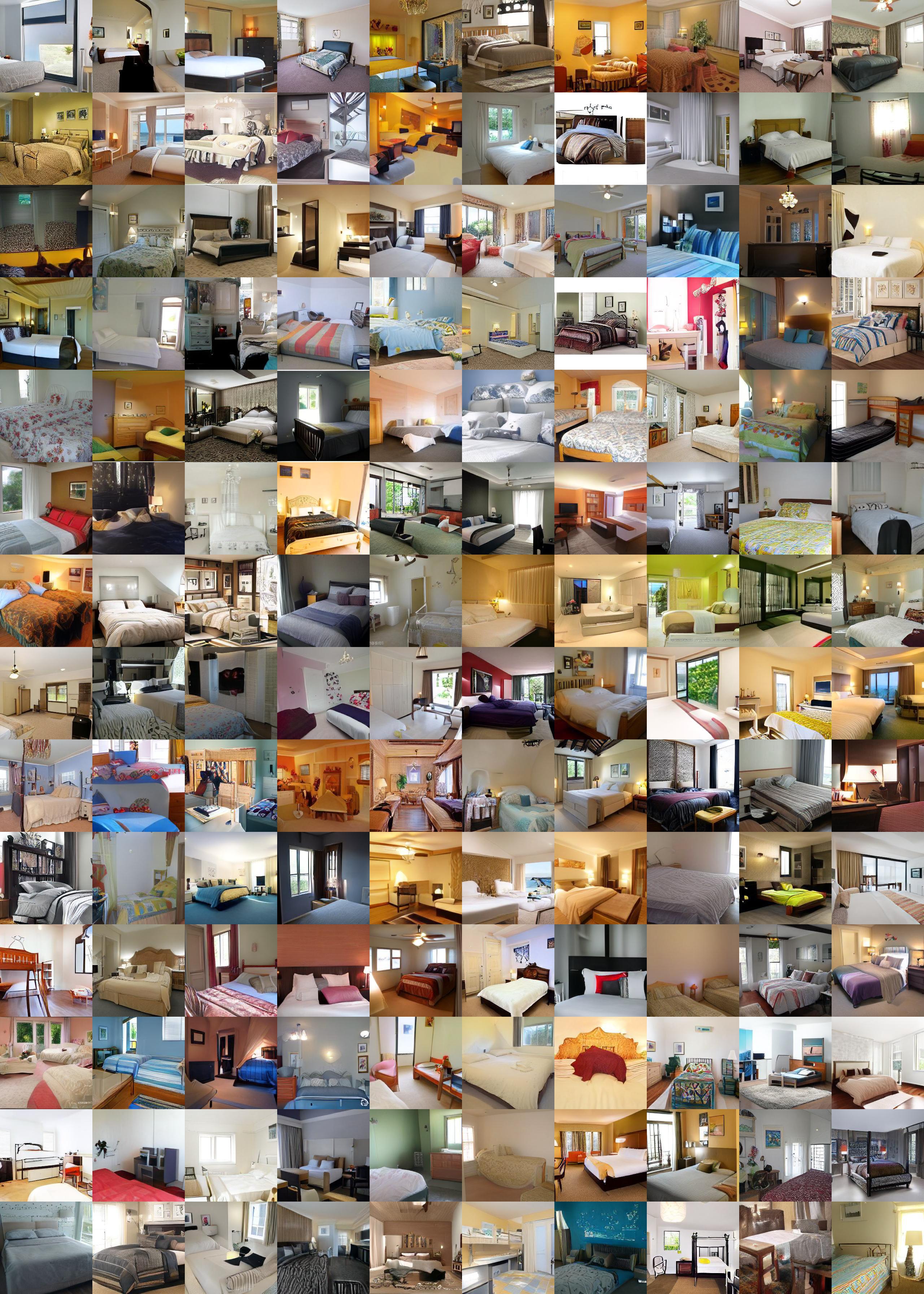}
\caption{Additional unconditional image generation results and comparisons at $256\times256$ with the LSUN Bedrooms dataset. }
\label{fig:app_beds256}
\end{figure}

\begin{figure}[t]
    \centering
     \begin{subfigure}[b]{\textwidth}
         \centering
         \includegraphics[width=\textwidth]{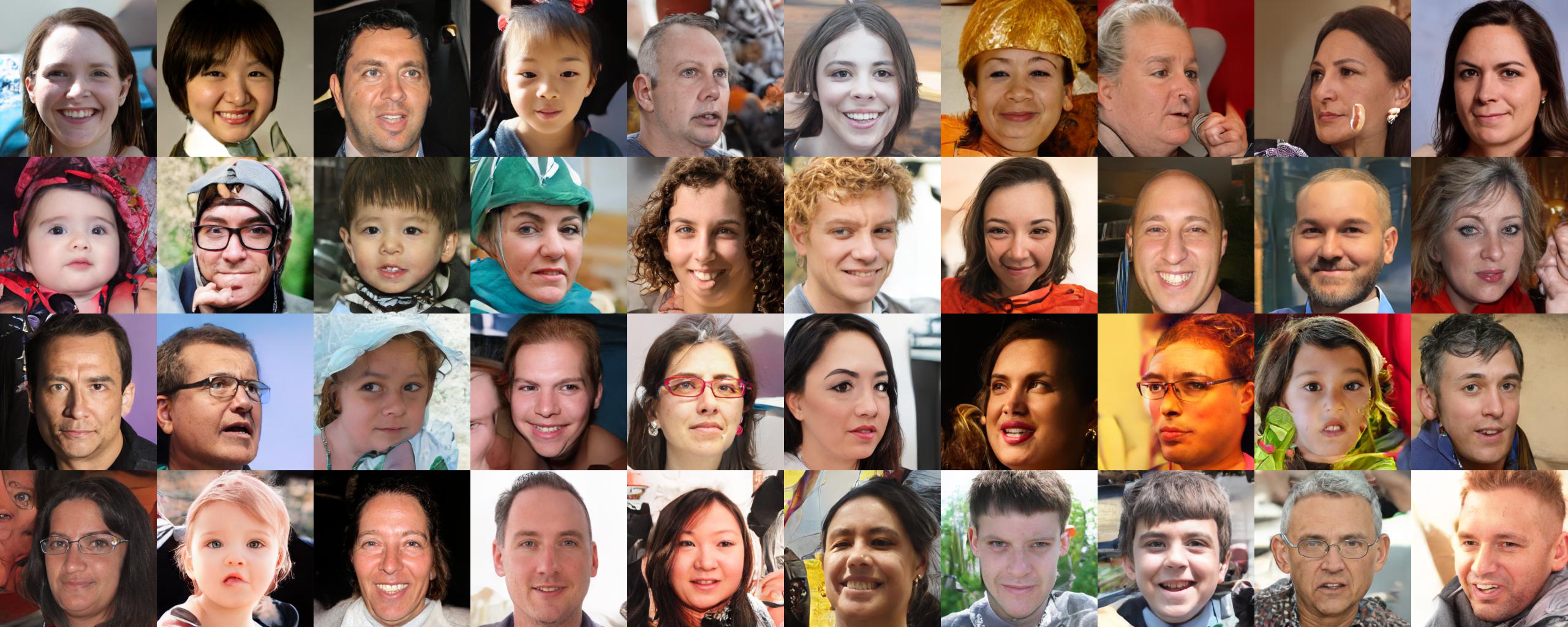}
         \caption{Absorbing diffusion.}
     \end{subfigure}
     \\
     \vspace{2mm}
     \begin{subfigure}[b]{\textwidth}
         \centering
         \includegraphics[width=\textwidth]{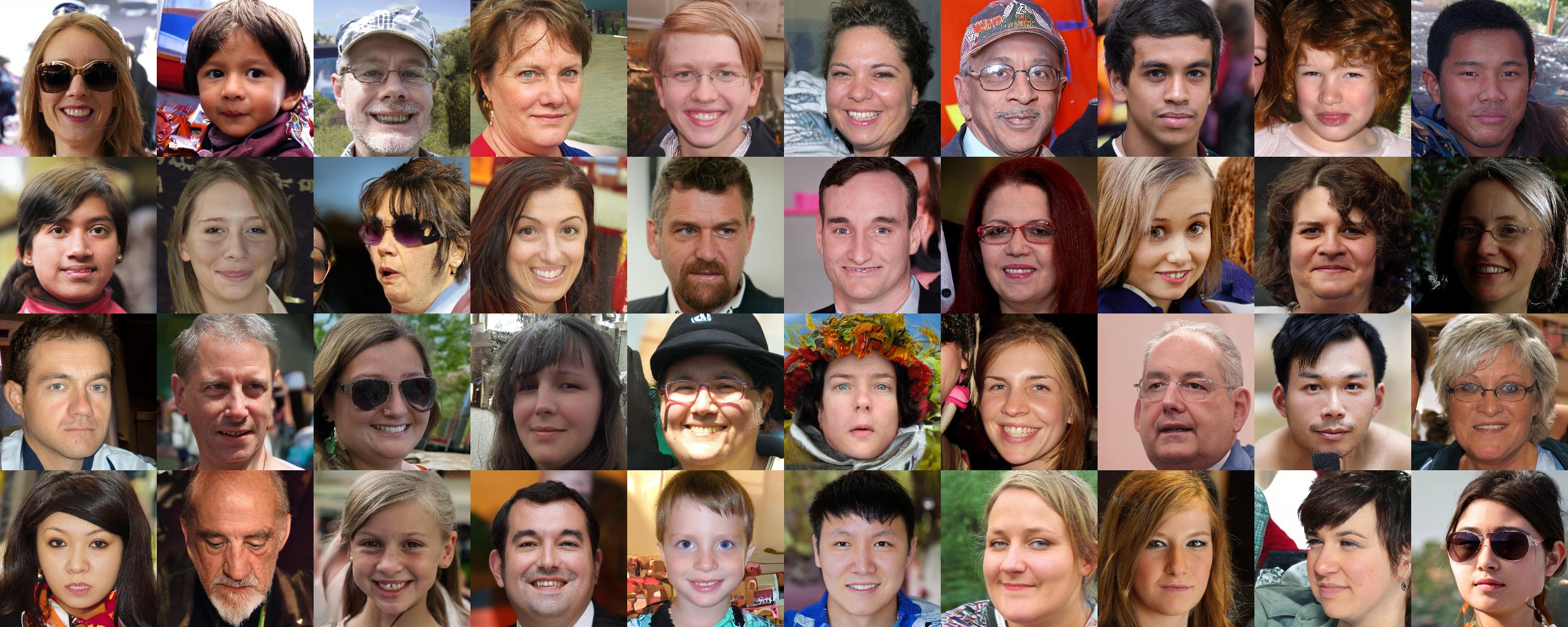}
         \caption{Latent diffusion.}
     \end{subfigure}
     \\
     \vspace{2mm}
     \begin{subfigure}[b]{\textwidth}
         \centering
         \includegraphics[width=\textwidth]{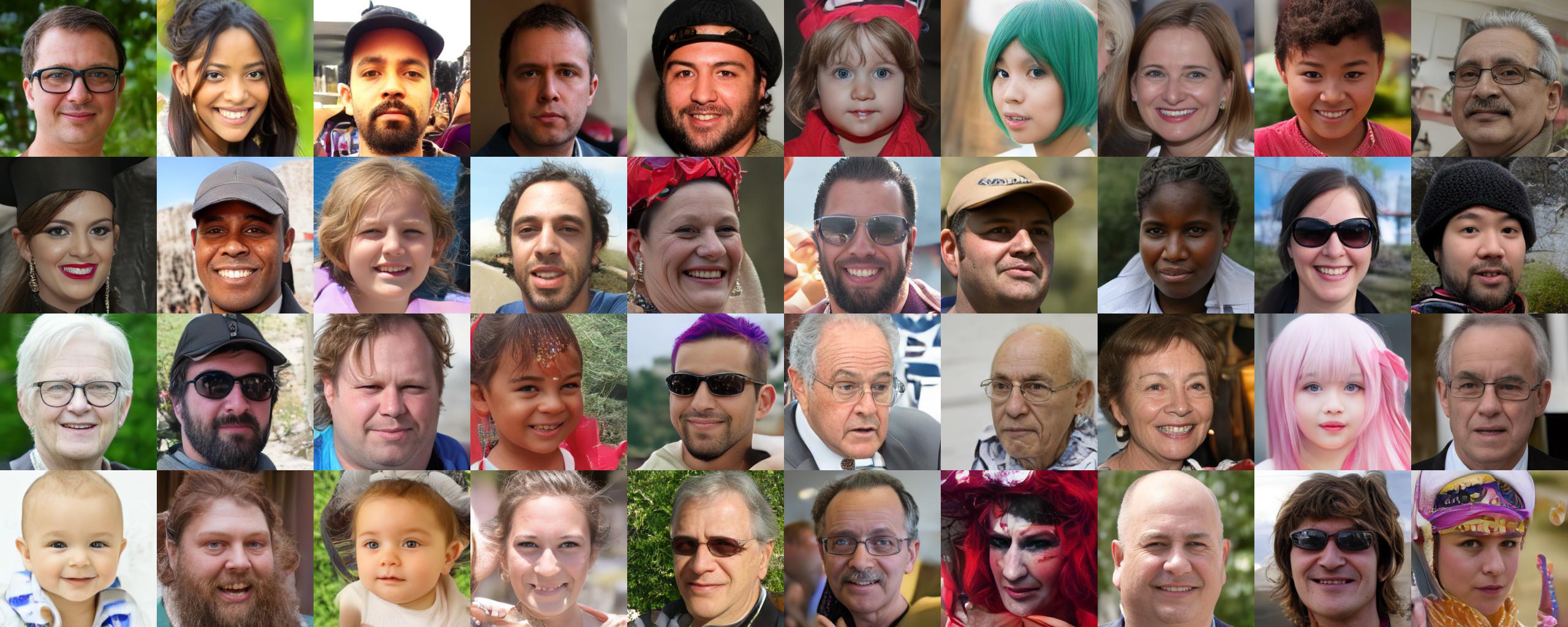}
         \caption{Ours.}
     \end{subfigure}
    \caption{Additional unconditional image generation results and comparisons at $256\times256$ with the FFHQ dataset.}
    \label{fig:app_ffhq}
\end{figure}

We present additional qualitative results of $256 \times 256$ LSUN Bedrooms and LSUN Churches unconditional image generation experiments in Figure~\ref{fig:app_churches256} and Figure~\ref{fig:app_beds256}, and qualitative comparisons of FFHQ unconditional image generation against Absorbing Diffusion models \cite{unleashing} and Latent Diffusion Models \cite{latent} in Figure~\ref{fig:app_ffhq}.

\subsection{High-resolution Image Generation}
\label{highres_imgs}
\begin{figure}[t]
    \centering
	\includegraphics[width=0.93\linewidth]{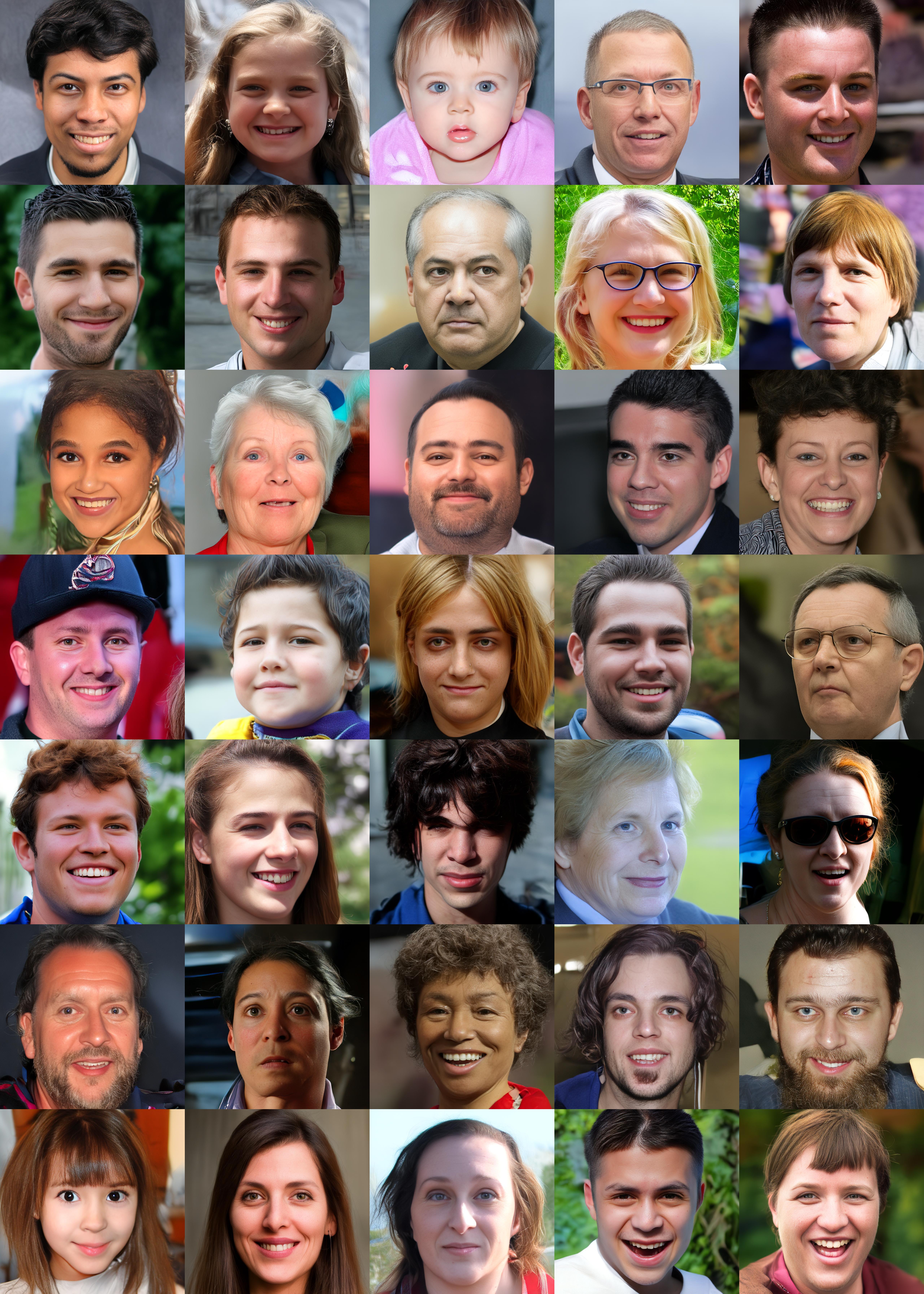}
\caption{Additional high-resolution image generation results at $1024\times1024$ with the FFHQ dataset ($\tau = 0.8$). }
\label{fig:app_ffhq1024}
\end{figure}

\begin{figure}[t]
    \centering
	\includegraphics[width=0.93\linewidth]{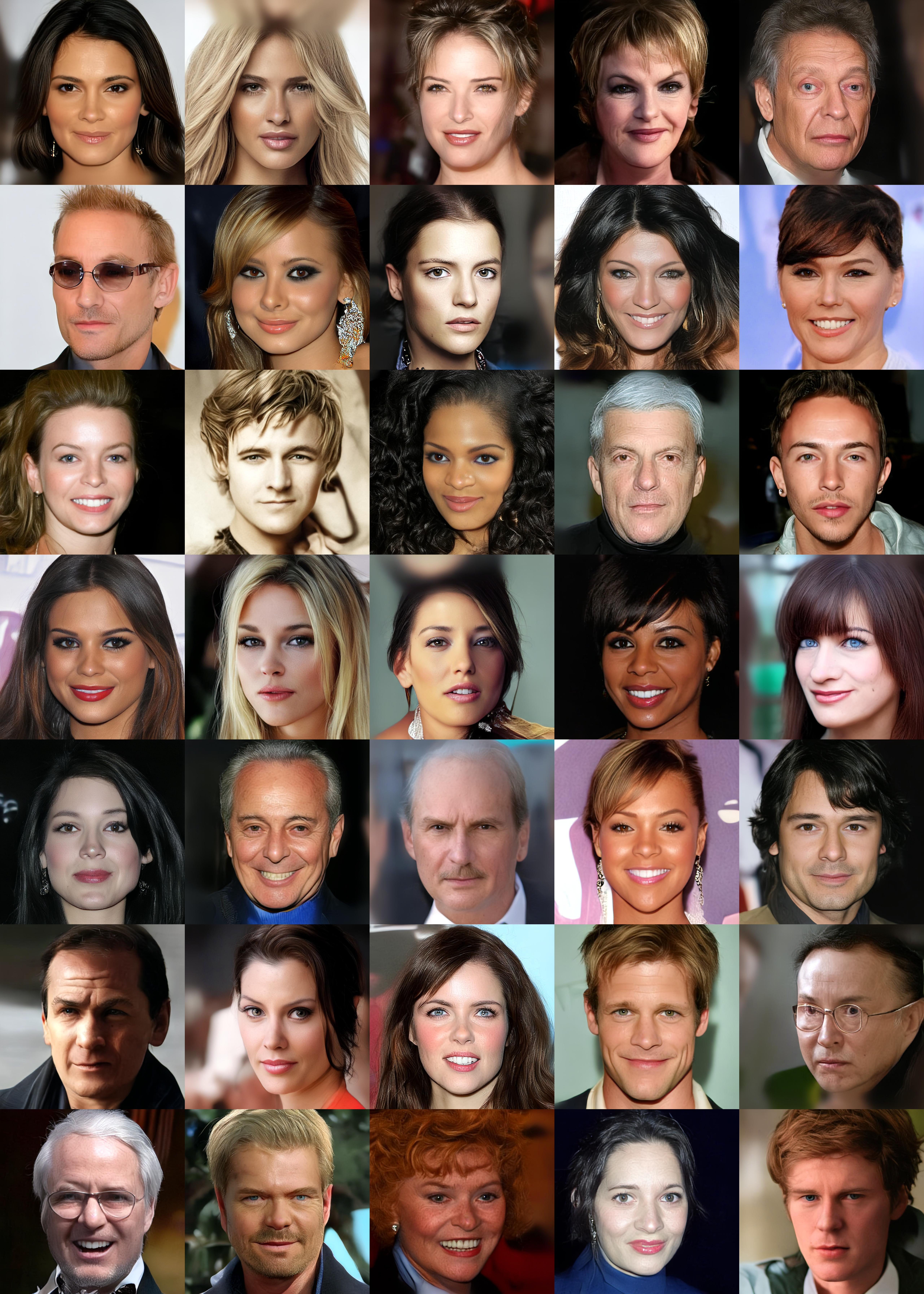}
\caption{Additional high-resolution image generation results at $1024\times1024$ with the CelebA-HQ dataset ($\tau = 0.8$). }
\label{fig:app_celeb1024}
\end{figure}

We present additional qualitative results of $1024 \times 1024$ unconditional image generation experiments in Figure~\ref{fig:app_ffhq1024} and Figure~\ref{fig:app_celeb1024}.

\subsection{Conditional Image Generation}
\label{cond_imgs}

\begin{figure}[t]
    \centering
     \begin{subfigure}[b]{0.19\textwidth}
         \centering
         \includegraphics[width=\textwidth]{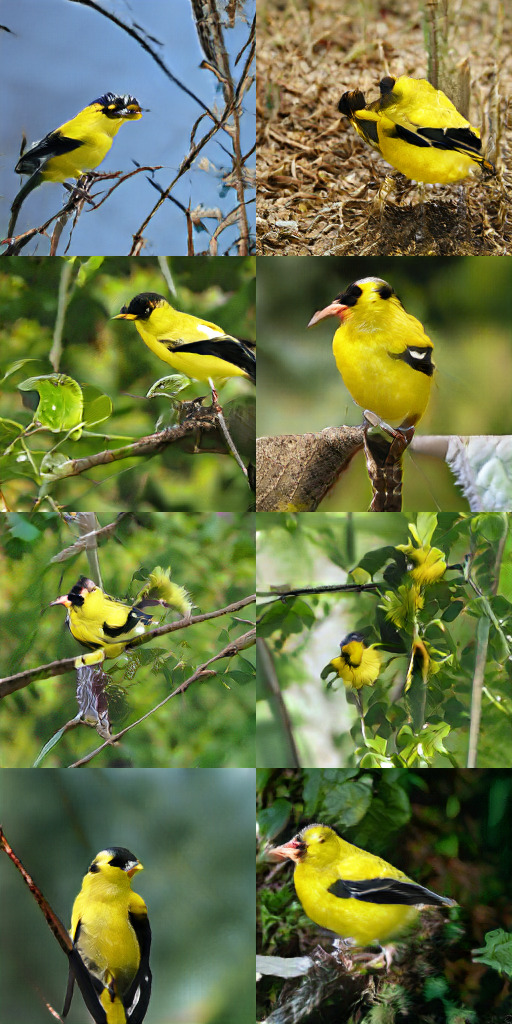}
         \caption{BigGAN-deep.}
     \end{subfigure}
     \begin{subfigure}[b]{0.19\textwidth}
         \centering
         \includegraphics[width=\textwidth]{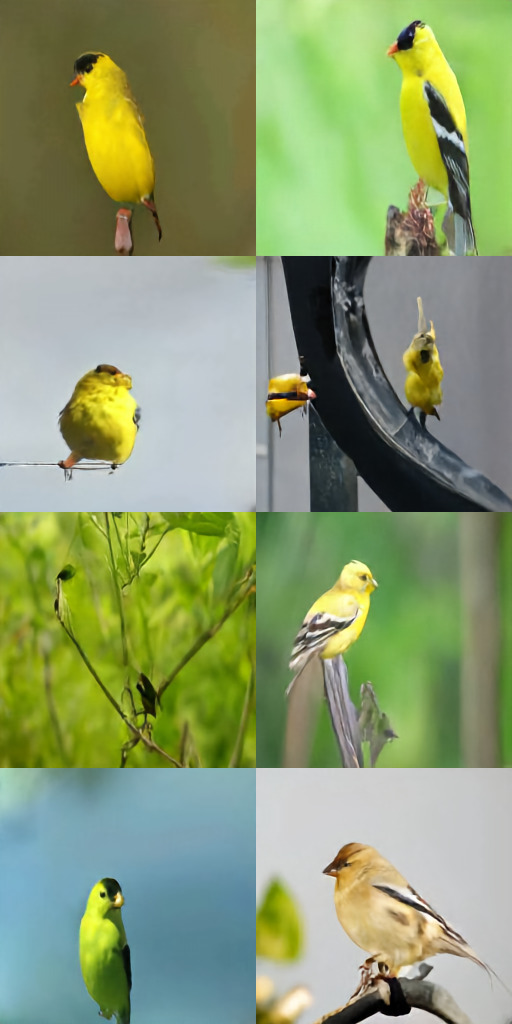}
         \caption{VQ-VAE-2.}
     \end{subfigure}
     \begin{subfigure}[b]{0.19\textwidth}
         \centering
         \includegraphics[width=\textwidth]{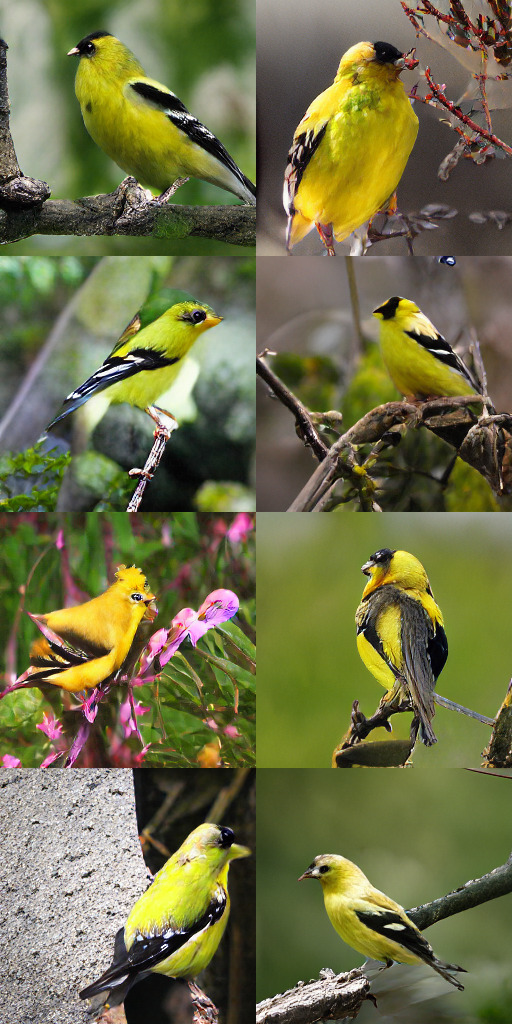}
         \caption{VQGAN.}
     \end{subfigure}
     \begin{subfigure}[b]{0.19\textwidth}
         \centering
         \includegraphics[width=\textwidth]{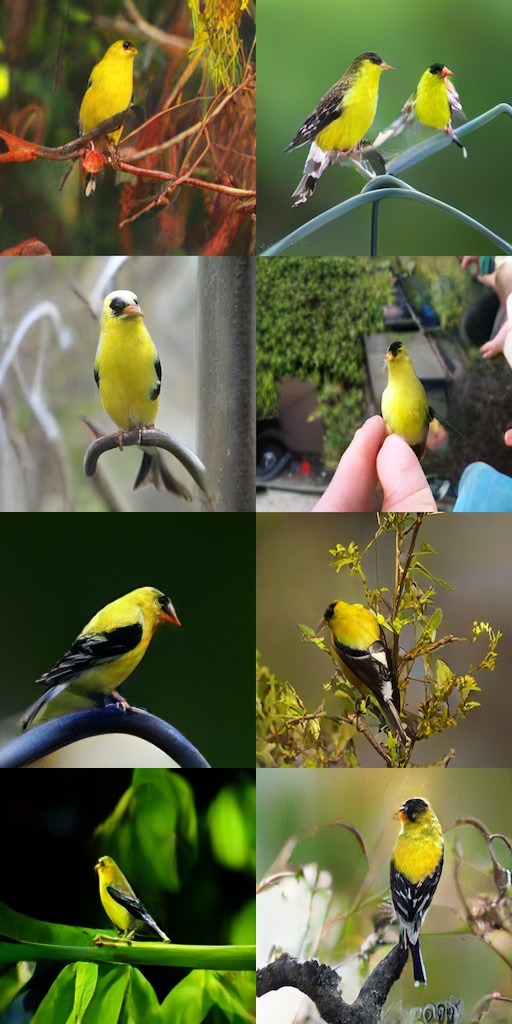}
         \caption{MaskGIT.}
     \end{subfigure}
     \begin{subfigure}[b]{0.19\textwidth}
         \centering
         \includegraphics[width=\textwidth]{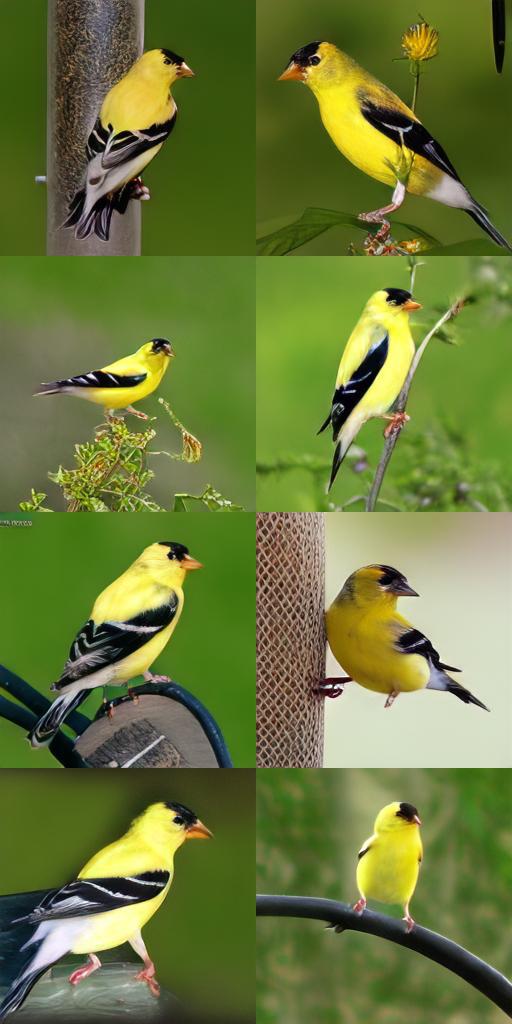}
         \caption{Ours.}
     \end{subfigure}
\caption{Qualitative comparisons on class-conditional image generation with ImageNet class IDs: 22, 108, and 11.}
\label{fig:conds_app}
\end{figure}

\begin{figure}[t]
    \centering
     \begin{subfigure}[b]{0.19\textwidth}
         \centering
         \includegraphics[width=\textwidth]{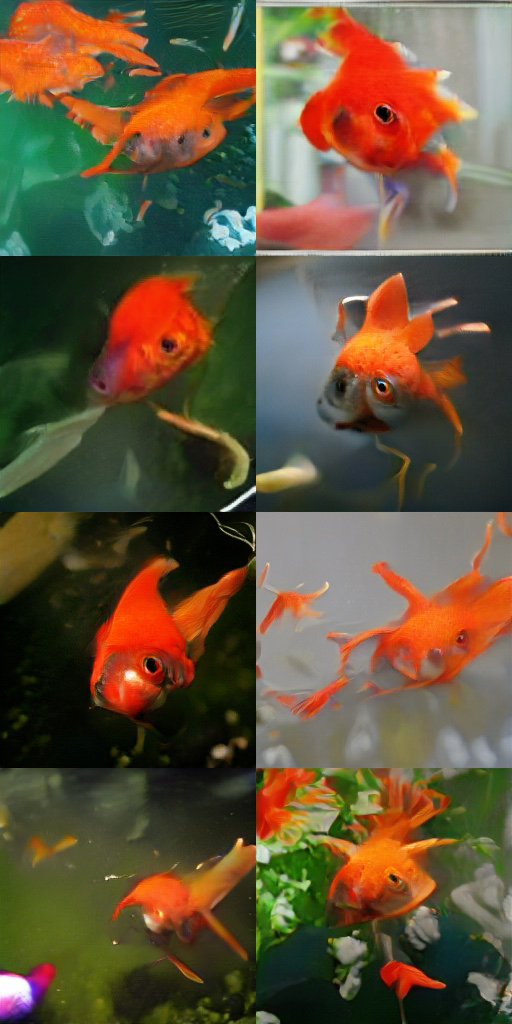}
         \caption{BigGAN-deep.}
     \end{subfigure}
     \begin{subfigure}[b]{0.19\textwidth}
         \centering
         \includegraphics[width=\textwidth]{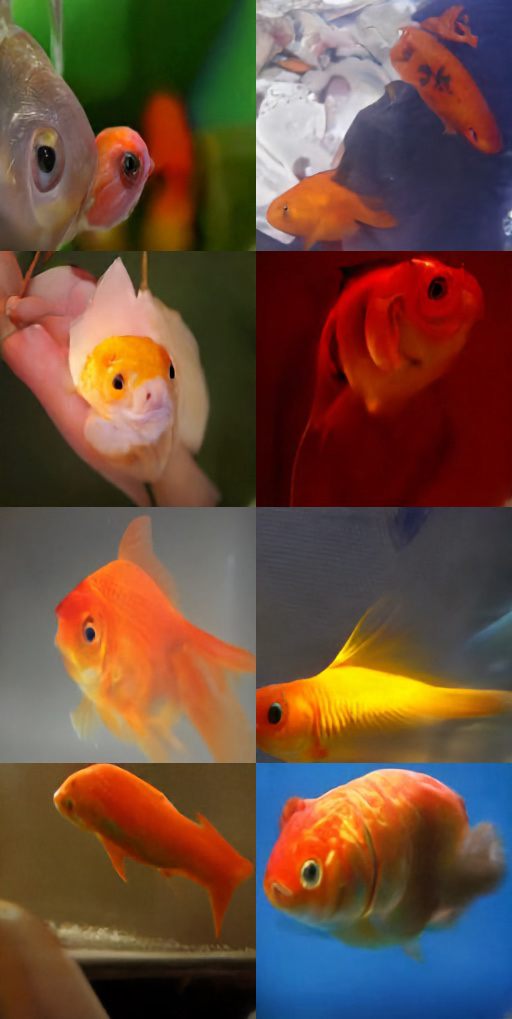}
         \caption{VQ-VAE-2.}
     \end{subfigure}
     \begin{subfigure}[b]{0.19\textwidth}
         \centering
         \includegraphics[width=\textwidth]{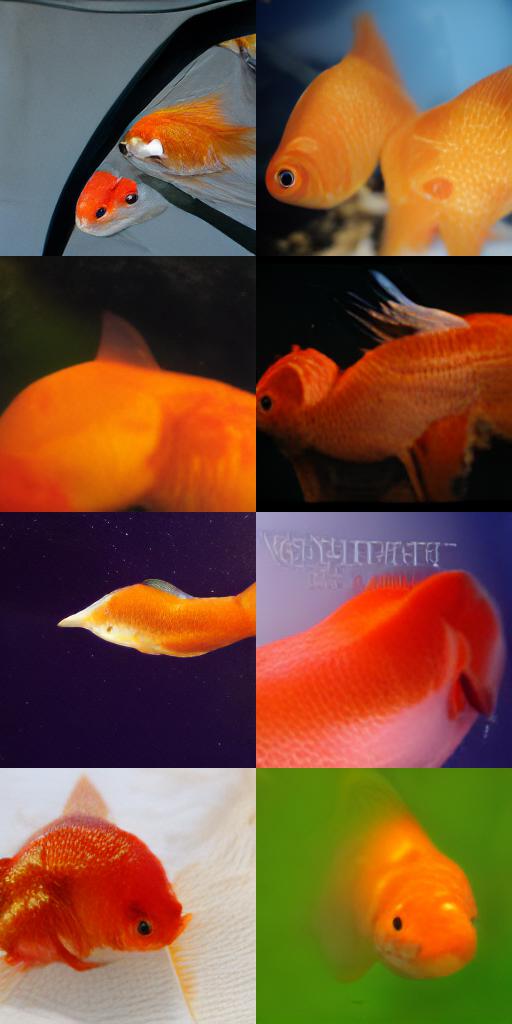}
         \caption{VQGAN.}
     \end{subfigure}
     \begin{subfigure}[b]{0.19\textwidth}
         \centering
         \includegraphics[width=\textwidth]{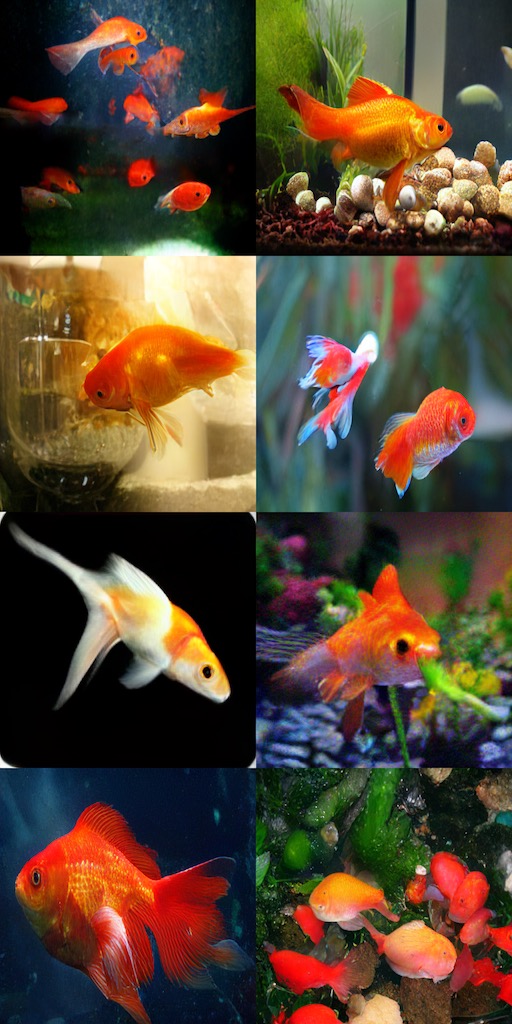}
         \caption{MaskGIT.}
     \end{subfigure}
     \begin{subfigure}[b]{0.19\textwidth}
         \centering
         \includegraphics[width=\textwidth]{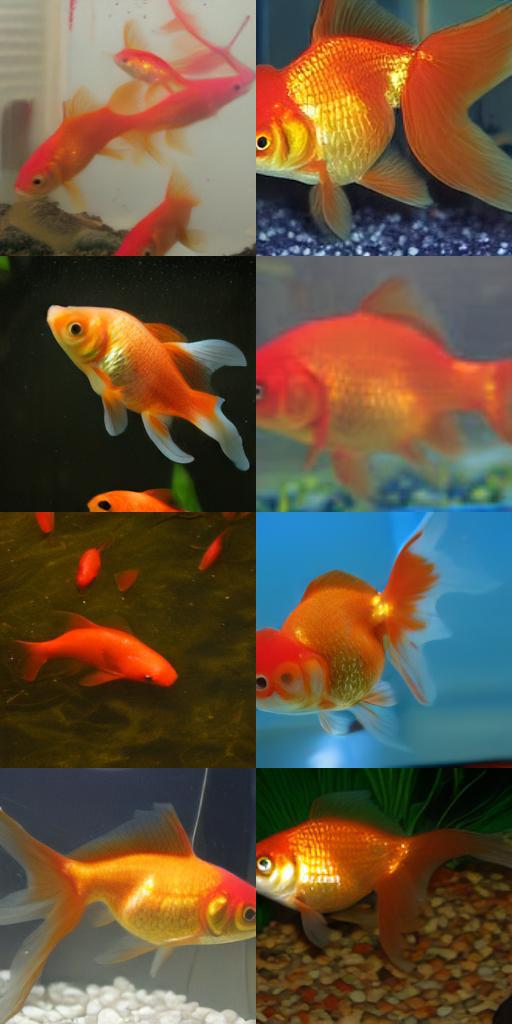}
         \caption{Ours.}
     \end{subfigure}
     \caption{Qualitative comparisons on class-conditional image generation with ImageNet class IDs: 141, 154, and 1.}
    \label{fig:conds_app2}
\end{figure}

We present additional qualitative results of class-conditional image generation experiments and comparisons with BigGAN-deep \cite{biggan}, VQ-VAE-2 \cite{vqvae2}, VQGAN \cite{vqgan} and MaskGIT \cite{maskgit} in Figure~\ref{fig:conds_app} and Figure~\ref{fig:conds_app2}.

\subsection{Image Inpainting}
\label{inpaint_imgs}

We present additional image inpainting results with different mask patterns in Figure~\ref{fig:app_inpaint}.

\subsection{Nearest Neighbors}

To further show that our model is generating novel samples instead of overfitting to the training datasets, we compare the generated images with the corresponding training datasets using LPIPS, and visualize the top-10 nearest neighbors in Figure~\ref{fig:nn}.

\clearpage
\begin{figure}[t]
    \centering
	\begin{subfigure}[b]{0.19\textwidth}
         \centering
         \includegraphics[width=\textwidth]{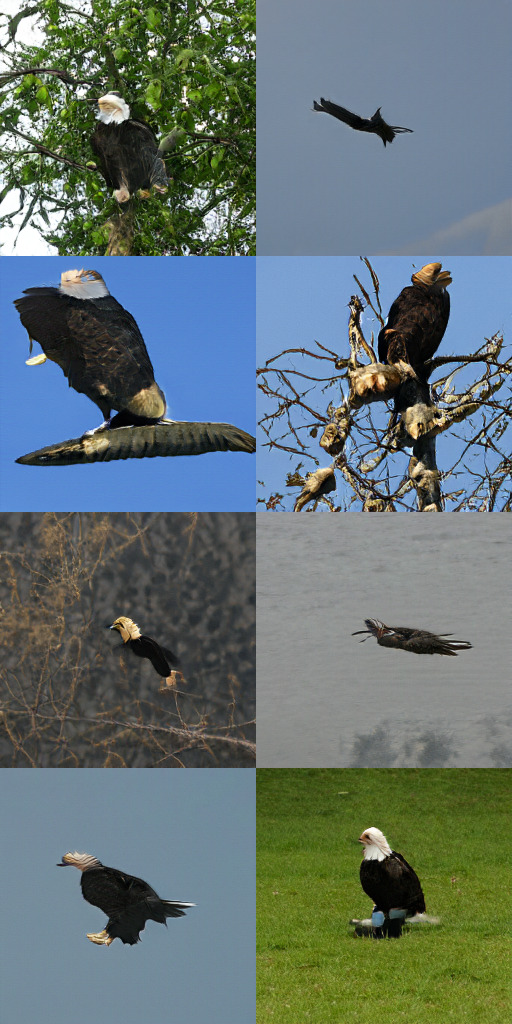}
         \caption{BigGAN-deep.}
     \end{subfigure}
     \begin{subfigure}[b]{0.19\textwidth}
         \centering
         \includegraphics[width=\textwidth]{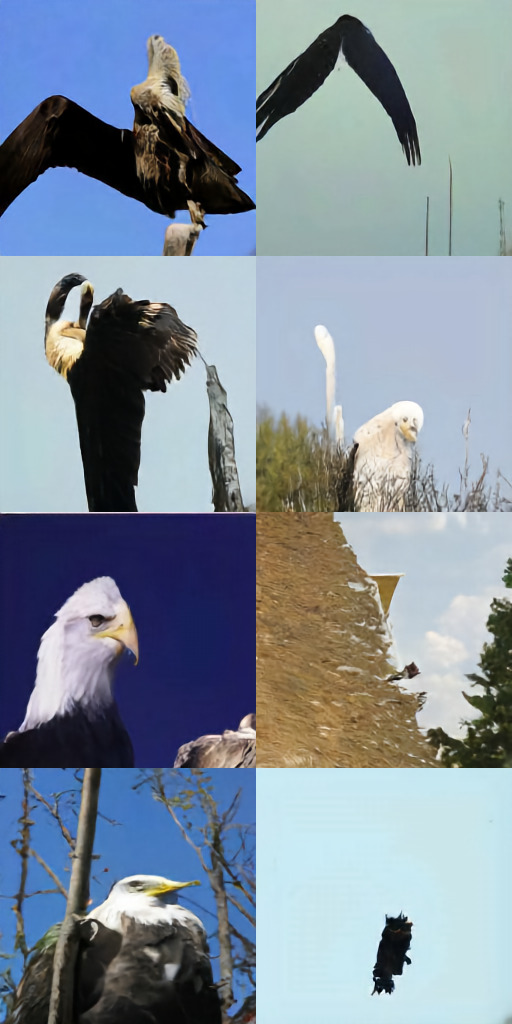}
         \caption{VQ-VAE-2.}
     \end{subfigure}
     \begin{subfigure}[b]{0.19\textwidth}
         \centering
         \includegraphics[width=\textwidth]{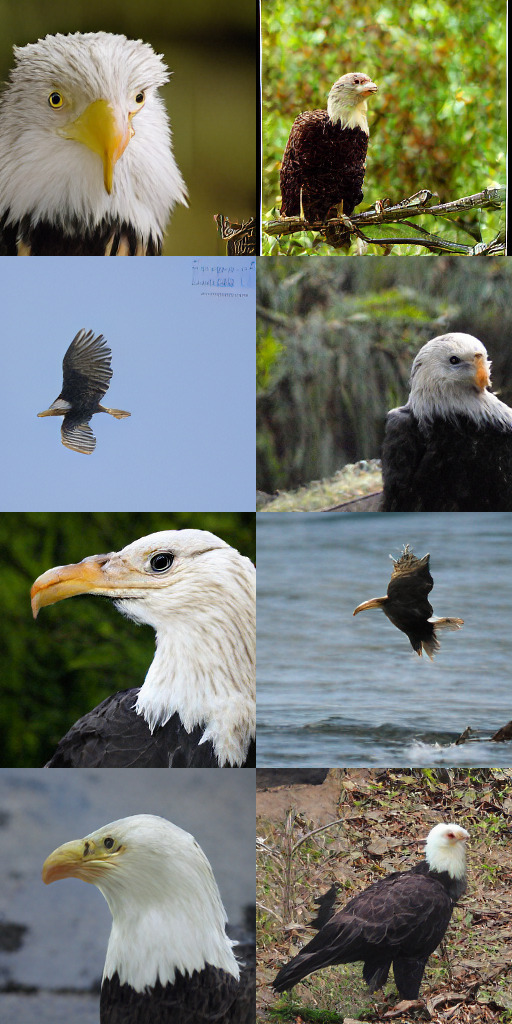}
         \caption{VQGAN.}
     \end{subfigure}
     \begin{subfigure}[b]{0.19\textwidth}
         \centering
         \includegraphics[width=\textwidth]{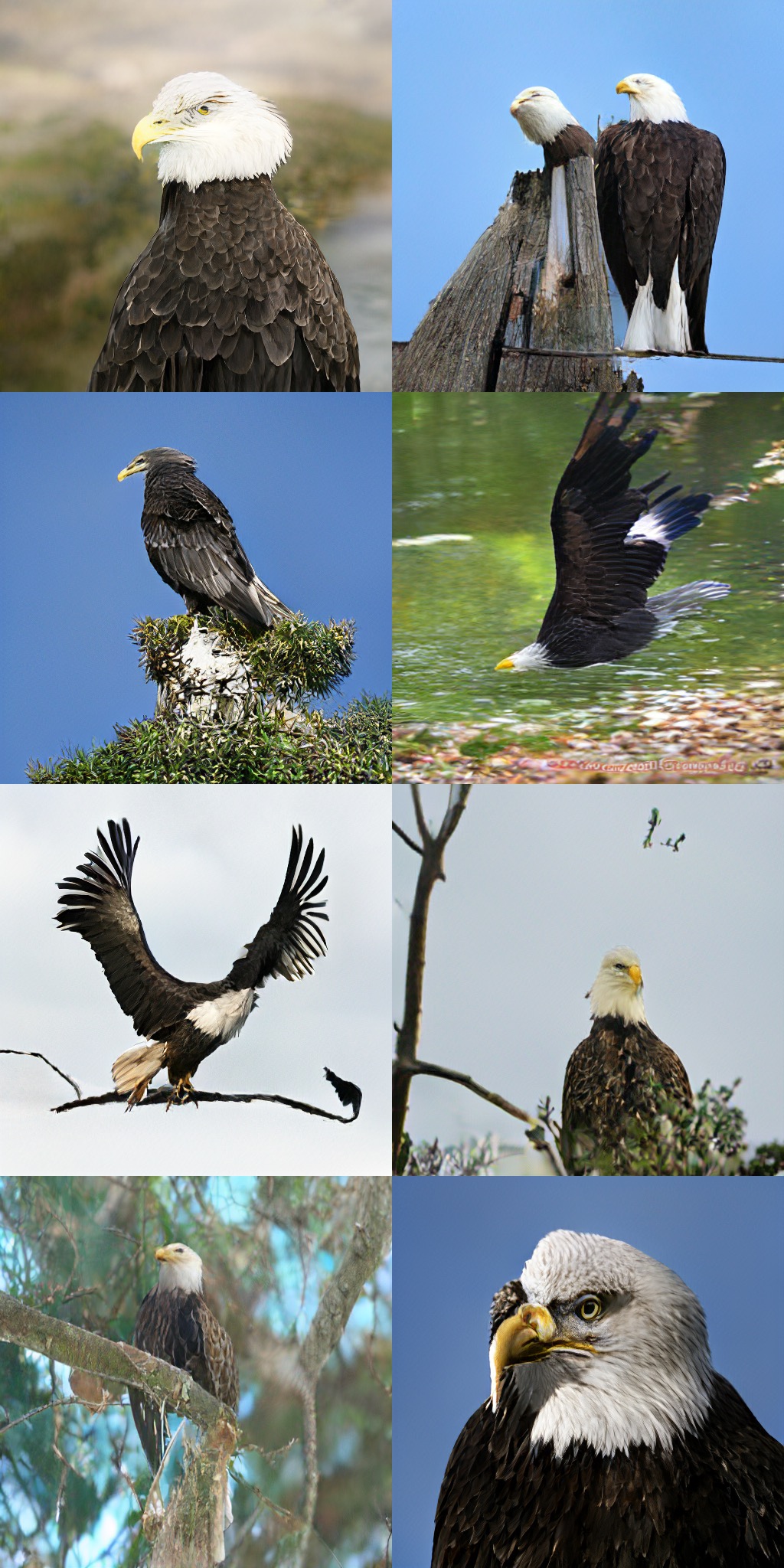}
         \caption{MaskGIT.}
     \end{subfigure}
     \begin{subfigure}[b]{0.19\textwidth}
         \centering
         \includegraphics[width=\textwidth]{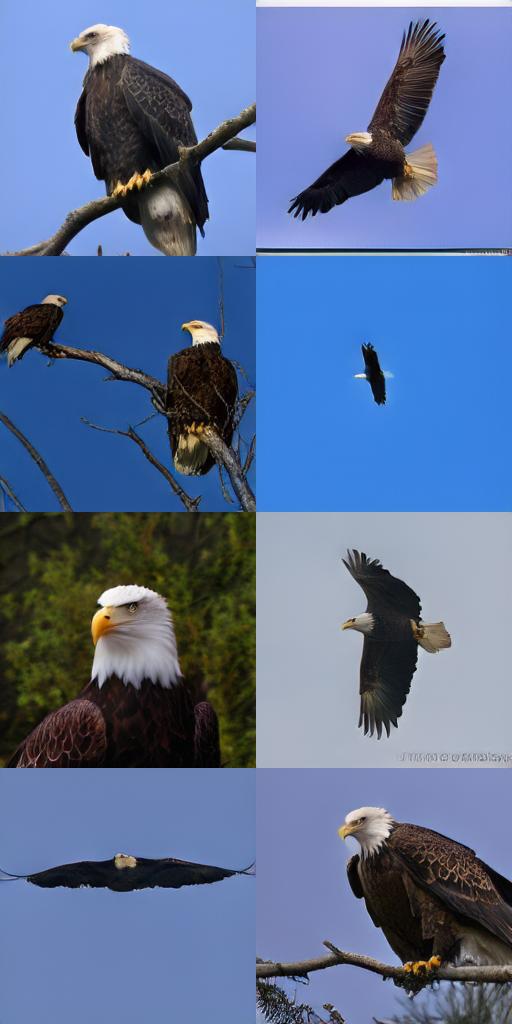}
         \caption{Ours.}
     \end{subfigure}
\end{figure}

\begin{figure}[t]
    \centering
     \begin{subfigure}[b]{0.19\textwidth}
         \centering
         \includegraphics[width=\textwidth]{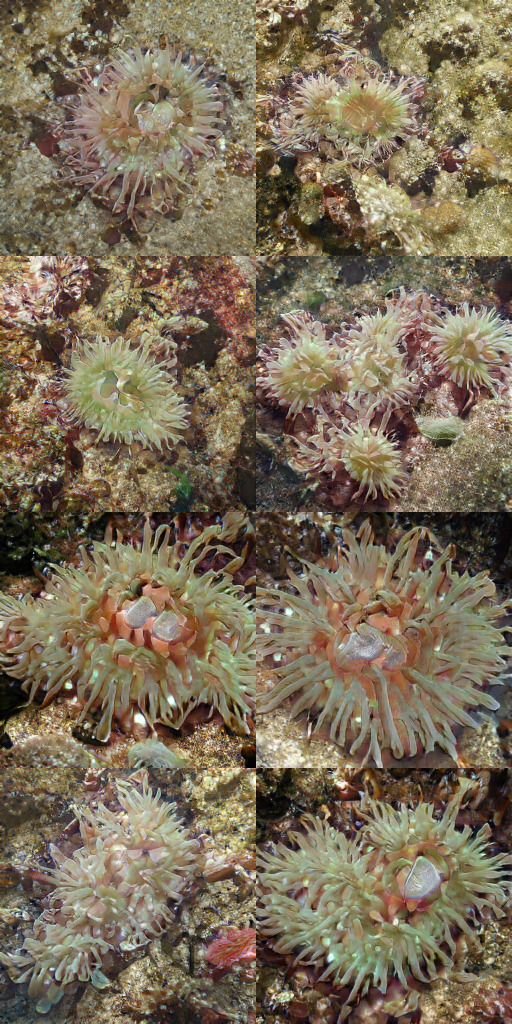}
         \caption{BigGAN-deep.}
     \end{subfigure}
     \begin{subfigure}[b]{0.19\textwidth}
         \centering
         \includegraphics[width=\textwidth]{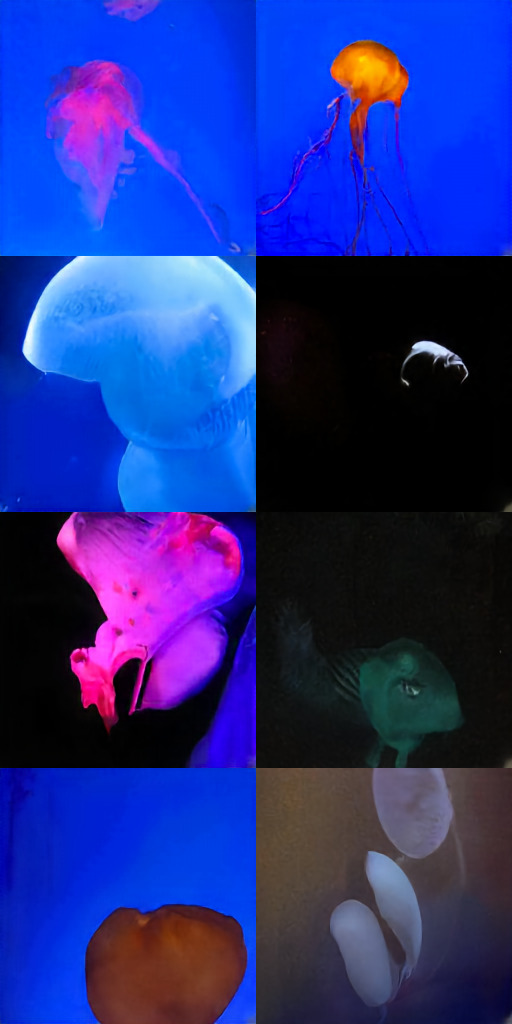}
         \caption{VQ-VAE-2.}
     \end{subfigure}
     \begin{subfigure}[b]{0.19\textwidth}
         \centering
         \includegraphics[width=\textwidth]{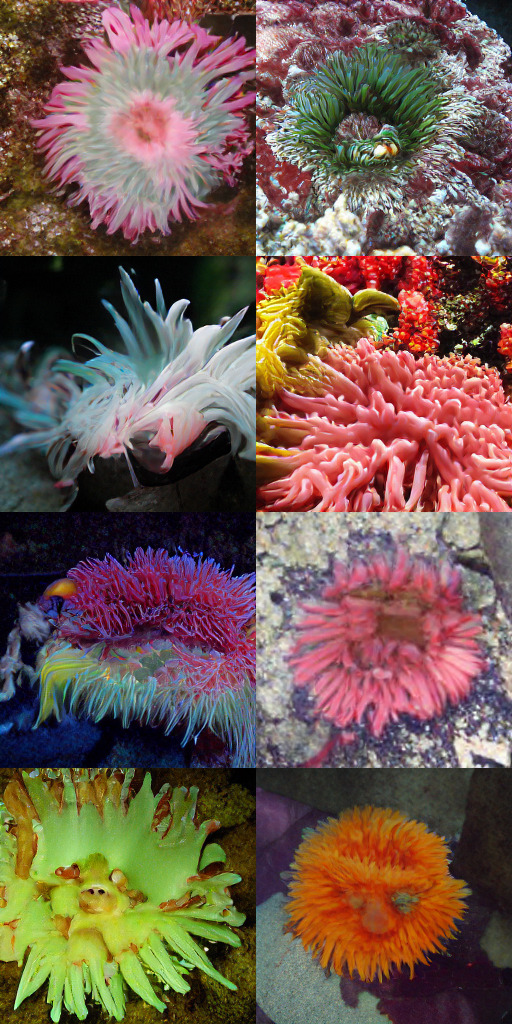}
         \caption{VQGAN.}
     \end{subfigure}
     \begin{subfigure}[b]{0.19\textwidth}
         \centering
         \includegraphics[width=\textwidth]{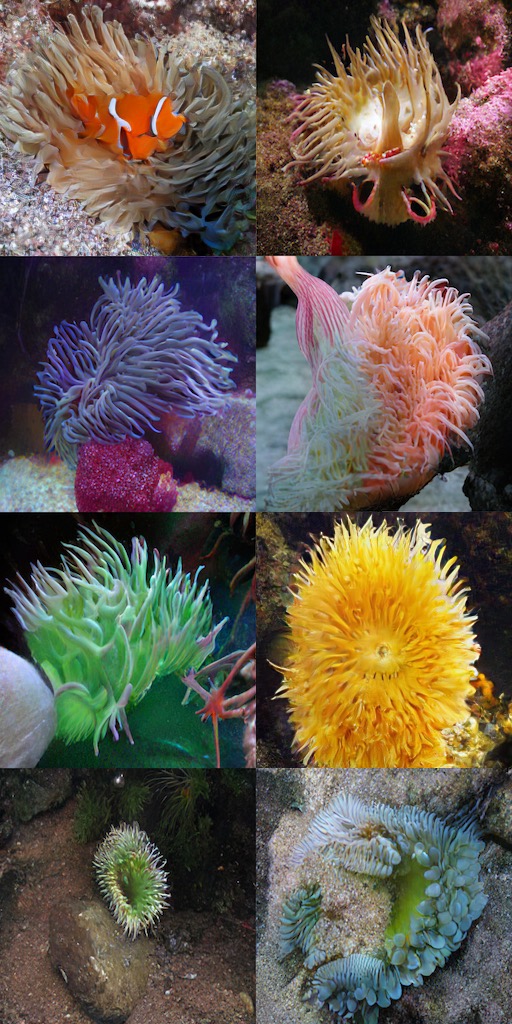}
         \caption{MaskGIT.}
     \end{subfigure}
     \begin{subfigure}[b]{0.19\textwidth}
         \centering
         \includegraphics[width=\textwidth]{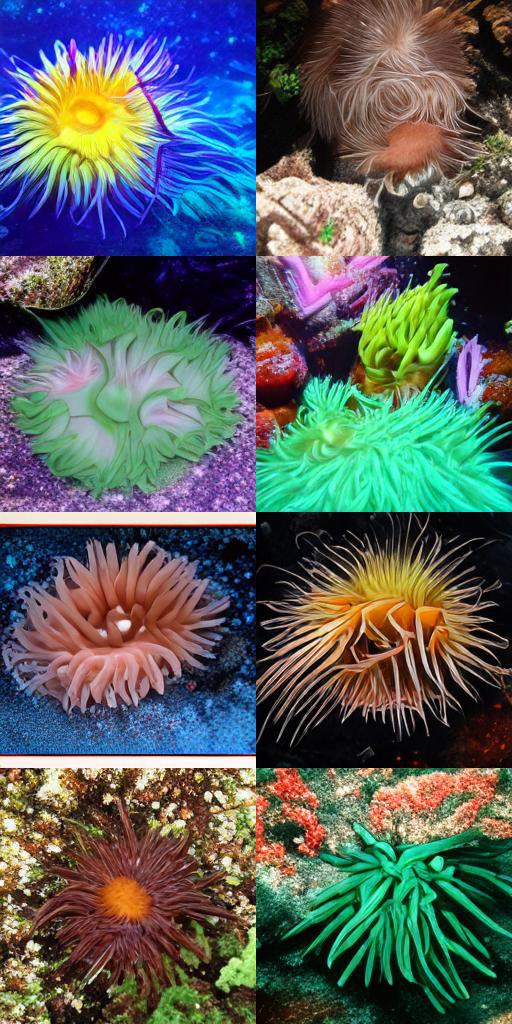}
         \caption{Ours.}
     \end{subfigure}
\end{figure}

\clearpage

\begin{figure}[t]
    \centering
     \begin{subfigure}[b]{0.19\textwidth}
         \centering
         \includegraphics[width=\textwidth]{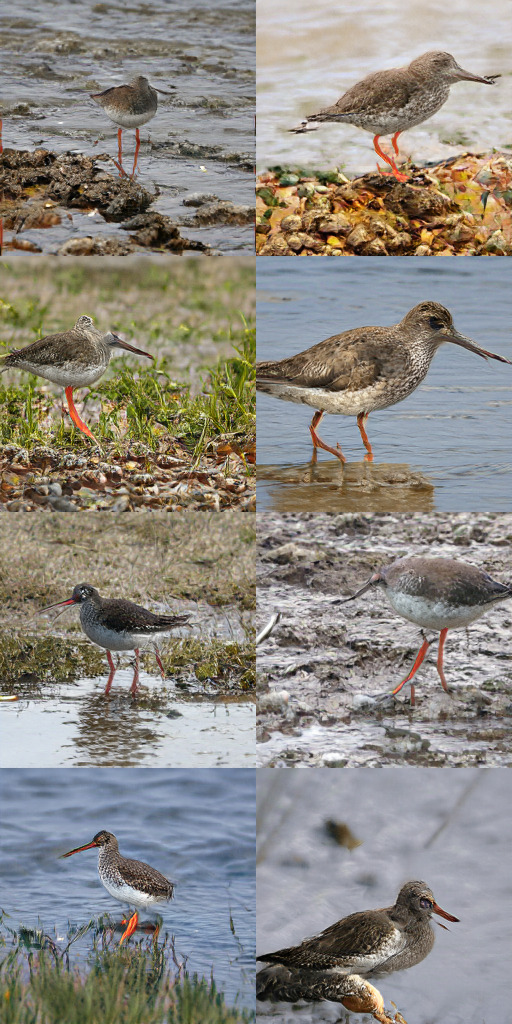}
         \caption{BigGAN-deep.}
     \end{subfigure}
     \begin{subfigure}[b]{0.19\textwidth}
         \centering
         \includegraphics[width=\textwidth]{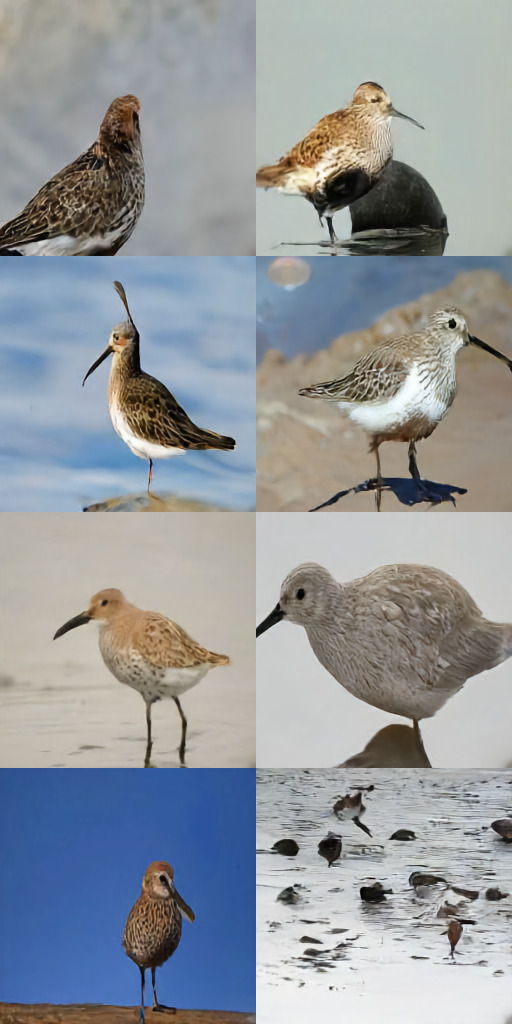}
         \caption{VQ-VAE-2.}
     \end{subfigure}
     \begin{subfigure}[b]{0.19\textwidth}
         \centering
         \includegraphics[width=\textwidth]{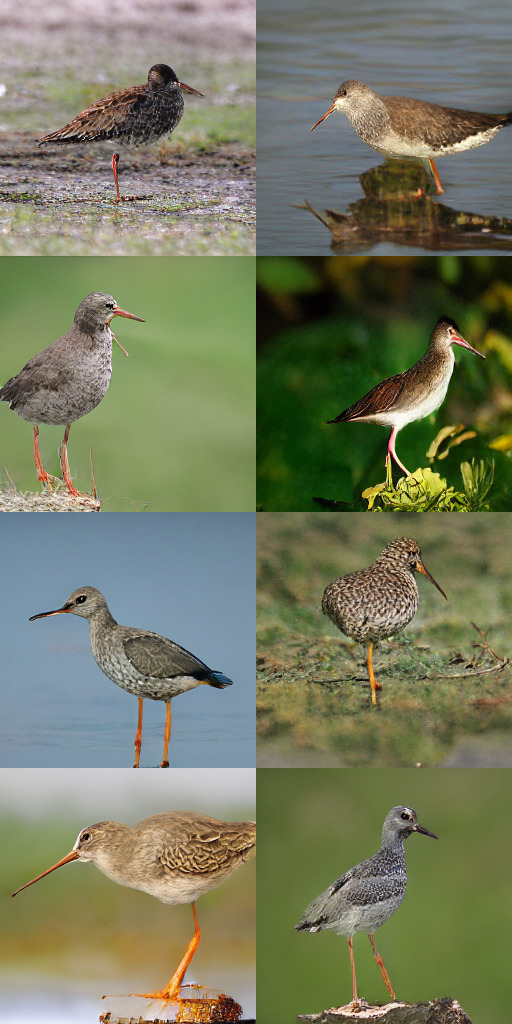}
         \caption{VQGAN.}
     \end{subfigure}
     \begin{subfigure}[b]{0.19\textwidth}
         \centering
         \includegraphics[width=\textwidth]{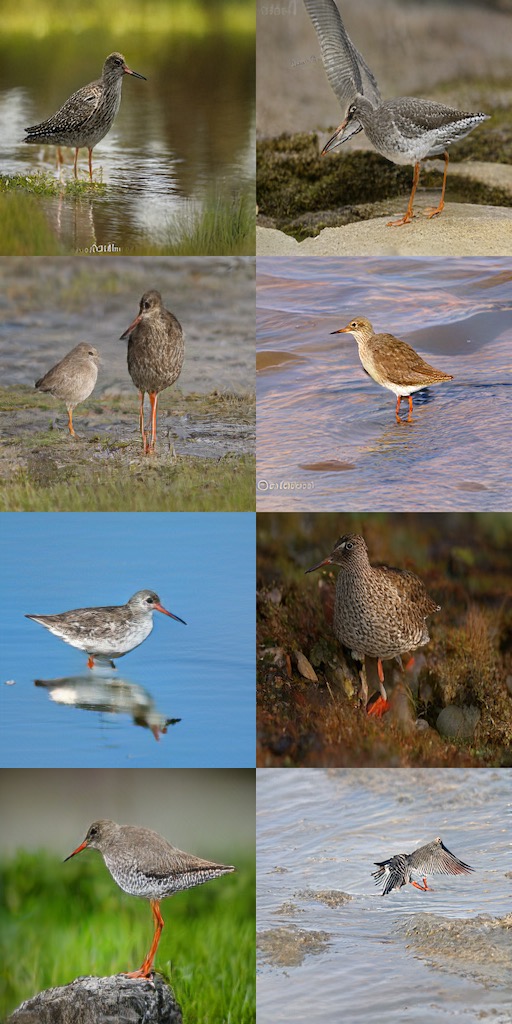}
         \caption{MaskGIT.}
     \end{subfigure}
     \begin{subfigure}[b]{0.19\textwidth}
         \centering
         \includegraphics[width=\textwidth]{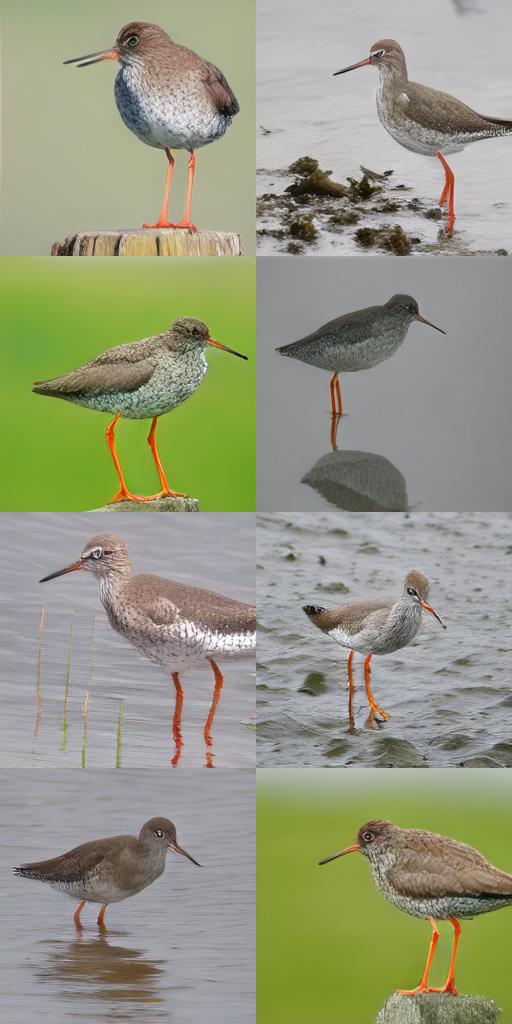}
         \caption{Ours.}
     \end{subfigure}
\end{figure}

\begin{figure}[t]
    \centering
     \begin{subfigure}[b]{0.19\textwidth}
         \centering
         \includegraphics[width=\textwidth]{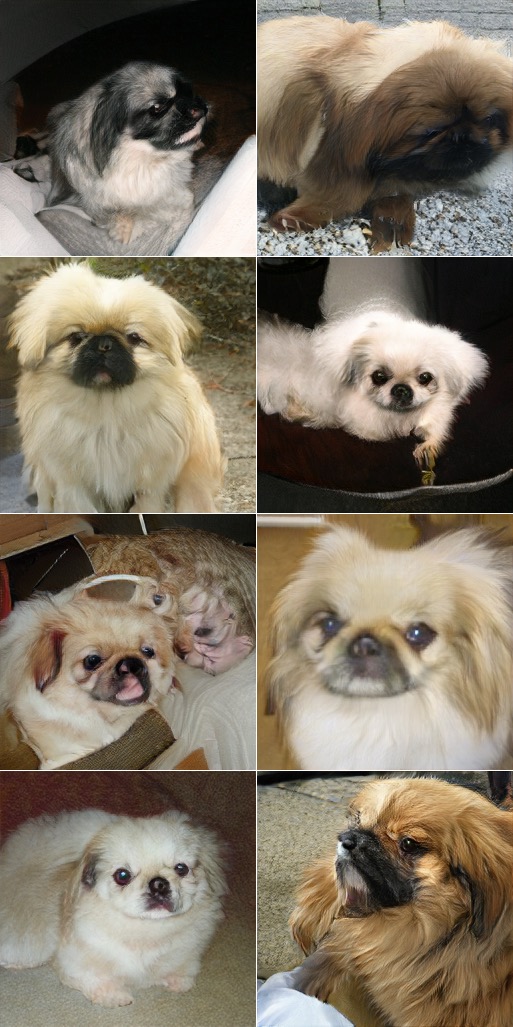}
         \caption{BigGAN-deep.}
     \end{subfigure}
     \begin{subfigure}[b]{0.19\textwidth}
         \centering
         \includegraphics[width=\textwidth]{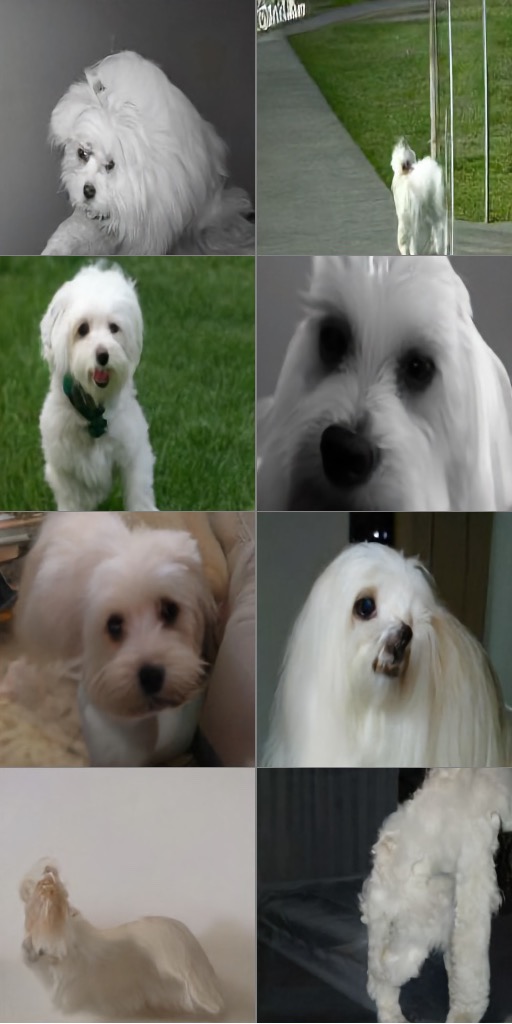}
         \caption{VQ-VAE-2.}
     \end{subfigure}
     \begin{subfigure}[b]{0.19\textwidth}
         \centering
         \includegraphics[width=\textwidth]{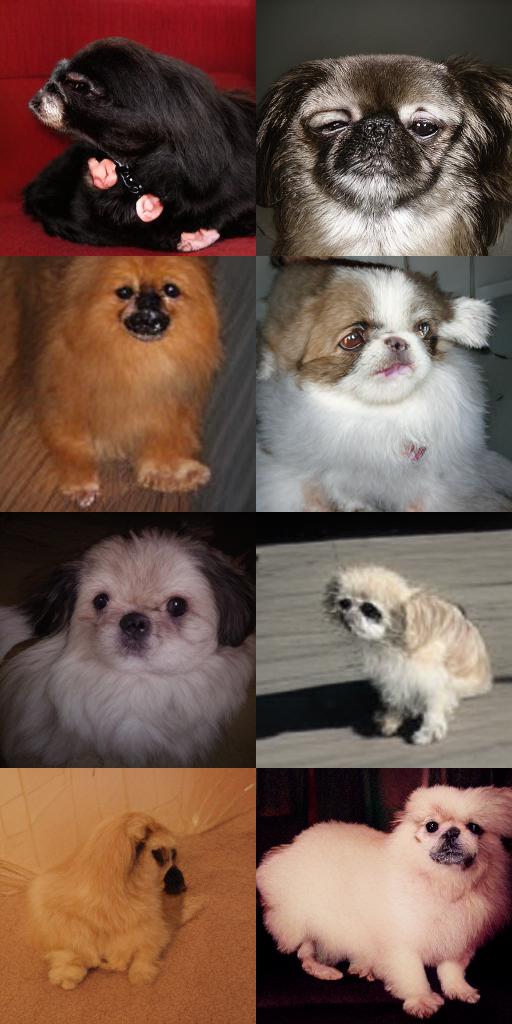}
         \caption{VQGAN.}
     \end{subfigure}
     \begin{subfigure}[b]{0.19\textwidth}
         \centering
         \includegraphics[width=\textwidth]{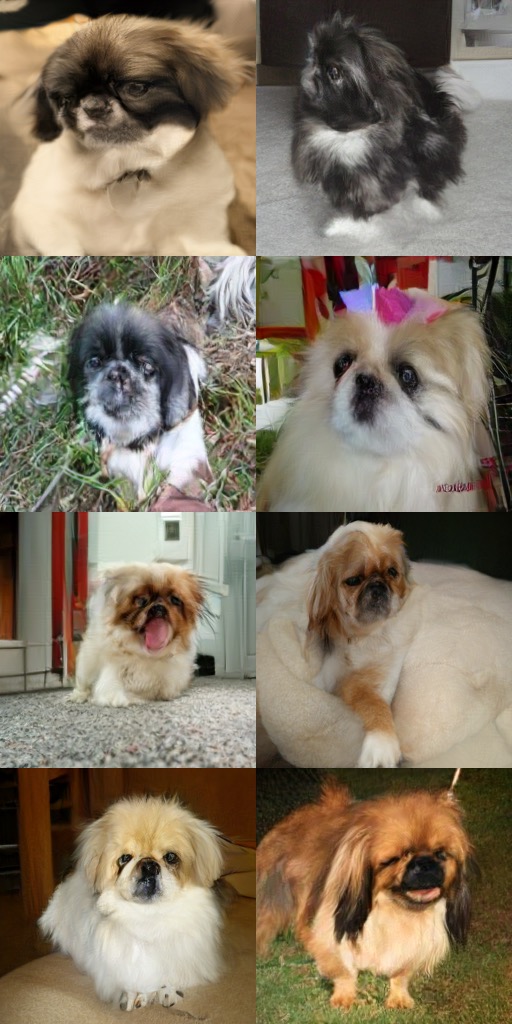}
         \caption{MaskGIT.}
     \end{subfigure}
     \begin{subfigure}[b]{0.19\textwidth}
         \centering
         \includegraphics[width=\textwidth]{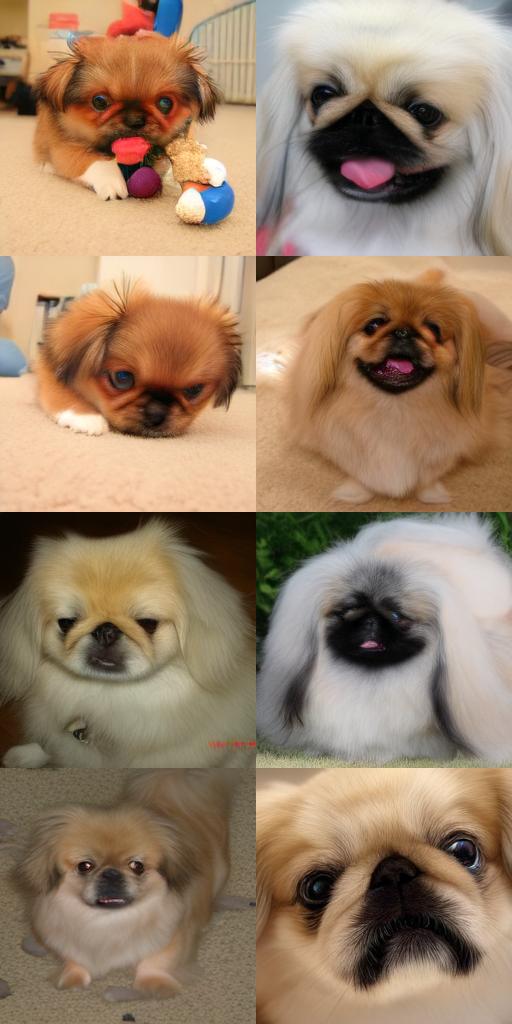}
         \caption{Ours.}
     \end{subfigure}
\end{figure}

\clearpage

\newcommand{\TS}{1.0\linewidth}
\begin{figure}[t]
	\centering
	\resizebox{\linewidth}{!}{
		\begin{tabular}{c|c|c|c} 
		\includegraphics[height=\TS]{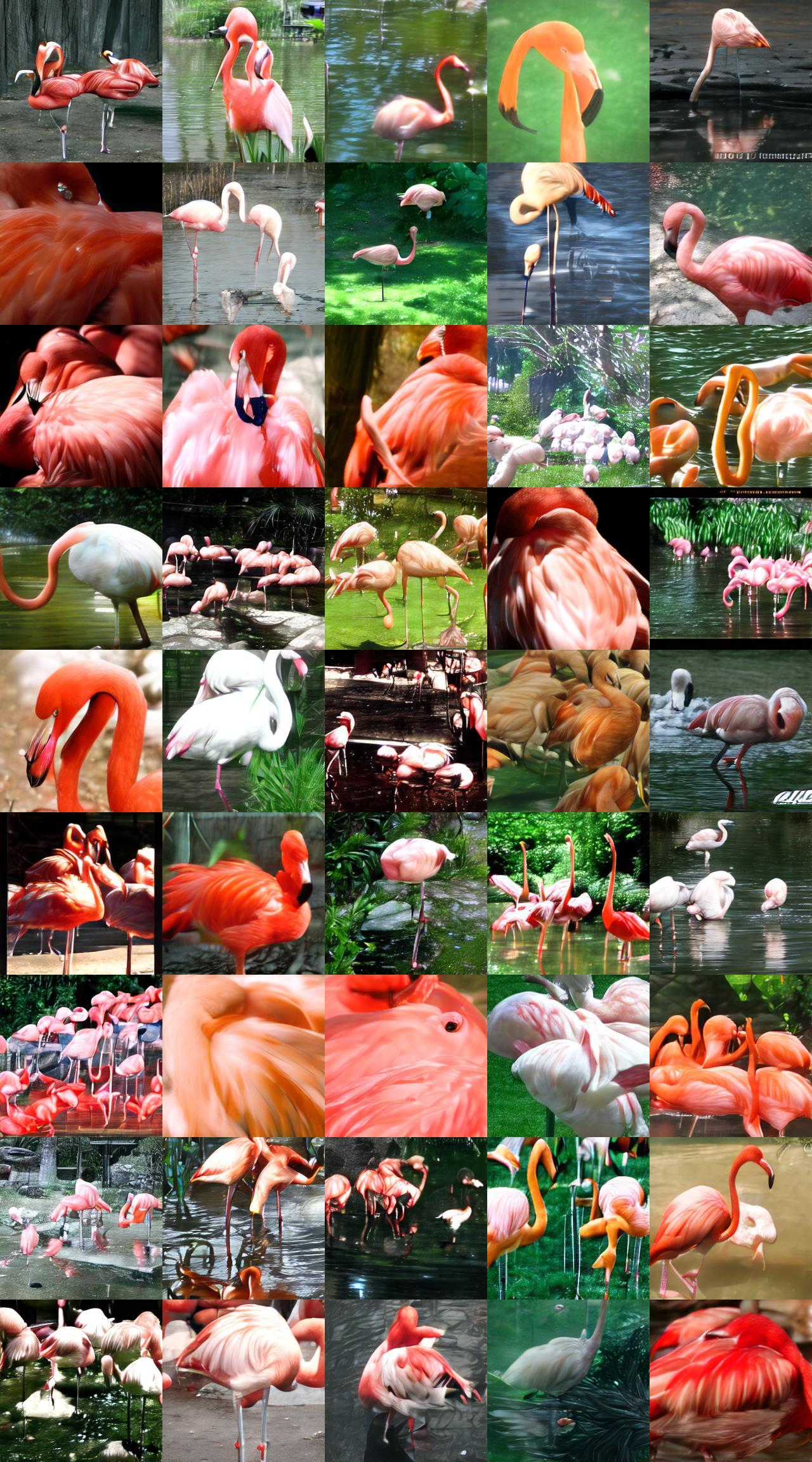} &
        \includegraphics[height=\TS]{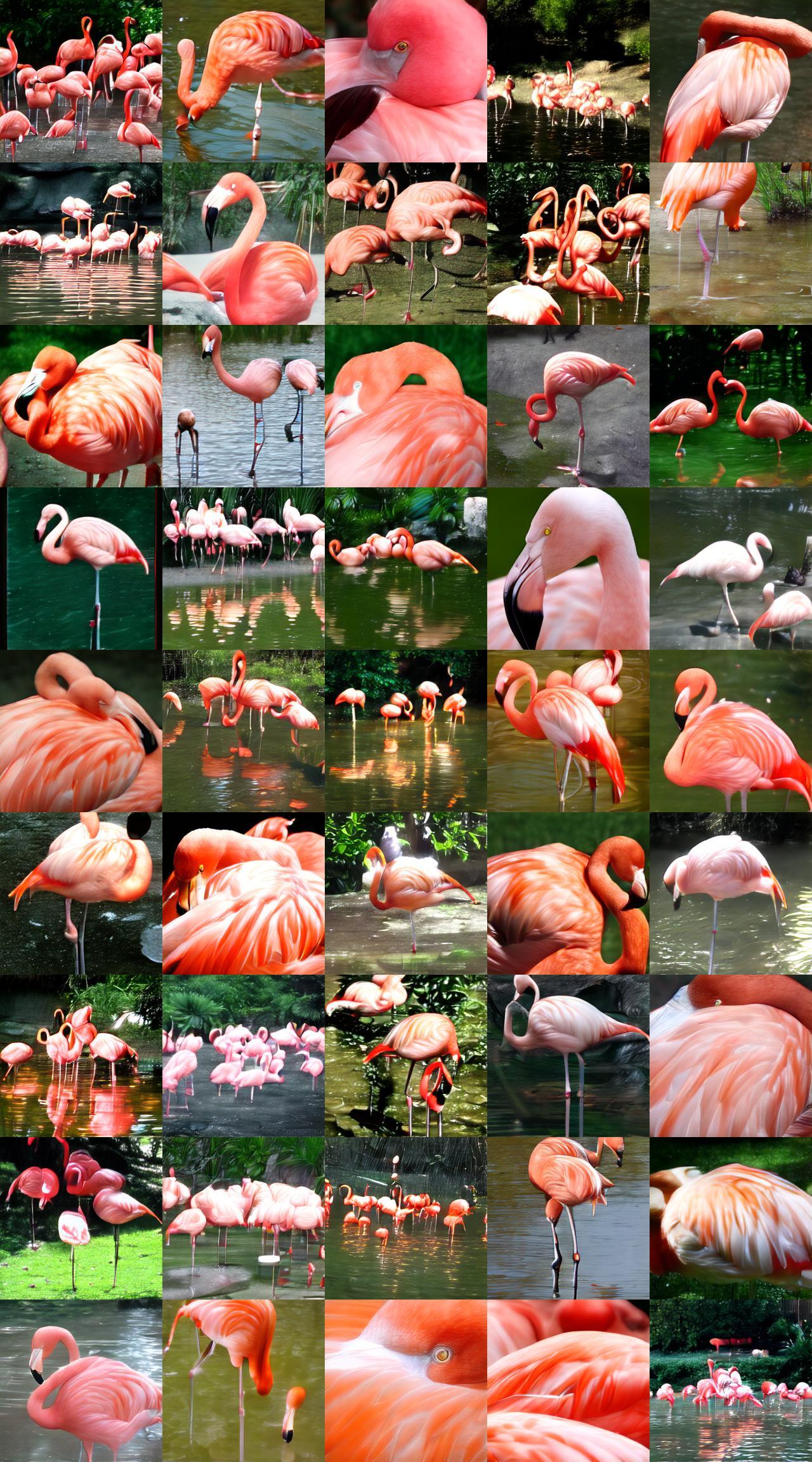} &
        \includegraphics[height=\TS]{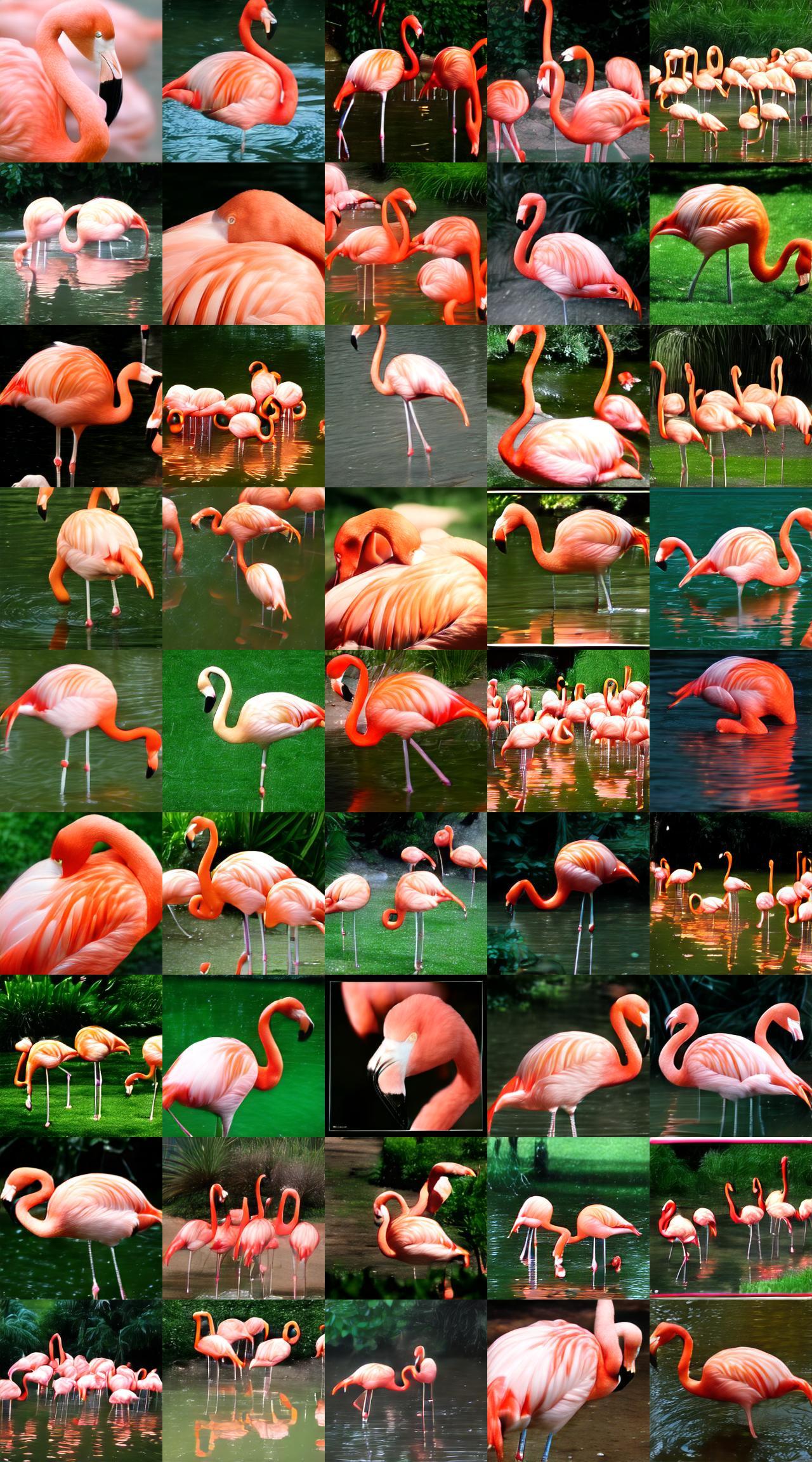} &
        \includegraphics[height=\TS]{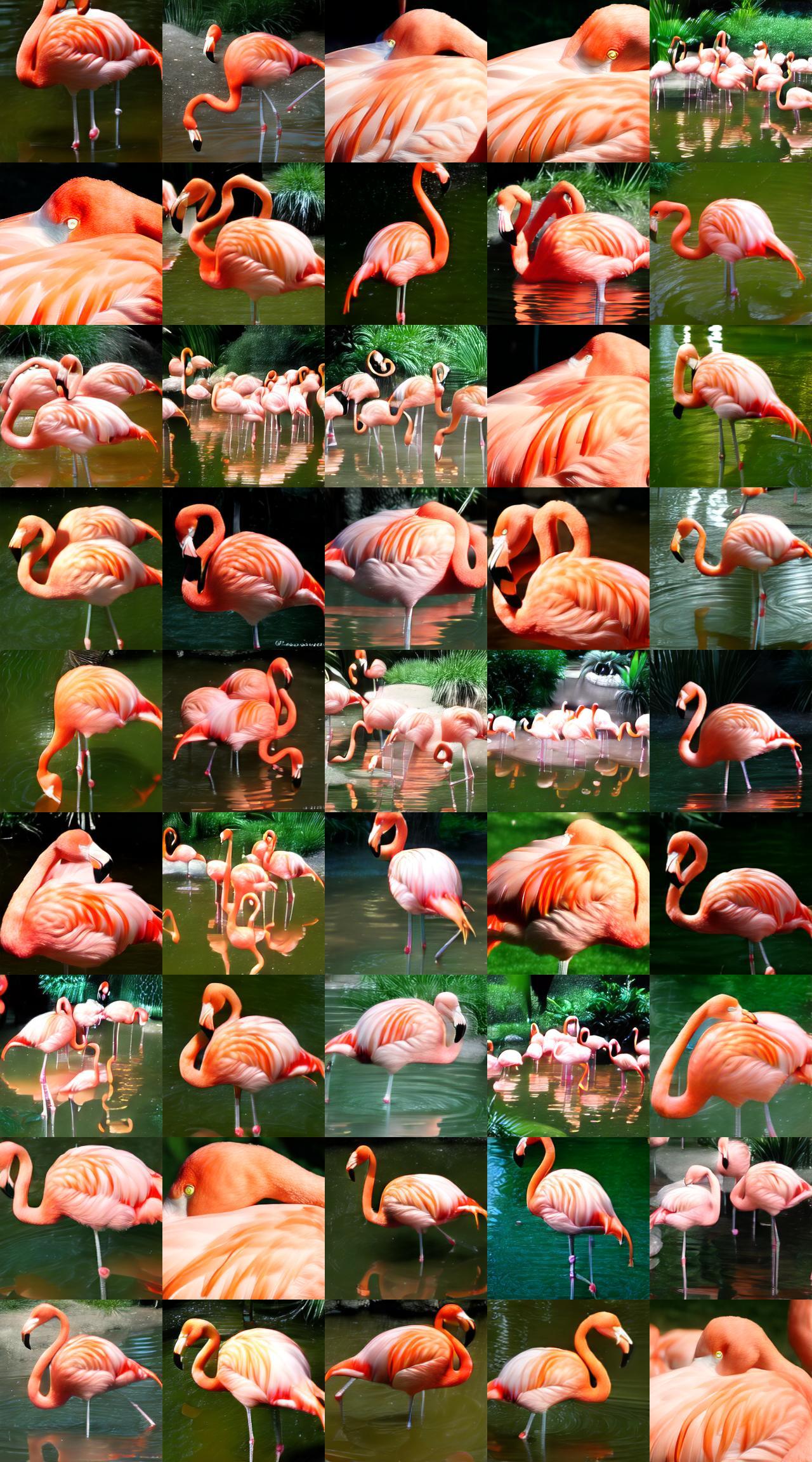} \\
        \includegraphics[height=\TS]{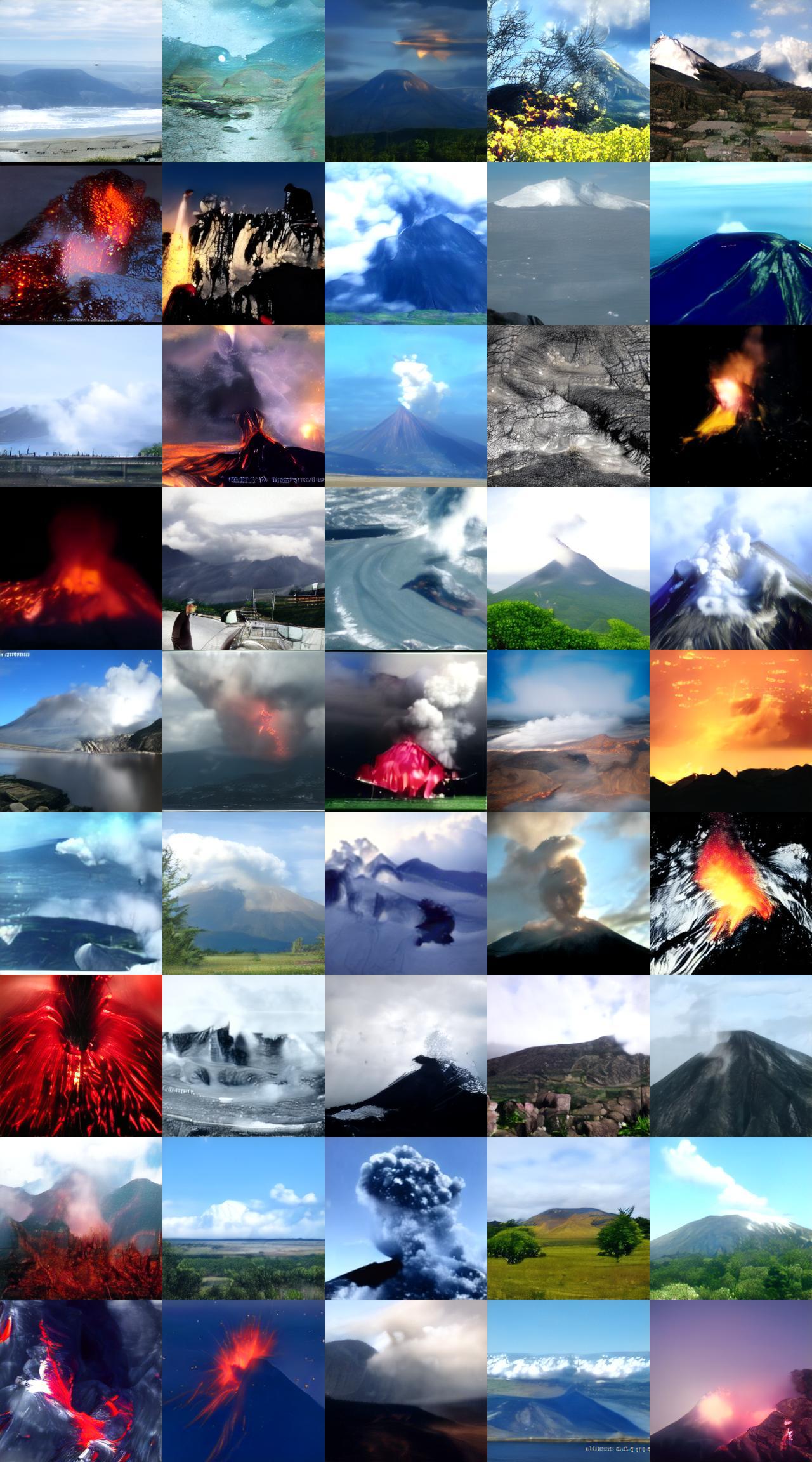} &
        \includegraphics[height=\TS]{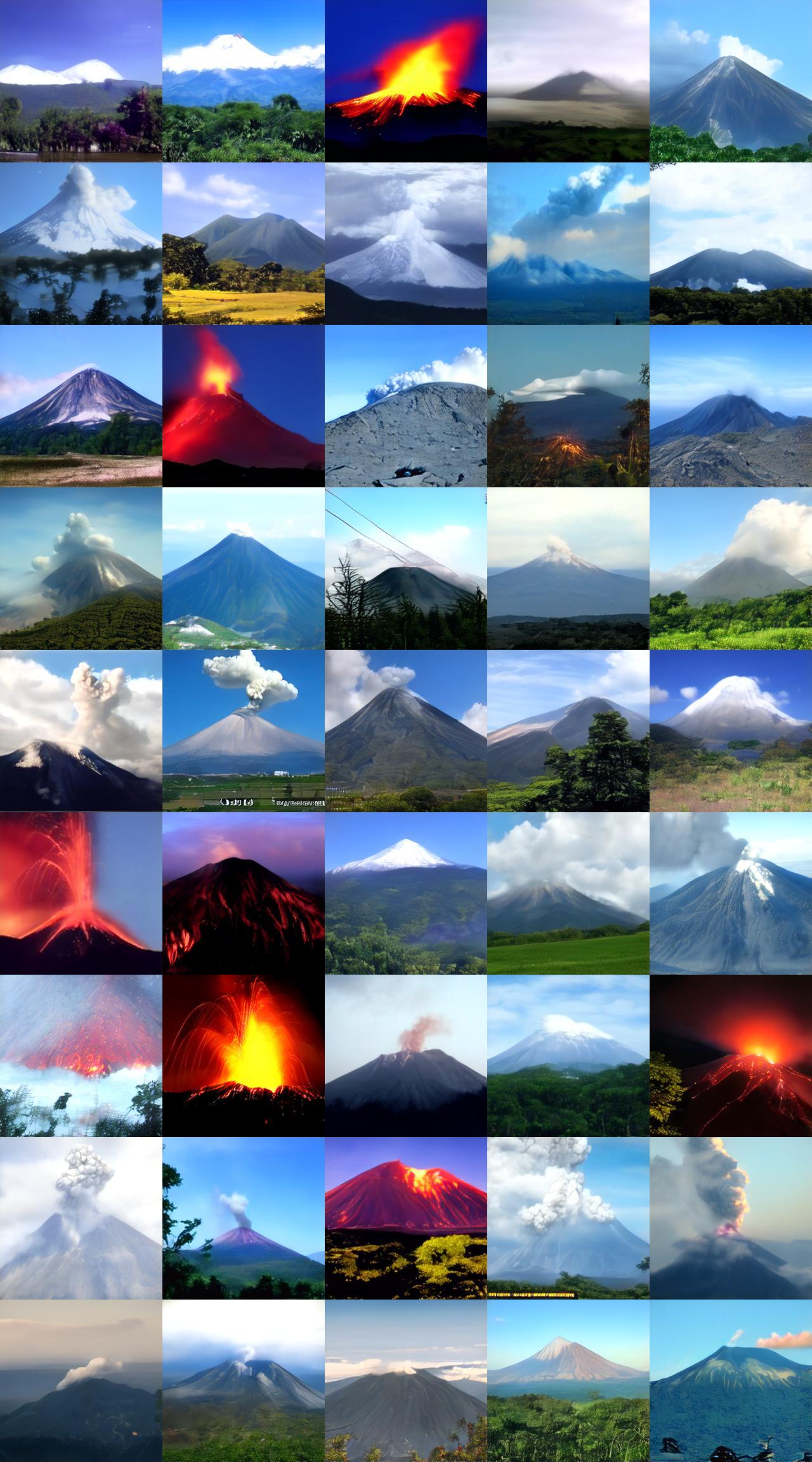} &
        \includegraphics[height=\TS]{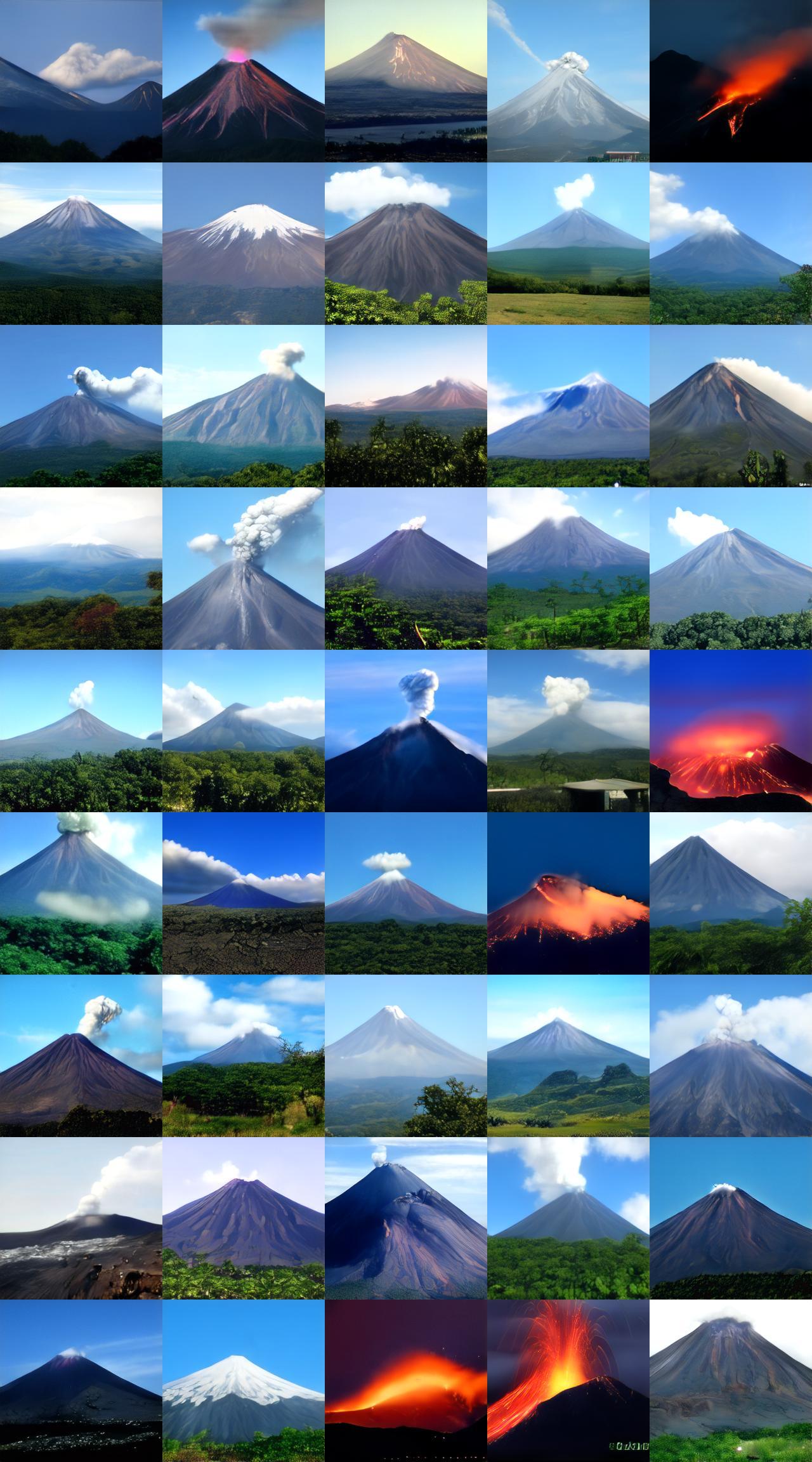} &
        \includegraphics[height=\TS]{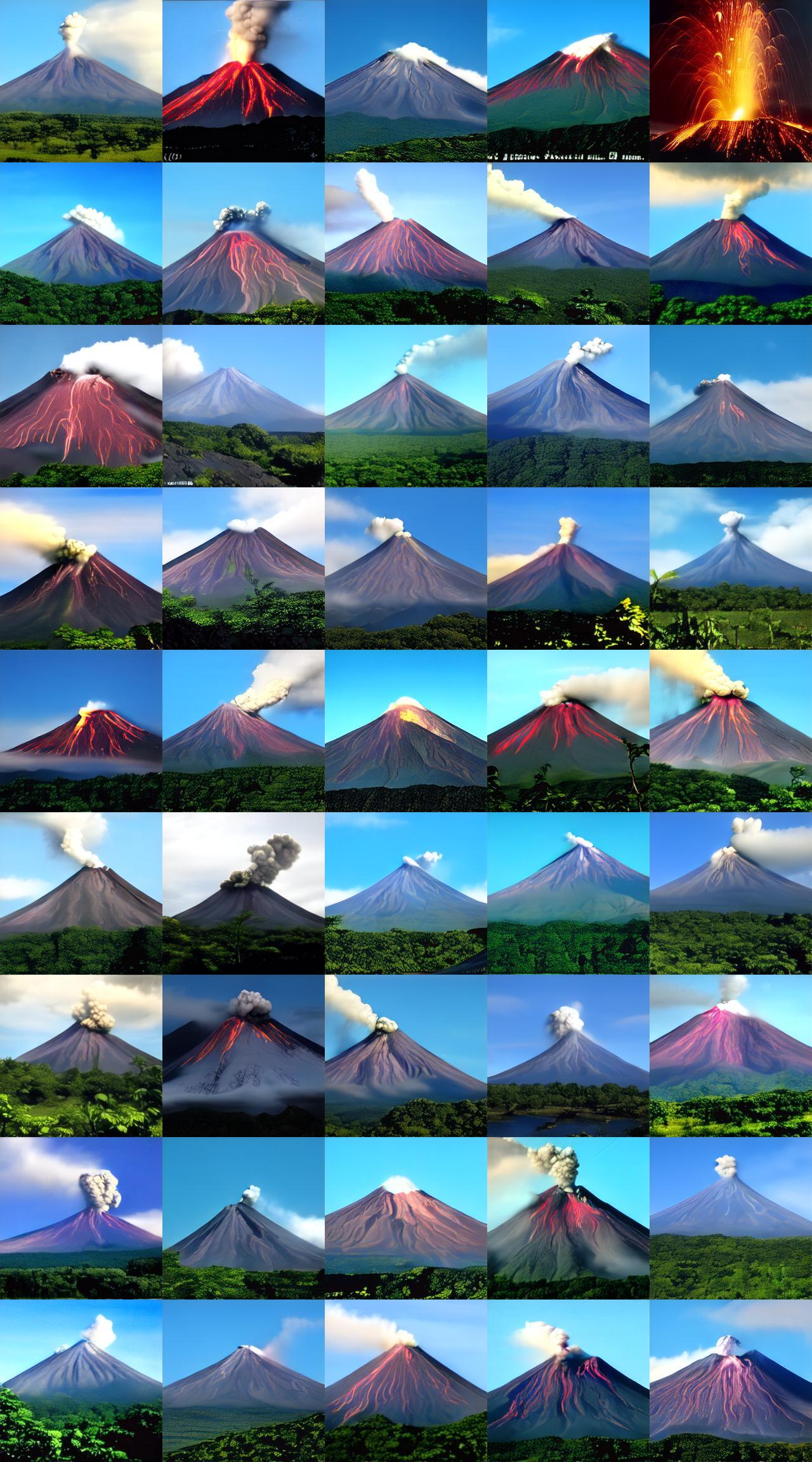} \\
        \includegraphics[height=\TS]{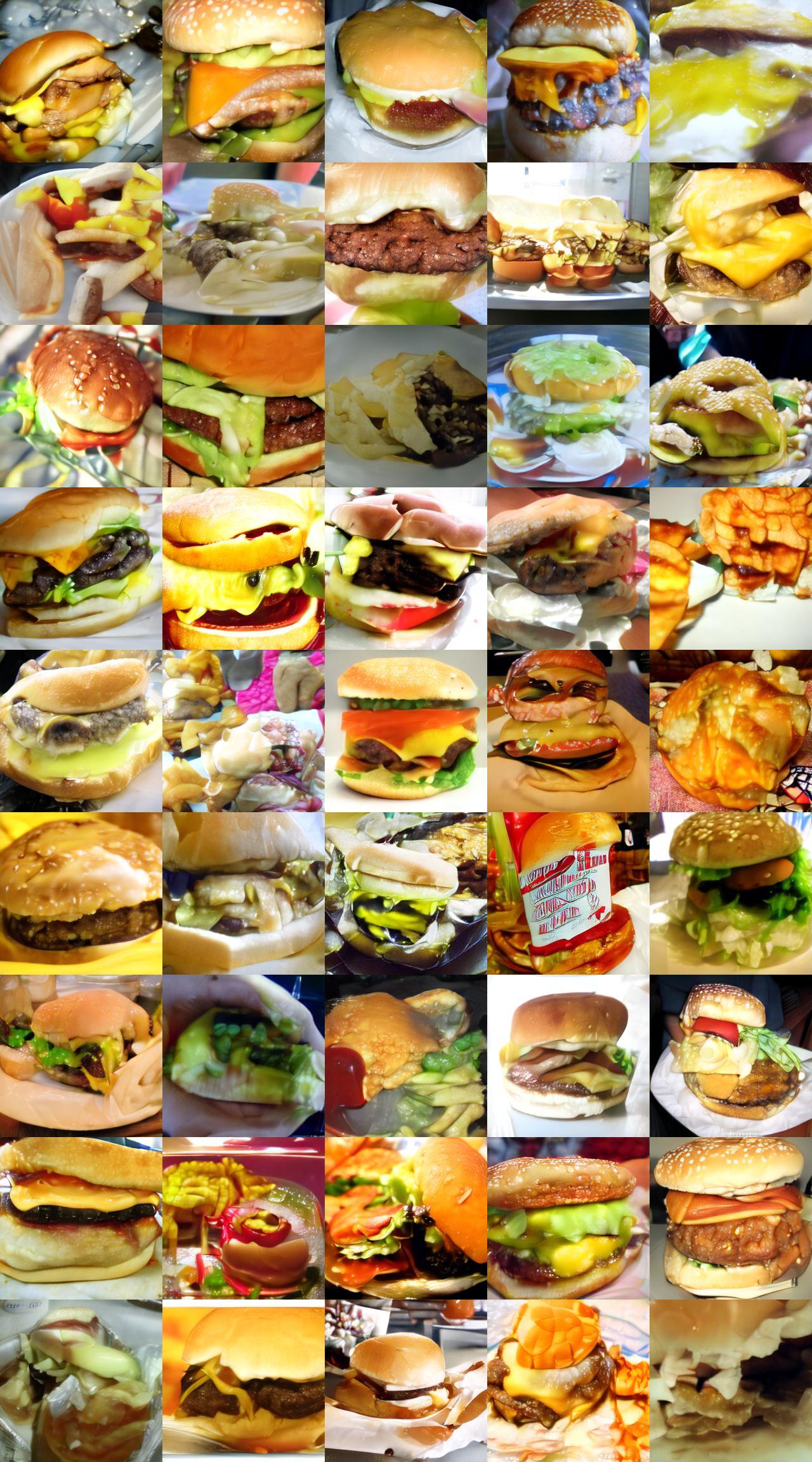} &
        \includegraphics[height=\TS]{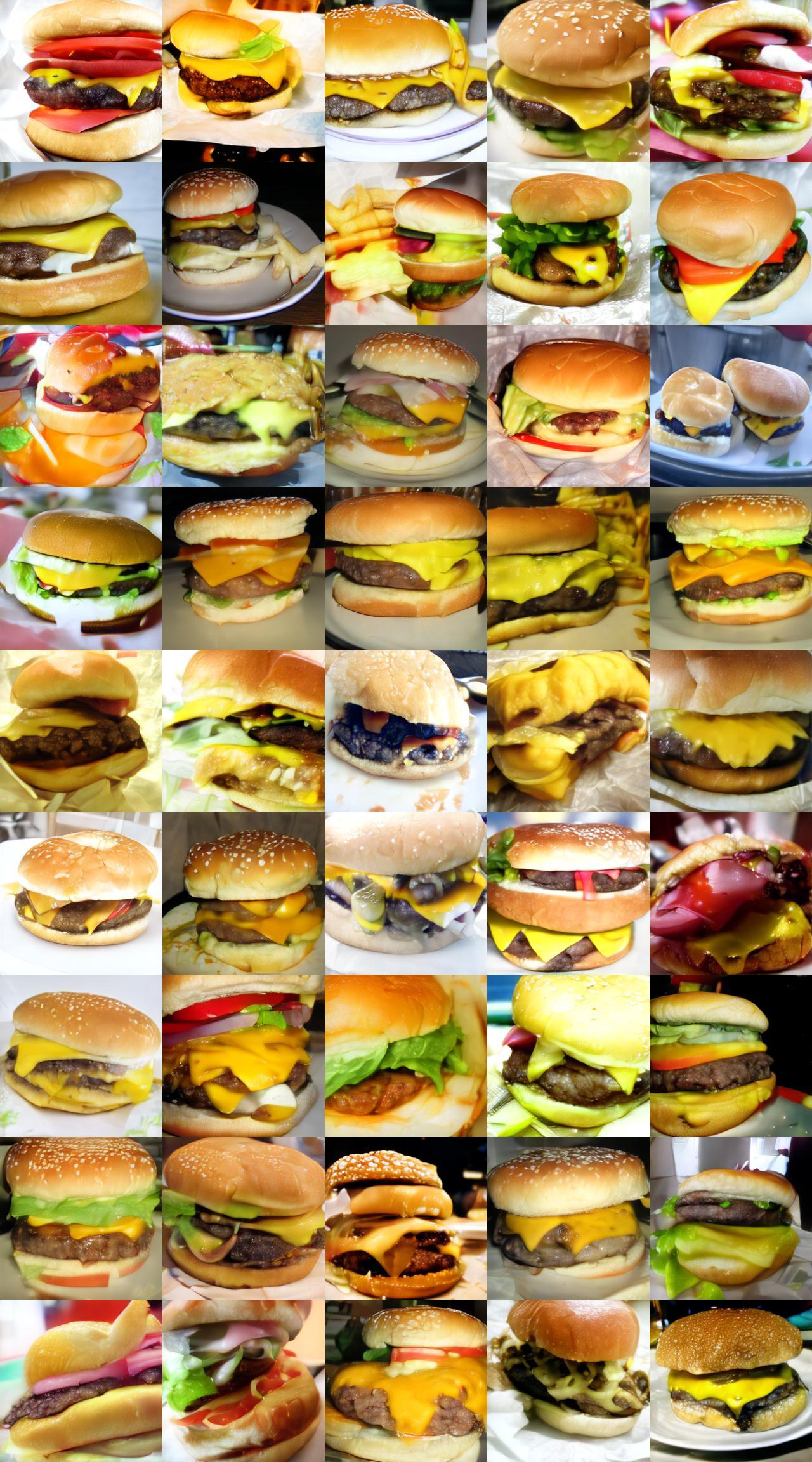} &
        \includegraphics[height=\TS]{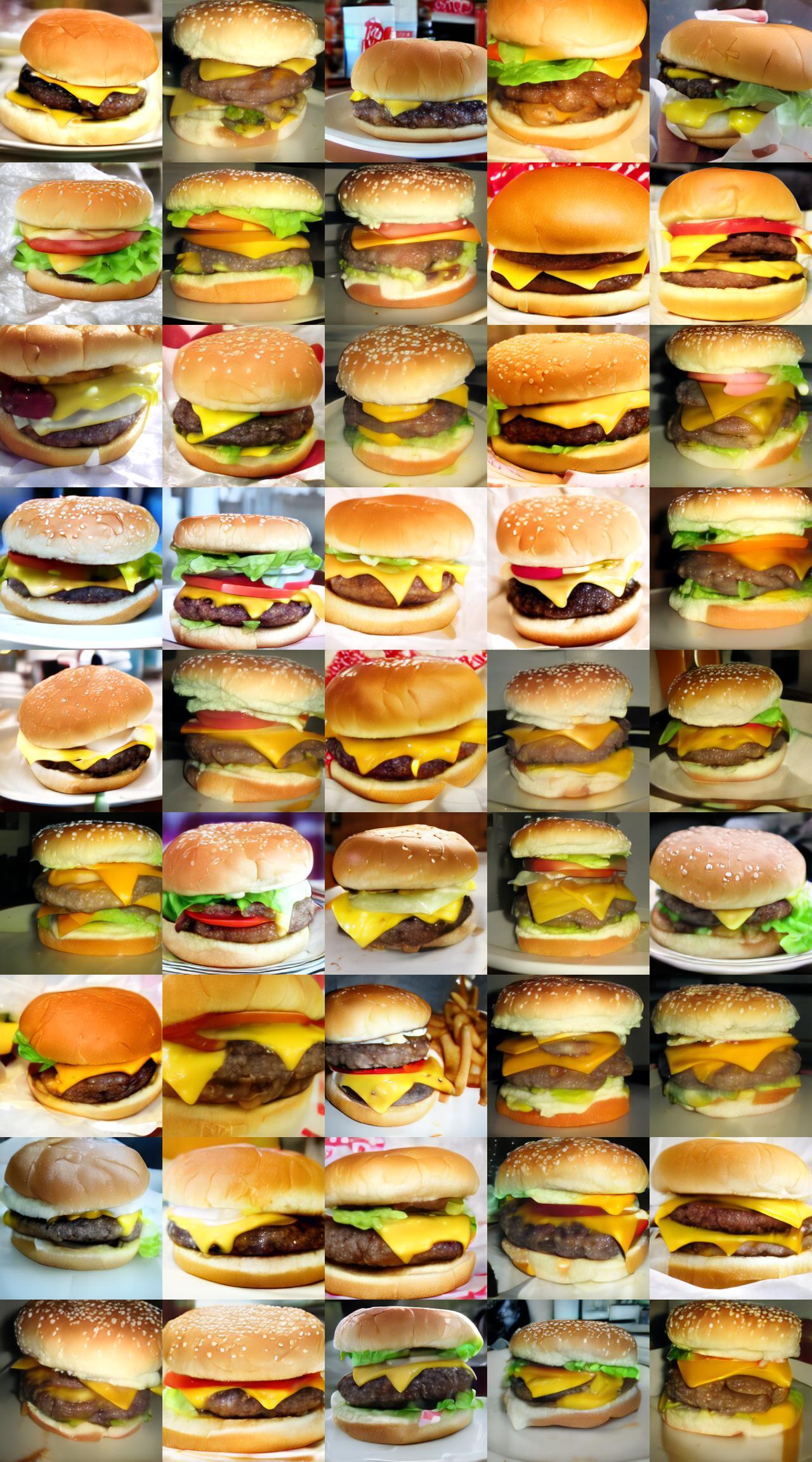} &
        \includegraphics[height=\TS]{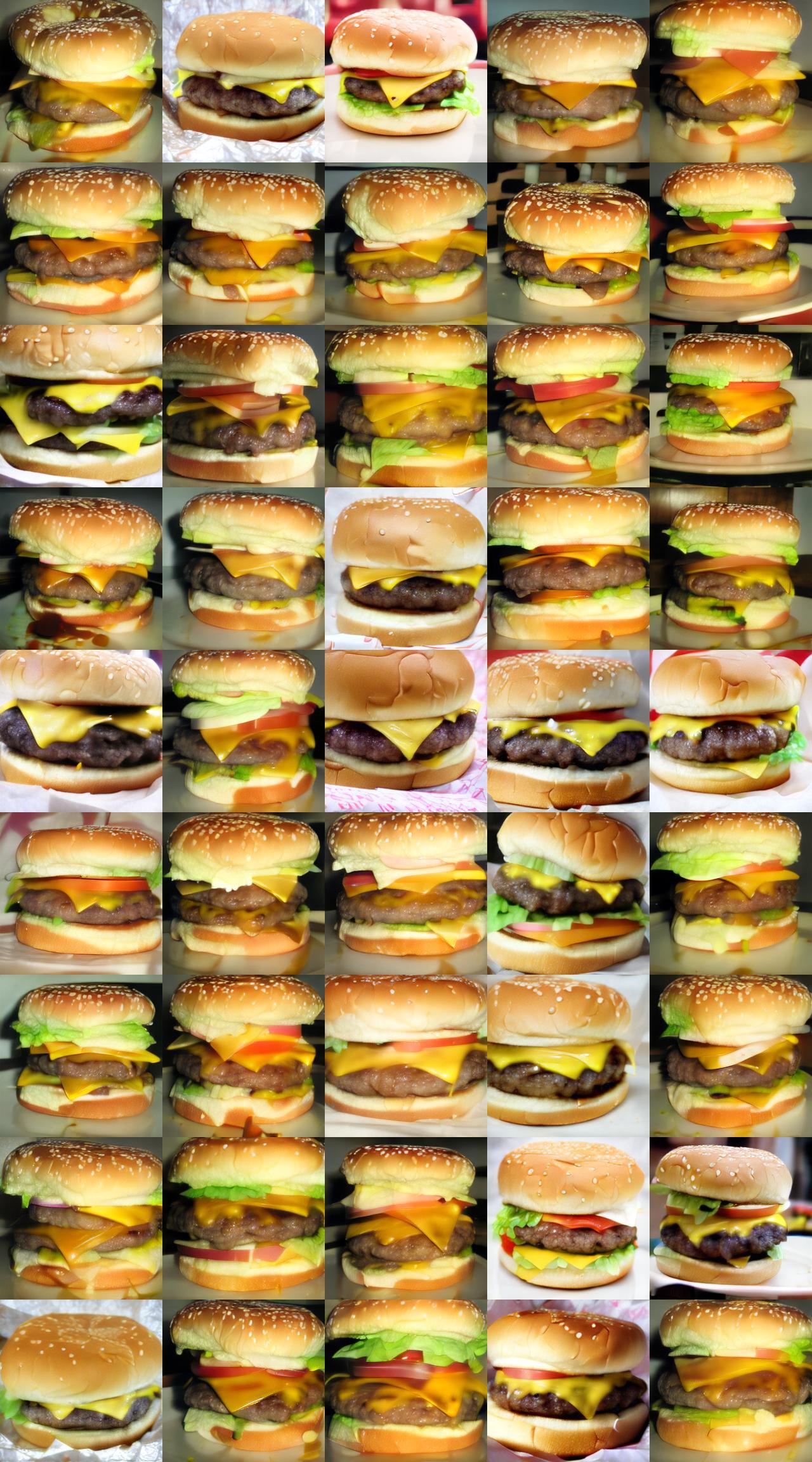} \\
        
        \Huge $\omega = 0.0$ & \Huge $\omega = 1.0$ & \Huge $\omega = 2.5$ & \Huge $\omega = 10.0$ \\
	\end{tabular}}
	\caption{Label-conditioned image generation with different scales of classifier-free guidance ($\omega$). Larger $\omega$ improves the sample quality at the cost of lower diversity. }
	\label{fig:app_omega}
\end{figure}

\newcommand{\TSA}{0.25\linewidth}
\begin{figure}[t]
	\centering
	\resizebox{\linewidth}{!}{
		\begin{tabular}{c|c|c}
        \includegraphics[height=\TSA]{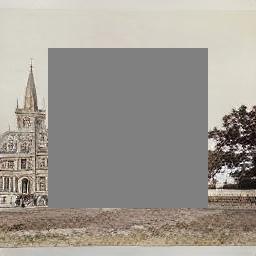} & 
		\includegraphics[height=\TSA]{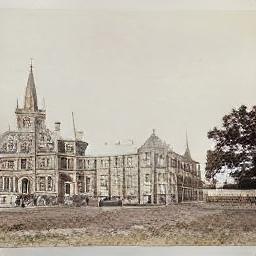} & 
		\includegraphics[height=\TSA]{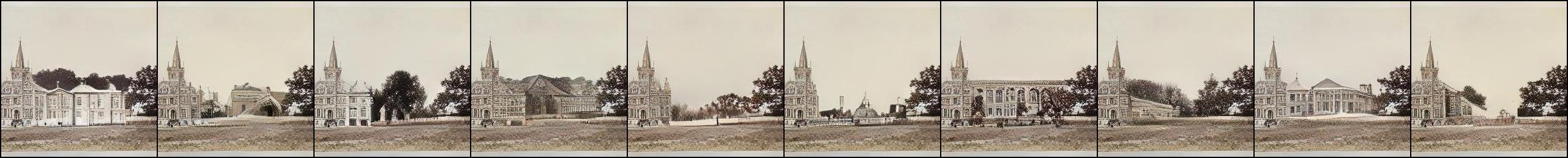} \\
		\includegraphics[height=\TSA]{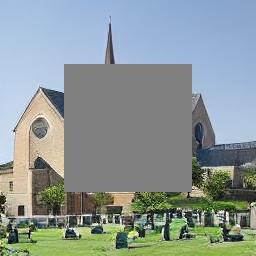} & 
		\includegraphics[height=\TSA]{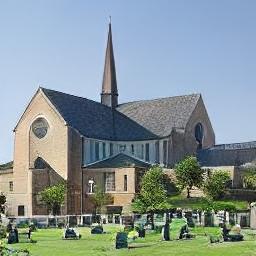} & 
		\includegraphics[height=\TSA]{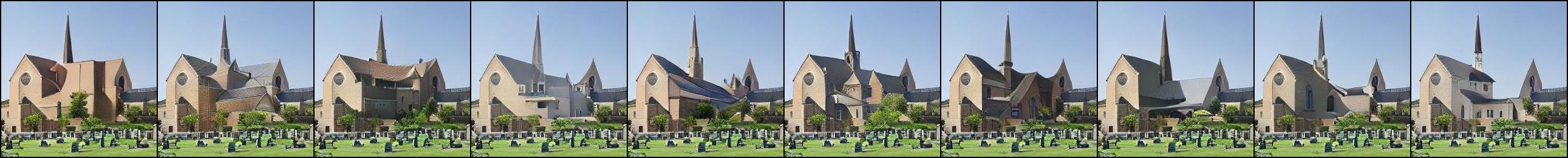} \\
		\includegraphics[height=\TSA]{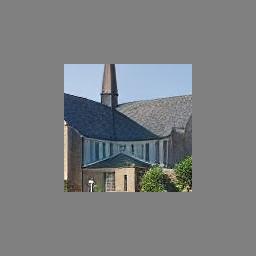} & 
		\includegraphics[height=\TSA]{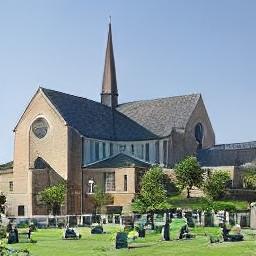} & 
		\includegraphics[height=\TSA]{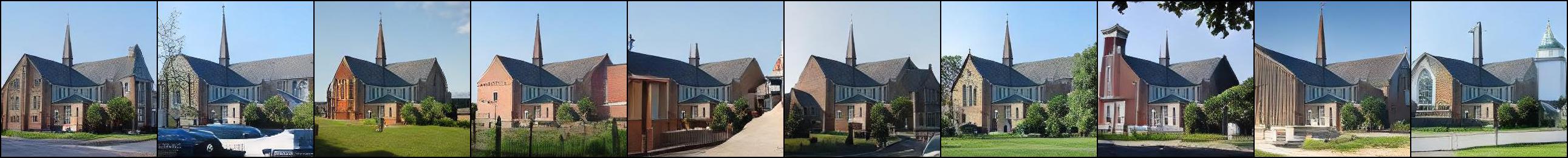} \\
		\includegraphics[height=\TSA]{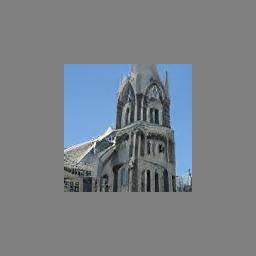} & 
		\includegraphics[height=\TSA]{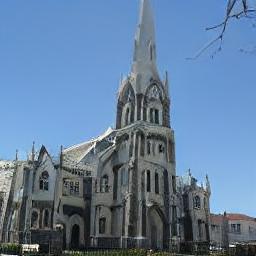} & 
		\includegraphics[height=\TSA]{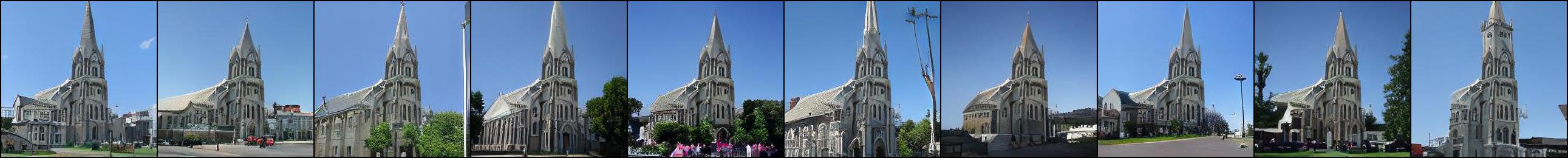} \\
		\includegraphics[height=\TSA]{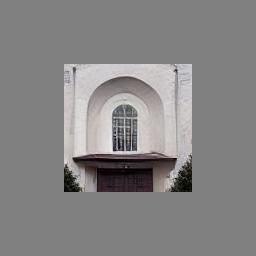} & 
		\includegraphics[height=\TSA]{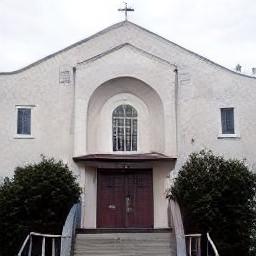} & 
		\includegraphics[height=\TSA]{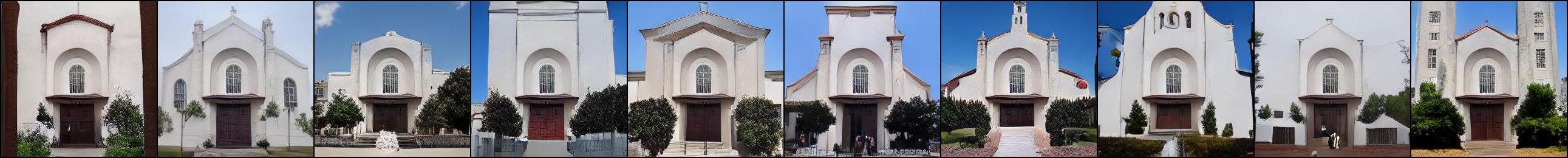} \\
		\includegraphics[height=\TSA]{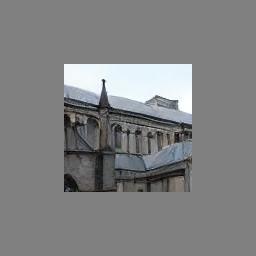} & 
		\includegraphics[height=\TSA]{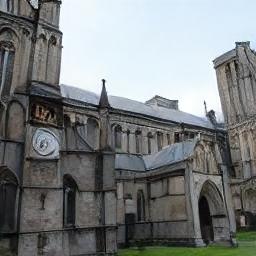} & 
		\includegraphics[height=\TSA]{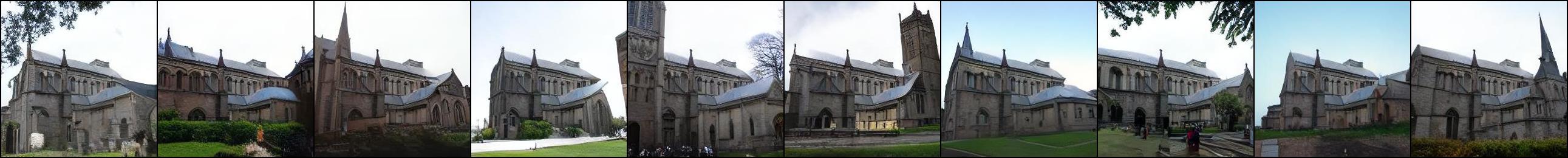} \\
		\includegraphics[height=\TSA]{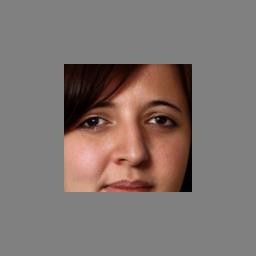} & 
		\includegraphics[height=\TSA]{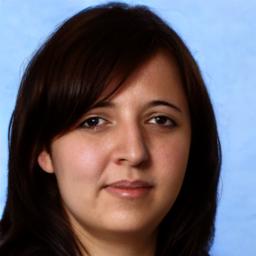} & 
		\includegraphics[height=\TSA]{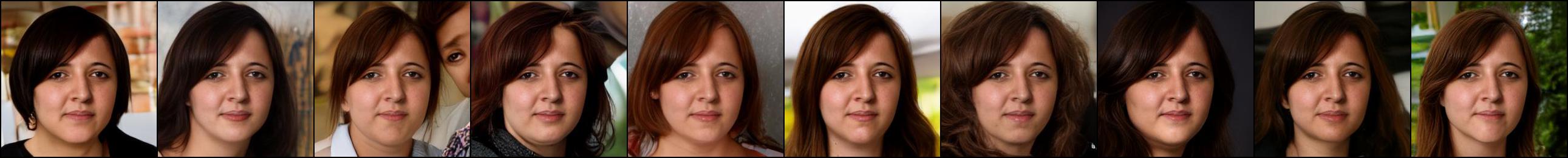} \\
		\includegraphics[height=\TSA]{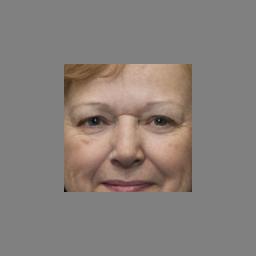} & 
		\includegraphics[height=\TSA]{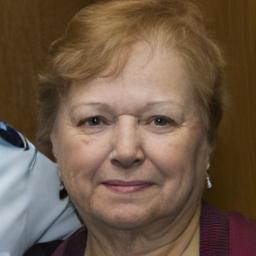} & 
		\includegraphics[height=\TSA]{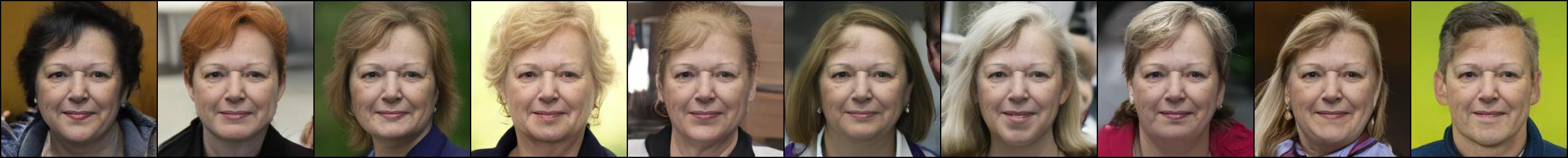} \\
		\includegraphics[height=\TSA]{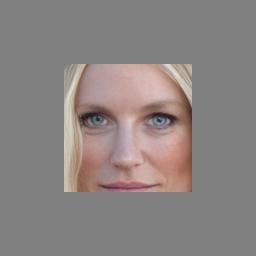} & 
		\includegraphics[height=\TSA]{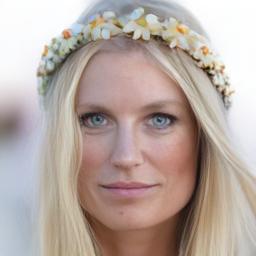} & 
		\includegraphics[height=\TSA]{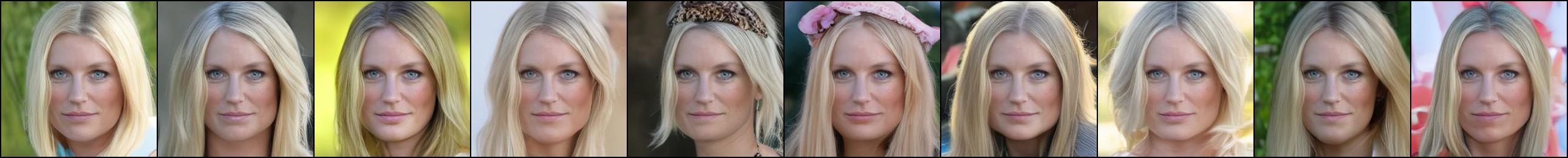} \\
		\includegraphics[height=\TSA]{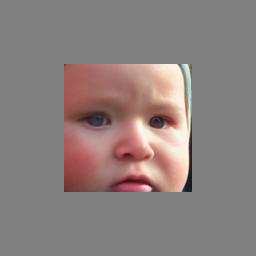} & 
		\includegraphics[height=\TSA]{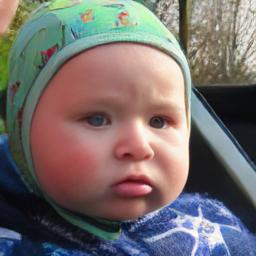} & 
		\includegraphics[height=\TSA]{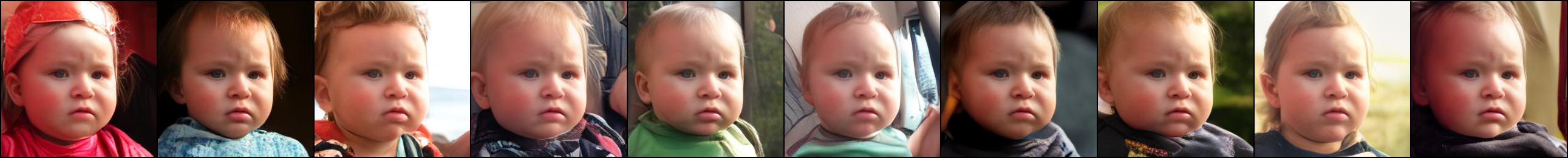} \\
		\includegraphics[height=\TSA]{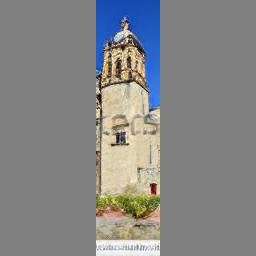} & 
		\includegraphics[height=\TSA]{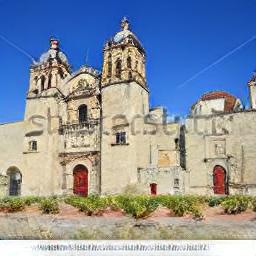} & 
		\includegraphics[height=\TSA]{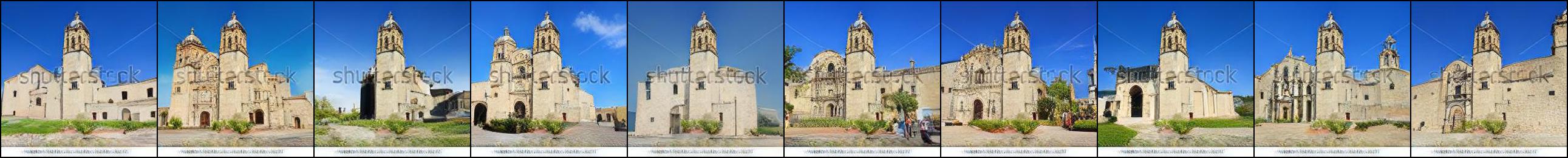} \\
		\includegraphics[height=\TSA]{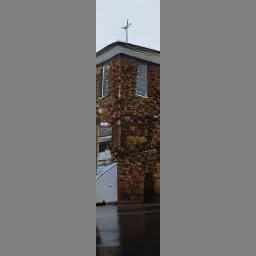} & 
		\includegraphics[height=\TSA]{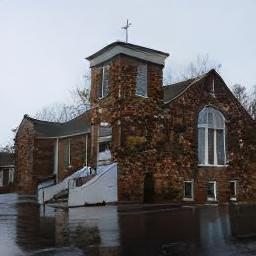} & 
		\includegraphics[height=\TSA]{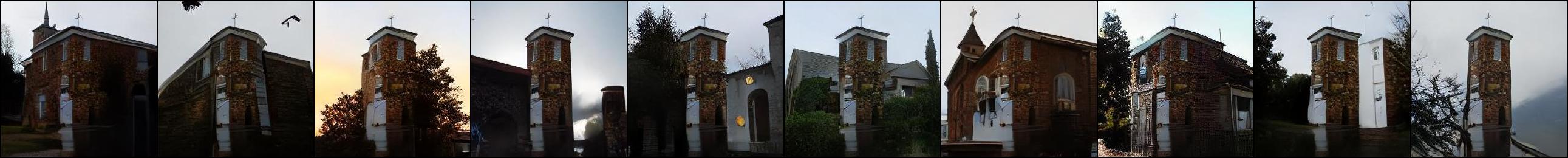} \\
		\includegraphics[height=\TSA]{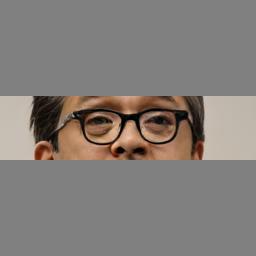} & 
		\includegraphics[height=\TSA]{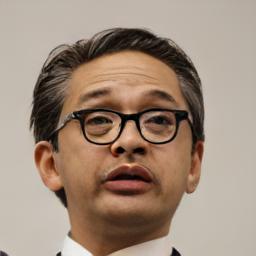} & 
		\includegraphics[height=\TSA]{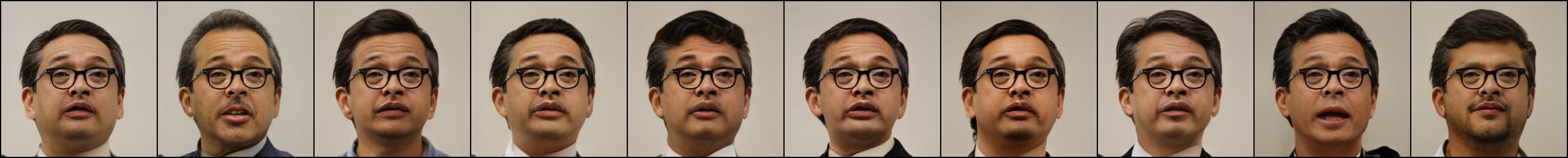} \\
        \includegraphics[height=\TSA]{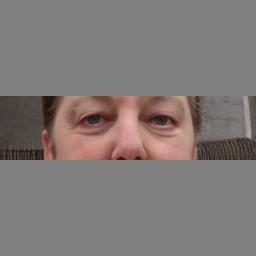} & 
		\includegraphics[height=\TSA]{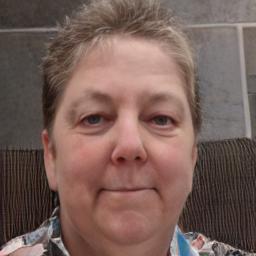} & 
		\includegraphics[height=\TSA]{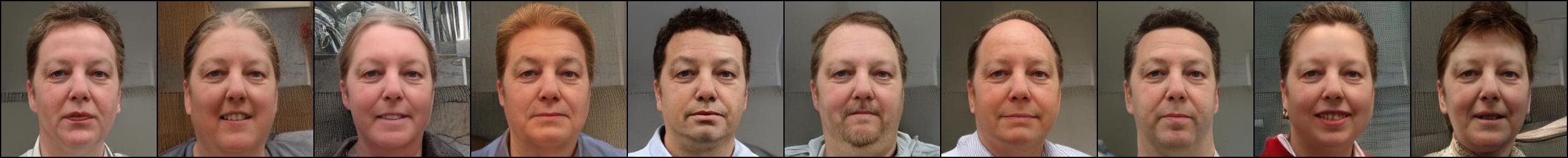} \\
        \Huge Condition & \Huge Original & \Huge Inpainting results \\
	\end{tabular}}
	\caption{Conditional image inpainting with different masking patterns.}
	\label{fig:app_inpaint}
\end{figure}

\clearpage
\begin{figure}[t]
	\centering
	\resizebox{\linewidth}{!}{
		\begin{tabular}{c|c}
		\includegraphics[height=\TSA]{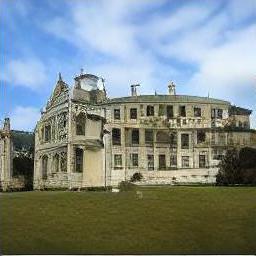} & 
		\includegraphics[height=\TSA]{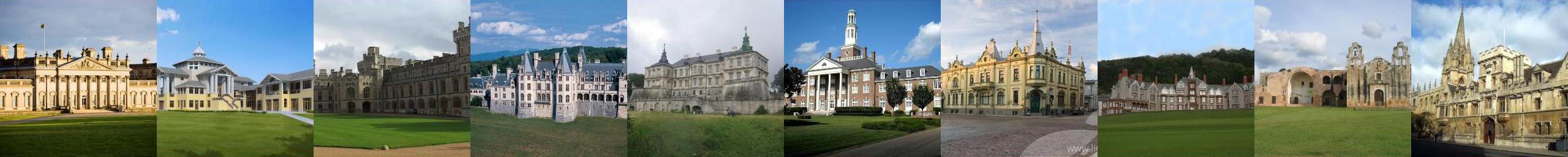} \\
		\includegraphics[height=\TSA]{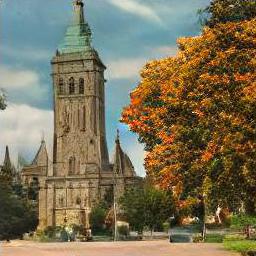} & 
		\includegraphics[height=\TSA]{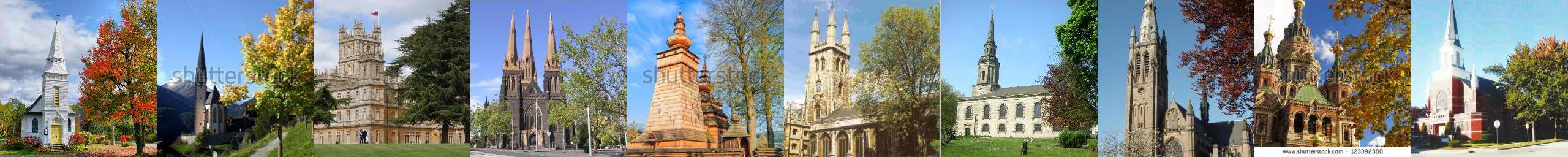} \\
		\includegraphics[height=\TSA]{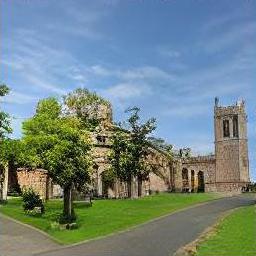} & 
		\includegraphics[height=\TSA]{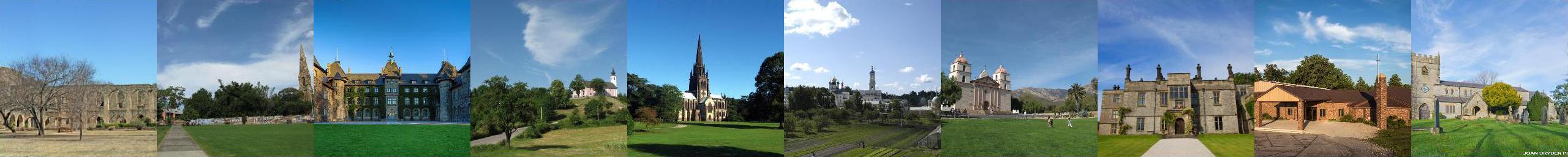} \\
		\includegraphics[height=\TSA]{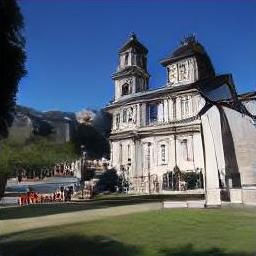} & 
		\includegraphics[height=\TSA]{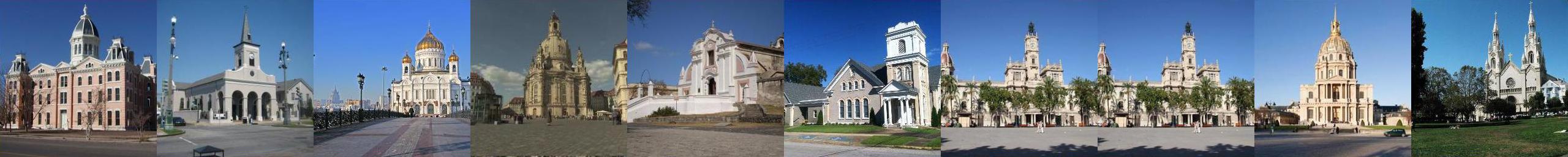} \\
		\includegraphics[height=\TSA]{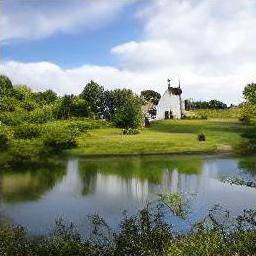} & 
		\includegraphics[height=\TSA]{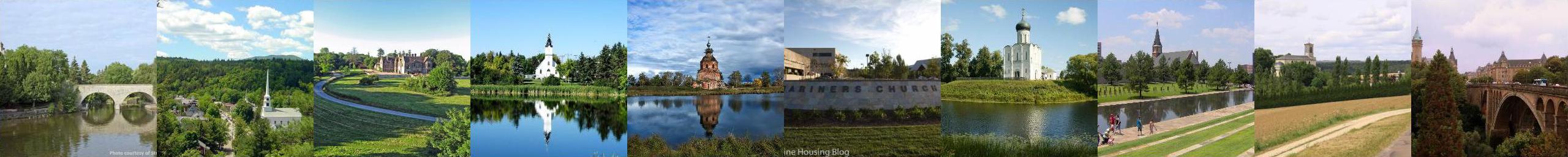} \\
		\includegraphics[height=\TSA]{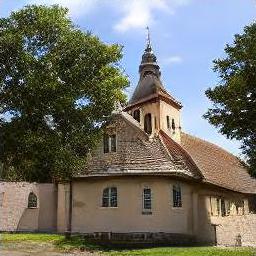} & 
		\includegraphics[height=\TSA]{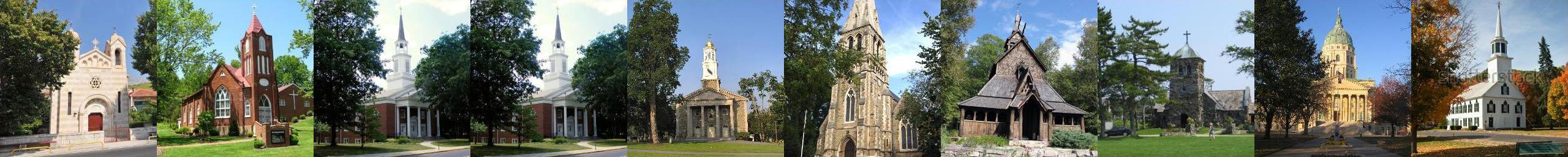} \\
		\includegraphics[height=\TSA]{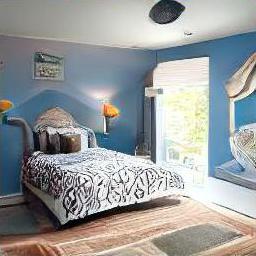} & 
		\includegraphics[height=\TSA]{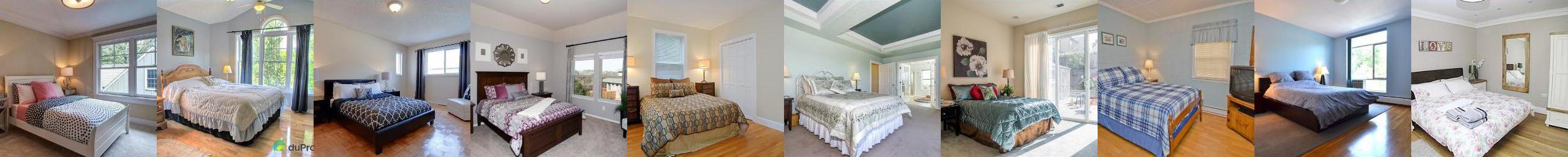} \\
		\includegraphics[height=\TSA]{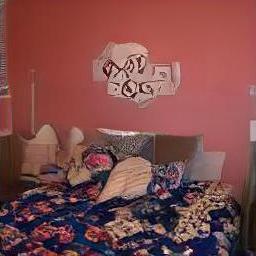} & 
		\includegraphics[height=\TSA]{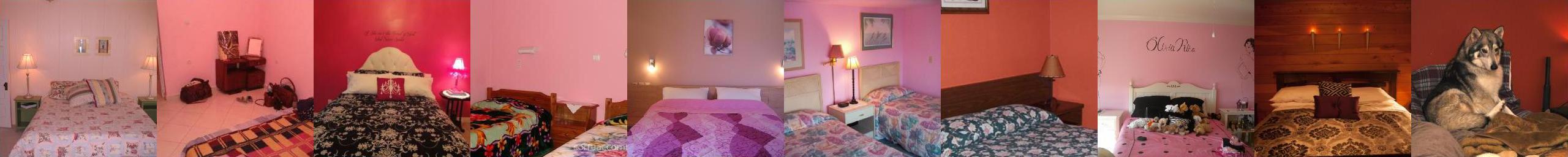} \\
		\includegraphics[height=\TSA]{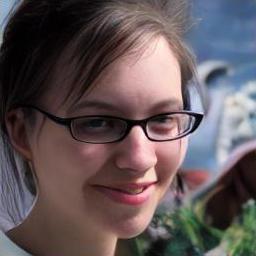} & 
		\includegraphics[height=\TSA]{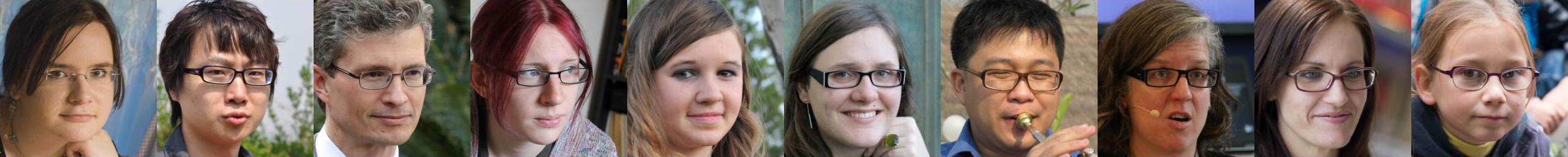} \\
		\includegraphics[height=\TSA]{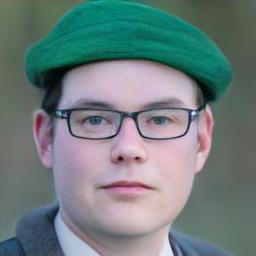} & 
		\includegraphics[height=\TSA]{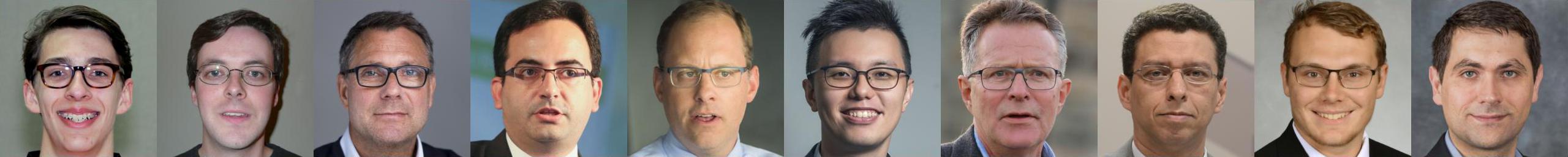} \\
		\includegraphics[height=\TSA]{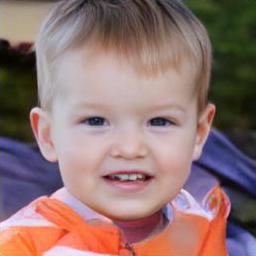} & 
		\includegraphics[height=\TSA]{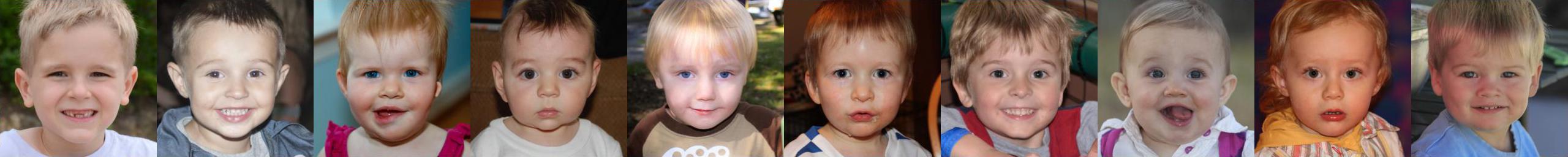} \\
		\includegraphics[height=\TSA]{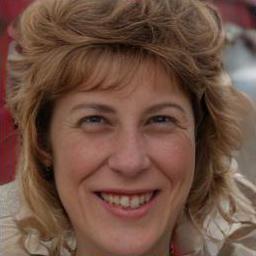} & 
		\includegraphics[height=\TSA]{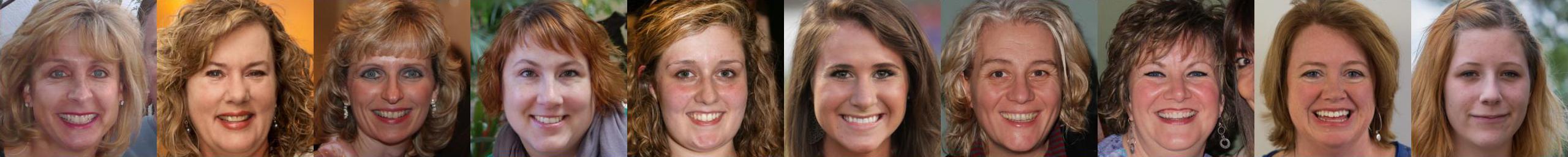} \\
		\includegraphics[height=\TSA]{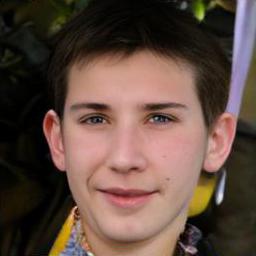} & 
		\includegraphics[height=\TSA]{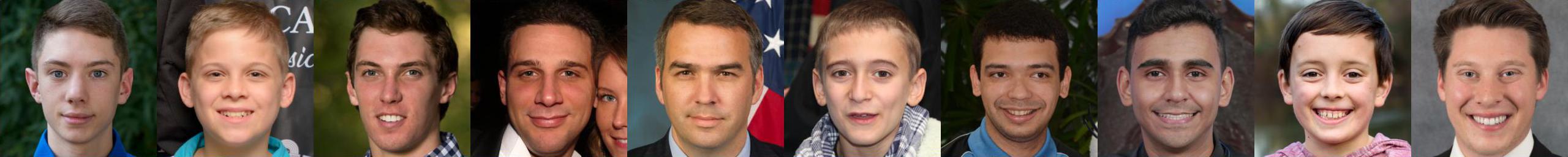} \\
		\includegraphics[height=\TSA]{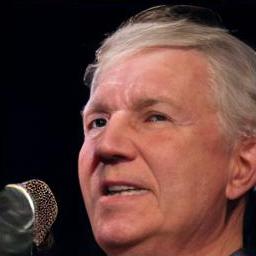} & 
		\includegraphics[height=\TSA]{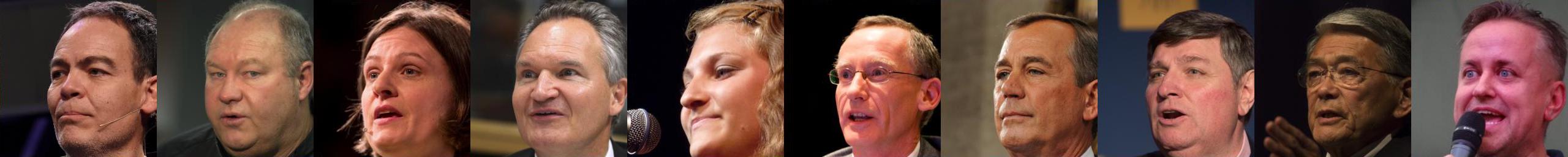} \\

        \Huge Generated & \Huge Nearest neighbours \\
	\end{tabular}}
	\caption{Top-10 nearest neighbours in the training datasets of our generated samples. Results show that our model is not overfitting to the training datasets.}
	\label{fig:nn}
\end{figure}

\end{document}